\documentclass[acmsmall]{acmart}

\usepackage{booktabs} % For formal tables
\usepackage{graphicx}
\usepackage{epsfig}
\usepackage{amsmath}
\usepackage{amsfonts}
\usepackage{array}
\usepackage{booktabs}
\usepackage{algorithm}
\usepackage{algorithmic}
\usepackage{multirow}
\usepackage{multicol}
\usepackage{subfigure}
\usepackage{color}
\usepackage{tcolorbox}
\usepackage{enumerate}
\usepackage{diagbox}
\usepackage{makecell}

\AtBeginDocument{%
  \providecommand\BibTeX{{%
    \normalfont B\kern-0.5em{\scshape i\kern-0.25em b}\kern-0.8em\TeX}}}

\setcopyright{acmcopyright}
\copyrightyear{2020}
\acmYear{2020}
\acmDOI{0000}

\acmJournal{TIST}
\acmVolume{00}
\acmNumber{0}
\acmArticle{000}
\acmMonth{0}

\begin{document}

\title{Graph Neural Networks: Taxonomy, Advances and Trends}


\author{Yu Zhou}
\authornote{Corresponding Author.}
\email{zhouyu@tyut.edu.cn}
\orcid{0000-0002-0304-0863}
\affiliation{%
  \institution{College of Data Science/Shanxi Spatial Information Network Engineering Technology Research Center, Taiyuan University of Technology}
  \streetaddress{No. 79 Yingze West Street, WanBaiLin District}
  \city{Taiyuan}
  \state{Shanxi}
  \country{China}
  \postcode{030024}
}

\author{Haixia Zheng}
\authornote{Corresponding Author.}
\email{zhenghaixia@tyut.edu.cn}
\affiliation{%
  \institution{College of Data Science/Shanxi Spatial Information Network Engineering Technology Research Center, Taiyuan University of Technology}
  \streetaddress{No. 79 Yingze West Street, WanBaiLin District}
  \city{Taiyuan}
  \state{Shanxi}
  \country{China}
  \postcode{030024}
}

\author{Xin Huang}
\email{huangxin@tyut.edu.cn}
\affiliation{%
  \institution{College of Data Science, Taiyuan University of Technology}
  \streetaddress{No. 79 Yingze West Street, WanBaiLin District}
  \city{Taiyuan}
  \state{Shanxi}
  \country{China}
  \postcode{030024}
}

\author{Shufeng Hao}
\email{haoshufeng@tyut.edu.cn}
\affiliation{%
	\institution{College of Data Science, Taiyuan University of Technology}
	\streetaddress{No. 79 Yingze West Street, WanBaiLin District}
	\city{Taiyuan}
	\state{Shanxi}
	\country{China}
	\postcode{030024}
}

\author{Dengao Li}
\email{lidengao@tyut.edu.cn}
\affiliation{%
  \institution{College of Data Science/Shanxi Spatial Information Network Engineering Technology Research Center, Taiyuan University of Technology}
  \streetaddress{No. 79 Yingze West Street, WanBaiLin District}
  \city{Taiyuan}
  \state{Shanxi}
  \country{China}
  \postcode{030024}
}

\author{Jumin Zhao}
\email{zhaojumin@tyut.edu.cn}
\affiliation{%
  \institution{College of Information and Computer/Shanxi Intelligent Perception Engineering Research Center, Taiyuan University of Technology}
  \streetaddress{No. 79 Yingze West Street, WanBaiLin District}
  \city{Taiyuan}
  \state{Shanxi}
  \country{China}
  \postcode{030024}
}

\renewcommand{\shortauthors}{Yu Zhou, et al.}

\begin{abstract}
  Graph neural networks provide a powerful toolkit for embedding real-world graphs into low-dimensional spaces according to specific tasks.
  Up to now, there have been several surveys on this topic. However, they usually lay emphasis on different angles so that the readers can not see a panorama of the graph neural networks.
  This survey aims to overcome this limitation, and provide a systematic and comprehensive review on the graph neural networks. First of all, we provide a novel taxonomy for the graph neural networks,
  and then refer to up to 250 relevant literatures to show the panorama of the graph neural networks. All of them are classified into the corresponding categories.
  In order to drive the graph neural networks into a new stage, we summarize four future research directions so as to overcome the facing challenges.
  It is expected that more and more scholars can understand and exploit the graph neural networks, and use them in their research community.
\end{abstract}

\begin{CCSXML}
<ccs2012>
   <concept>
       <concept_id>10010147.10010257.10010293.10010294</concept_id>
       <concept_desc>Computing methodologies~Neural networks</concept_desc>
       <concept_significance>500</concept_significance>
       </concept>
   <concept>
       <concept_id>10010147.10010257.10010293.10010319</concept_id>
       <concept_desc>Computing methodologies~Learning latent representations</concept_desc>
       <concept_significance>500</concept_significance>
       </concept>
 </ccs2012>
\end{CCSXML}

\ccsdesc[500]{Computing methodologies~Neural networks}
\ccsdesc[500]{Computing methodologies~Learning latent representations}

\keywords{Graph Convolutional Neural Network, Graph Recurrent Neural Network, Graph Pooling Operator, Graph Attention Mechanism, Graph Neural Network}

\maketitle

%%%%%%%%%%%%%%%%%%%%%%%%%%%%%%%%%%%%%%%%%%%%%%%%%%%%%%%%%%%%%%%%%%%%%%%%%%%%%%%%
\section{Introduction}
%%%%%%%%%%%%%%%%%%%%%%%%%%%%%%%%%%%%%%%%%%%%%%%%%%%%%%%%%%%%%%%%%%%%%%%%%%%%%%%%

Graph, as a complex data structure, consists of nodes (or vertices) and edges (or links). It can be used to model lots of complex systems in real world, e.g. social networks, protein-protein interaction networks,
brain networks, road networks, physical interaction networks and knowledge graph etc. Thus, Analyzing the complex networks becomes an intriguing research frontier.
With the rapid development of deep learning techniques, many scholars employ the deep learning architectures to tackle the graphs.
Graph Neural Networks (GNNs) emerge under these circumstances. Up to now, the GNNs have evolved into a prevalent and powerful computational framework for tackling irregular data such as graphs and manifolds.

The GNNs can learn task-specific node/edge/graph representations via hierarchical iterative operators so that the traditional machine learning methods can be employed
to perform graph-related learning tasks, e.g. node classification, graph classification, link prediction and clustering etc. Although the GNNs has attained substantial success over the graph-related learning tasks,
they still face great challenges. Firstly, the structural complexity of graphs incurs expensive computational cost on large graphs.
Secondly, perturbing the graph structure and/or initial features incurs sharp performance decay. Thirdly, the Wesfeiler-Leman (WL) graph isomorphism test impedes the performance improvement of the GNNs.
At last, the blackbox work mechanism of the GNNs hinders safely deploying them to real-world applications.

In this paper, we generalize the conventional deep architectures to the non-Euclidean domains,
and summarize the architectures, extensions and applications, benchmarks and evaluation pitfalls and future research directions of the graph neural networks.
Up to now, there have been several surveys on the GNNs. However, they usually discuss the GNN models from different angles and with different emphasises.
To the best of our knowledge, the first survey on the GNNs was conducted by Michael M. Bronstein et al\cite{1}.
Peng Cui et al\cite{2} reviewed different kinds of deep learning models applied to graphs from three aspects: semi-supervised learning methods including
graph convolutional neural networks, unsupervised learning methods including graph auto-encoders, and recent advancements including graph recurrent neural networks and graph reinforcement learning.
This survey laid emphasis on semi-supervised learning models, i.e. the spatial and spectral graph convolutional neural networks, yet comparatively less emphasis on the other two aspects.
Due to the space limit, this survey only listed a few of key applications of the GNNs, but ignored the diversity of the applications.
Maosong Sun et al\cite{3} provided a detailed review of the spectral and spatial graph convolutional neural networks from three aspects: graph types, propagation step and training method,
and divided its applications into three scenarios: structural scenarios, non-structural scenarios and other scenarios. However, this article did not involve the other GNN architectures
such as graph auto-encoders, graph recurrent neural networks and graph generative networks.
Philip S. Yu et al\cite{4} conducted a comprehensive survey on the graph neural networks, and investigated available datasets, open-source implementations and practical applications.
However, they only listed a few of core literatures on each research topic.
Davide Bacciu et al\cite{367} gives a gentle introduction to the field of deep learning for graph data. The goal of this article is to introduce the main concepts
and building blocks to construct neural networks for graph data, and therefore it falls short of an exposition of recent works on graph neural networks.

\begin{figure}[htb]
  \centering
  \includegraphics[width=0.9\textwidth]{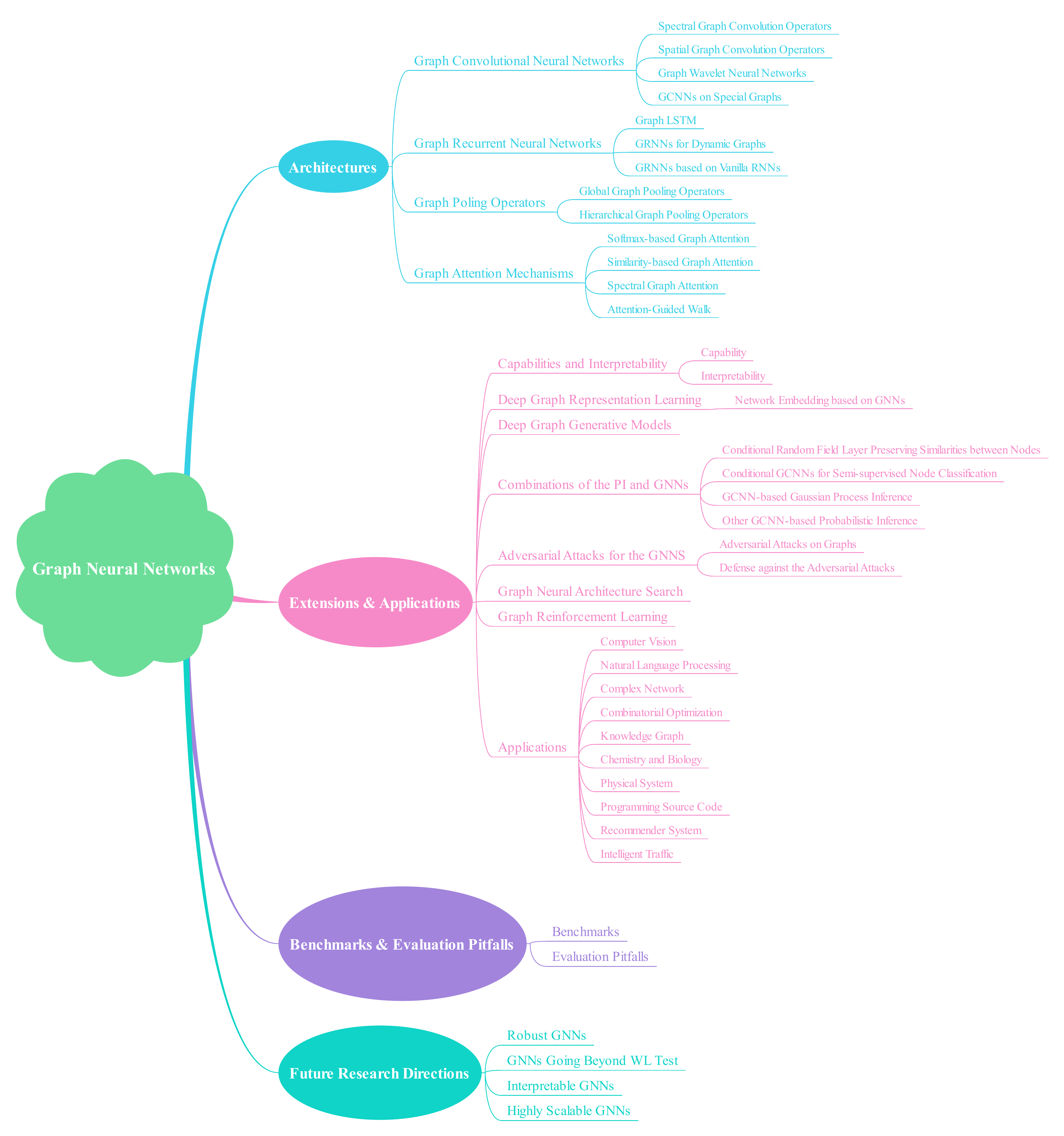}
  \caption{The architecture of this paper.}
  \label{GNN Panorama}
\end{figure}

It  is noted that all of the aforementioned surveys do not concern capability and interpretability of GNNs, combinations of the probabilistic inference and GNNs,
and adversarial attacks on graphs. In this article, we provide a panorama of GNNs for readers from 4 perspectives:
architectures, extensions and applications, benchmarks and evaluations pitfalls, future research directions,
as shown in Fig. \ref{GNN Panorama}. For the architectures of GNNs, we investigate the studies on graph convolutional neural networks (GCNNs), graph pooling operators, graph attention mechanisms and
graph recurrent neural networks (GRNNs). The extensions and applications demonstrate some notable research topics on the GNNs through integrating the above architectures.
Specifically, this perspective includes the capabilities and interpretability, deep graph representation learning, deep graph generative models, combinations of the Probabilistic Inference (PI) and the GNNs,
adversarial attacks for GNNs, Graph Neural Architecture Search and graph reinforcement learning and applications. In summary, our article provides a complete taxonomy for GNNs,
and comprehensively review the current advances and trends of the GNNs. These are our main differences from the aforementioned surveys.

\textbf{Contributions.} Our main contributions boils down to the following three-fold aspects.
\begin{enumerate}
  \item We propose a novel taxonomy for the GNNs, which has three levels. The first includes architectures, benchmarks and evaluation pitfalls, and applications. The architectures are classified into 9 categories, the benchmarks and evaluation pitfalls into 2 categories, and the applications into 10 categories. Furthermore, the graph convolutional neural networks, as a classic GNN architecture, are again classified into 6 categories.
  \item We provide a comprehensive review of the GNNs. All of the literatures fall into the corresponding categories. It is expected that the readers not only understand the panorama of the GNNs, but also comprehend the basic principles and various computation modules of the GNNs through reading this survey.
  \item We summarize four future research directions for the GNNs according to the current facing challenges, most of which are not mentioned the other surveys. It is expected that the research on the GNNs can progress into a new stage by overcoming these challenges.
\end{enumerate}

\textbf{Roadmap.} The remainder of this paper is organized as follows. First of all, we provide some basic notations and definitions that will be often used in the following sections.
Then, we start reviewing the GNNs from 4 aspects: architectures in section 3, extensions and applications in section 4, benchmarks and evaluation pitfalls in section 5
and future research directions in section 6. Finally, we conclude our paper.

%%%%%%%%%%%%%%%%%%%%%%%%%%%%%%%%%%%%%%%%%%%%%%%%%%%%%%%%%%%%%%%%%%%%%%%%%%%%%%%%
\section{Preliminaries}
%%%%%%%%%%%%%%%%%%%%%%%%%%%%%%%%%%%%%%%%%%%%%%%%%%%%%%%%%%%%%%%%%%%%%%%%%%%%%%%%

In this section, we introduce relevant notations so as to conveniently describe the graph neural network models.
A simple  graph can be denoted by $G=(V,E)$ where $V$ and $E$ respectively denote the set of $N$ nodes (or vertices) and $M$ edges.
Without loss of generality, let $V=\left\{v_1,\cdots,v_N\right\}$ and $E=\left\{e_1,\cdots,e_M\right\}$. Each edge $e_j\in E$ can be denoted by $e_j=\left(v_{s_j},v_{r_j}\right)$ where $v_{s_j}, v_{r_j}\in V$.
Let $A_G$ denote the adjacency matrix of $G$ where $A_G(s,r)=1$ iff there is an edge between $v_s$ and $v_r$. If $G$ is edge-weighted, $A_G(s,r)$ equals the weight value of the edge $\left(v_s,v_r\right)$.
If $G$ is directed, $\left(v_{s_j},v_{r_j}\right)\neq\left(v_{r_j}, v_{s_j}\right)$ and therefore $A_G$ is asymmetric.
A directed edge $e_j=\left(v_{s_j},v_{r_j}\right)$ is also called an arch, i.e. $e_j=\left\langle v_{s_j},v_{s_j}\right\rangle$.
Otherwise $\left(v_{s_j},v_{r_j}\right)=\left(v_{r_j}, v_{s_j}\right)$ and $A_G$ is symmetric. For a node $v_s\in V$, let $N_G(v_s)$ denote the set of neighbors of $v_s$, and $d_G(v_s)$ denote the degree of $v_s$.
If $G$ is directed, let $N_G^+(v_s)$ and $N_G^-(v_s)$ respectively denote the incoming and outgoing neighbors of $v_s$, and $d_G^+(v_s)$ and $d_G^-(v_s)$ respectively denote the incoming and outgoing degree of $v_s$.
Given a vector $a=(a_1,\cdots,a_N)\in\mathbb{R}^N$, $\text{diag}(a)$ (or $\text{diag}(a_1,\cdots,a_N)$) denotes a diagonal matrix consisting of the elements $a_n, n=1,\cdots,N$.

A vector $x\in\mathbb{R}^{N}$ is called a 1-dimensional graph signal on $G$. Similarly, $X\in\mathbb{R}^{N\times d}$ is called a $d\text{-dimensiaonl}$ graph signal on $G$.
In fact, $X$ is also called a feature matrix of nodes on $G$. Without loss of generality, let $X[j,k]$ denote the $(j,k)\text{-th}$ entry of the matrix $X\in\mathbb{R}^{N\times d}$,
$X[j,:]\in\mathbb{R}^d$ denote the feature vector of the node $v_j$ and $X[:,j]$ denote the $1\text{-dimensional}$ graph signal on $G$.
Let $\mathbb{I}_N$ denote a $N\times N$ identity matrix. For undirected graphs, $L_G=D_G-A_G$ is called the Laplacian matrix of $G$,
where $D_G[r,r]=\sum_{c=1}^N A_G[r,c]$. For a 1-dimensional graph signal $x$, its smoothness $s(x)$ is defined as
\begin{equation}\label{SmoothnessOfGraphSignal}
s(x)=x^TL_Gx=\frac{1}{2}\sum_{r,c=1}^NA_G(r,c)\left(x[r]-x[c]\right)^2.
\end{equation}
The normalization of $L_G$ is defined by $\overline{L}_G=\mathbb{I}_N-D_G^{-\frac{1}{2}}A_GD_G^{-\frac{1}{2}}$.
$\overline{L}_G$ is a real symmetric semi-positive definite matrix. So, it has $N$ ordered real non-negative eigenvalues $\left\{\lambda_n:n=1,\cdots,N\right\}$
and corresponding orthonormal eigenvectors $\left\{u_n:n=1,\cdots,N\right\}$, namely $\overline{L}_G=U\Lambda U^T$ where $\Lambda=\text{diag}(\lambda_1,\cdots,\lambda_N)$
and $U=\left(u_1,\cdots,u_N\right)$ denotes a orthonormomal matrix. Without loss of generality, $0=\lambda_1\leq\lambda_2\leq\cdots\leq\lambda_N=\lambda_{\text{max}}$.
The eigenvectors $u_n, n=1,\cdots,N$ are also called the graph Fourier bases of $G$. Obviously, the graph Fourier basis are also the 1-dimensional graph signal on $G$.
The graph Fourier transform\cite{5} for a given graph signal $x$ can be denoted by
\begin{equation}\label{GraphFourierTransform}
\hat{x}\triangleq\mathcal{F}(x)=U^Tx.
\end{equation}
The inverse graph Fourier transform can be correspondingly denoted by
\begin{equation}\label{InverseGraphFourierTransform}
x\triangleq\mathcal{F}^{-1}(\hat{x})=U\hat{x}.
\end{equation}
Note that the eigenvalue $\lambda_n$ actually measures the smoothness of the graph Fourier mode $u_n$.
Throughout this paper, let $\rho(\cdot)$ denote an activation function, $\bowtie$ denote the concatenation of at least two vectors, and $\left<\right>$ denote the inner product of two vectors/matrices.
We somewhere use the function $\text{Concat}(\cdot)$ to denote the concatenation of two vectors as well.

%%%%%%%%%%%%%%%%%%%%%%%%%%%%%%%%%%%%%%%%%%%%%%%%%%%%%%%%%%%%%%%%%%%%%%%%%%%%%%%%
\section{Architectures}
%%%%%%%%%%%%%%%%%%%%%%%%%%%%%%%%%%%%%%%%%%%%%%%%%%%%%%%%%%%%%%%%%%%%%%%%%%%%%%%%

%%%%%%%%%%%%%%%%%%%%%%%%%%%%%%%%%%%%%%%%%%%%%%%%%%%%%%%%%%%%%%%%%%%%%%%%%%%%%%%%
\subsection{Graph Convolutional Neural Networks (GCNNs)}
%%%%%%%%%%%%%%%%%%%%%%%%%%%%%%%%%%%%%%%%%%%%%%%%%%%%%%%%%%%%%%%%%%%%%%%%%%%%%%%%

The GCNNs play pivotal roles on tackling the irregular data (e.g. graph and manifold). They are motivated by the Convolutional Neural Networks (CNNs) to learn hierarchical representations of irregular data.
There have been some efforts to generalize the CNN to graphs \cite{27,31,48}. However, they are usually computationally expensive and cannot capture spectral or spatial features.
Below, we introduce the GCNNs from the next 6 aspects: spectral GCNNs, spatial GCNNs, Graph wavelet neural networks and GCNNs on special graphs.

%%%%%%%%%%%%%%%%%%%%%%%%%%%%%%%%%%%%%%%%%%%%%%%%%%%%%%%%%%%%%%%%%%%%%%%%%%%%%%%%
\subsubsection{Spectral Graph Convolution Operators}
%%%%%%%%%%%%%%%%%%%%%%%%%%%%%%%%%%%%%%%%%%%%%%%%%%%%%%%%%%%%%%%%%%%%%%%%%%%%%%%%

\begin{figure}[htb]
  \centering
  \includegraphics[width=1.0\textwidth]{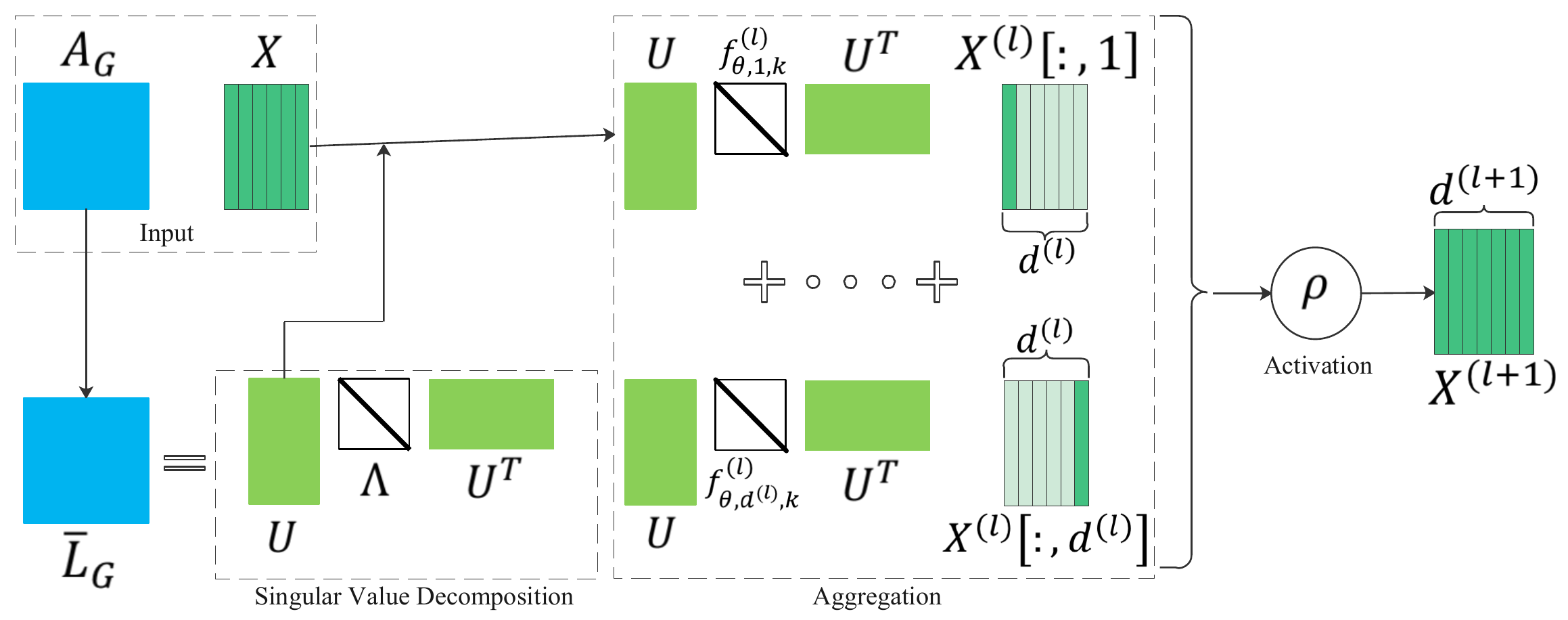}
  \caption{Computational framework of the spectral GCNN.}
  \label{Spectral GCNN}
\end{figure}

The spectral graph convolution operator is defined via the graph Fourier transform. For two graph signals $x$ and $y$ on $G$,
their spectral graph convolution $x\ast_Gy$ is defined by
\begin{equation}\label{Spectral Graph Convolution}
\begin{array}{rcl}
x\ast_G y & = & \mathcal{F}^{-1}(\mathcal{F}(x)\circledast\mathcal{F}(y)) \\ [2mm]
          & = & U\left(U^Tx\circledast U^Ty\right) \\[2mm]
          & = & U\text{diag}(U^Ty)U^Tx,
\end{array}
\end{equation}
where $\circledast$ denotes the element-wise Hadamard product \cite{7,18,27}.
The spectral graph convolution can be rewritten as
\begin{displaymath}
x\ast_G f_{\theta}=Uf_{\theta}U^Tx,
\end{displaymath}
where $f_{\theta}$ is a diagonal matrix consisting of the learnable parameters.
That is, the signal $x$ is filtered by the spectral graph filter (or graph convolution kernel) $f_{\theta}$.
For a $d^{(l)}\text{-dimensional}$ graph signal $X^{(l)}$ on $G$,
the output $X^{(l+1)}$ yielded by a graph convolution layer, namely $d^{(l+1)}\text{-dimensional}$ graph signal on $G$,
can be written as
\begin{equation}
\label{Spectral GCNN Formula}
X^{(l+1)}[:,k]=\rho\left(\sum_{j=1}^{d^{(l)}}Uf_{\theta,j,k}^{(l)}U^TX^{(l)}[:,j]\right),
\end{equation}
where $f_{\theta,j,k}^{(l)}$ is a spectral graph filter, i.e. a $N\times N$ diagonal matrix consisting of learnable parameters corresponding to the $j\text{-th}$ graph signal
at $l\text{-th}$ layer and the $k\text{-th}$ graph signal at $(l+1)\text{-th}$ layer.
The computational framework of the spectral GCNN in Eq. (\ref{Spectral GCNN Formula}) is demonstrated in Fig. \ref{Spectral GCNN}.
It is worth noting that the calculation of the above graph convolution layer takes $O(N^3)$ time and $O(N^2)$ space
to perform the eigendecomposition of $\overline{L}_G$ especially for large graphs.
The article \cite{179} proposes a regularization technique, namely GraphMix, to augment the vanilla GCNN with a parameter-sharing Fully Connected Network (FCN).

\textbf{Spectral Graph Filter.} Many studies \cite{27} focus on designing different spectral graph filters.
In order to circumvent the eigendecomposition, the spectral graph filter $f_{\theta}$ can formulated as a $K\text{-localized}$ polynomial of the eigenvalues
of the normalized graph Laplacian $\overline{L}_G$ \cite{6,8,78},
i.e.
\begin{equation}\label{Chebyshev Spectral Filter}
f_{\theta}=f_{\theta}(\Lambda)\triangleq\sum_{k=0}^{K-1}\theta_k\Lambda^k.
\end{equation}
In practice, the $K\text{-localized}$ Chebyshev polynomial \cite{78} is a favorable choice of formulating the spectral graph filter, i.e.
\begin{displaymath}
f_{\theta}(\Lambda)=\sum_{k=0}^{K-1}\theta_kT_k(\widetilde{\Lambda}),
\end{displaymath}
where the Chebyshev polynomial is defined as
\begin{equation}\label{Chebeshev Polynomial}
T_0(x)=1,\quad T_1(x)=x,\quad T_k(x)=2xT_{k-1}(x)-T_{k-2}(x)
\end{equation}
and $\widetilde{\Lambda}=\frac{2}{\lambda_{\text{max}}}\Lambda-\mathbb{I}_N$.
The reason why $\widetilde{\Lambda}=\frac{2}{\lambda_{\text{max}}}\Lambda-\mathbb{I}_N$ is because it can map eigenvalues $\lambda\in\left[0,\lambda_{\text{max}}\right]$ into $[-1,1]$.
This filter is $K\text{-localized}$ in the sense that it leverages information from nodes which are at most $K\text{-hops}$ away.
In order to further decrease the computational cost, the $1\text{st-order}$ Chebyshev polynomial is used to define the spectral graph filter. Specifically, it lets $\lambda_{\text{max}}\approx 2$
(because the largest eigenvalue of $\overline{L}_G$ is less than or equal to 2 \cite{374}) and $\theta=\theta_0=-\theta_1$.
Moreover, the renormalization trick is used here to mitigate the limitations of the vanishing/exploding gradient,
namely substituting $\widetilde{D}_G^{-\frac{1}{2}}\widetilde{A}_G\widetilde{D}_G^{-\frac{1}{2}}$ for $\mathbb{I}_N+D_G^{-\frac{1}{2}}A_GD_G^{-\frac{1}{2}}$ where $\widetilde{A}_G=A_G+\mathbb{I}_N$
and $\widetilde{D}_G=\text{diag}\left(\sum_{k=1}^{N}\widetilde{A}[1,k],\cdots,\sum_{k=1}^{N}\widetilde{A}[N,k]\right)$.
As a result, the Graph Convolutional Network (GCN) \cite{6,375} can be defined as
\begin{equation}\label{GCN-Layer}
X^{(l+1)}=\rho\left(\widetilde{D}_G^{-\frac{1}{2}}\widetilde{A}_G\widetilde{D}_G^{-\frac{1}{2}}X^{(l)}\Theta^{(l)}\right).
\end{equation}
The Chebyshev spectral graph filter suffers from a drawback that the spectrum of $\overline{L}_G$ is linearly mapped into $\left[-1,1\right]$.
This drawback makes it hard to specialize in the low frequency bands.
In order to mitigate this problem, Michael M. Bronstein et al \cite{12} proposes the Cayley spectral graph filter
via the $\text{order-}r$ Cayley polynomial $f_{c,h}(\lambda)=c_0+2\text{Re}\left(\sum_{j=0}^{r}c_h\mathcal{C}(\lambda)^j\right)$ with
the Cayley transform $\mathcal{C}(\lambda)=\frac{\lambda-i}{\lambda+i}$.
Moreover, there are many other spectral graph filters, e.g. \cite{8,14,19,22,30,37,38,39,44,90}.
In addition, some studies employ the capsule network \cite{400} to construct capsule-inspired GNNs \cite{52,53,54}.

\textbf{Overcoming Time and Memory Challenges.} A chief challenge for GCNNs is that their training cost is strikingly expensive, especially on huge and sparse graphs.
The reason is that the GCNNs require full expansion of neighborhoods for the feed-forward computation of each node,
and large memory space for storing intermediate results and outputs.
In general, two approaches, namely sampling \cite{9,10,11,25} and decomposition \cite{23,24}, can be employed to mitigate the time and memory challenges for the spectral GCNNs.

\textbf{Depth Trap of Spectral GCNNs.} A bottleneck of GCNNs is that their performance maybe decease with ever-increasing number of layer.
This decay is often attributed to three factors: (1) overfitting resulting from the ever-increasing number of parameters;
(2) gradient vanishing/explosion during training; (3) oversmoothing making vertices from different clusters more and more indistinguishable.
The reason for oversmoothing is that performing the Laplacian smoothing many times forces the features of vertices
within the same connected component to stuck in stationary points \cite{16}.
There are some available approaches, e.g. \cite{33,373,379,398,399}, to circumvent the depth trap of the spectral GCNNs.

%%%%%%%%%%%%%%%%%%%%%%%%%%%%%%%%%%%%%%%%%%%%%%%%%%%%%%%%%%%%%%%%%%%%%%%%%%%%%%%%
\subsubsection{Spatial Graph Convolution Operators}
%%%%%%%%%%%%%%%%%%%%%%%%%%%%%%%%%%%%%%%%%%%%%%%%%%%%%%%%%%%%%%%%%%%%%%%%%%%%%%%%

Original spatial GCNNs \cite{47,76,77,380} constitutes a transition function, which must be a contraction map in order to ensure the uniqueness of states, and an update function.
In the following, we firstly introduce a generic framework of the spatial GCNN, and then investigate its variants.

\begin{figure}[htb]
  \centering
  \includegraphics[width=1.0\textwidth]{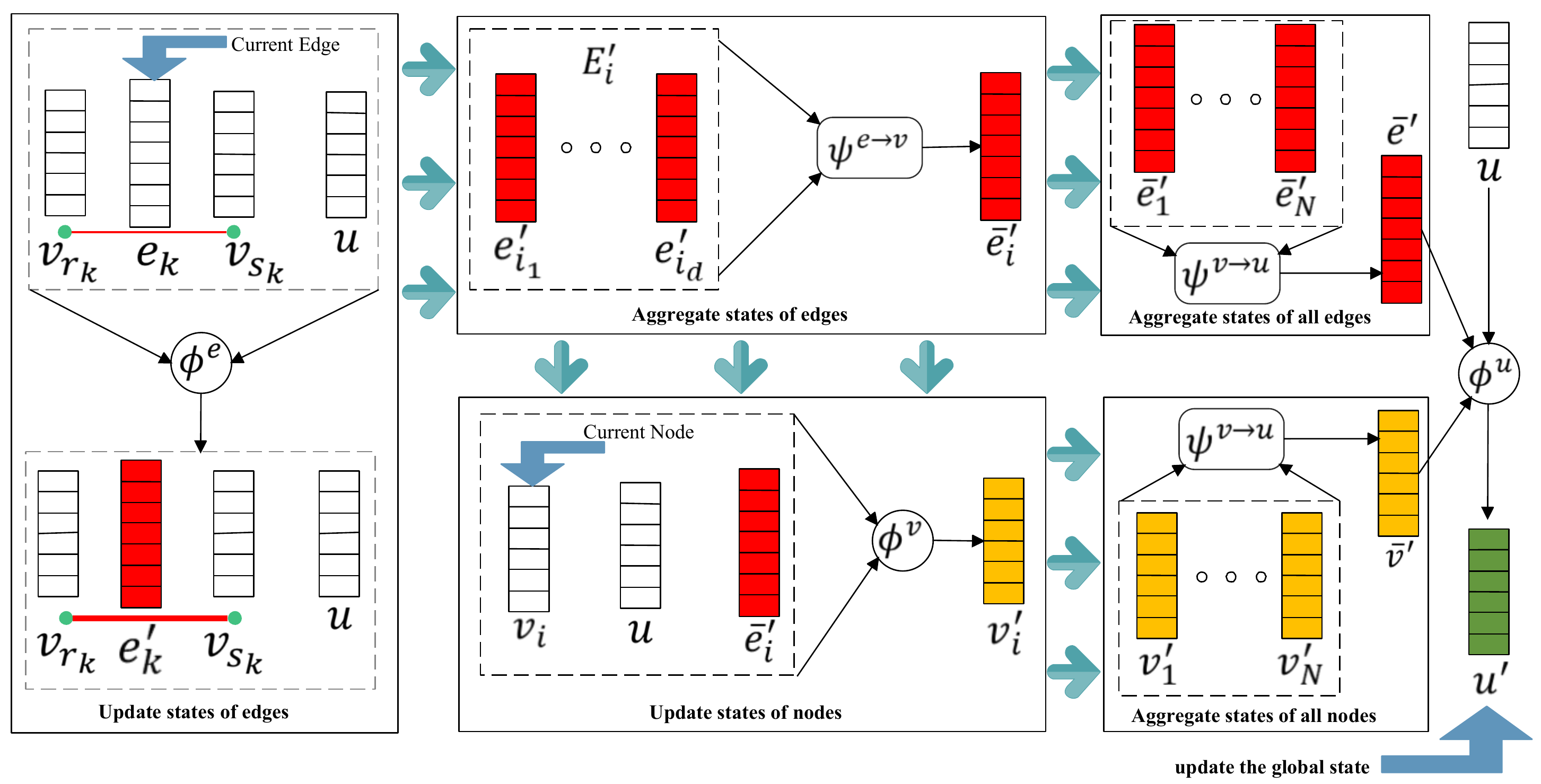}
  \caption{Computational framework of the spatial GCNN.}
  \label{Spatial GCNN}
\end{figure}

Graph networks (GNs) as generic architectures with relational inductive bias \cite{85} provide an elegant interface
for learning entities, relations and structured knowledge.
Specifically, GNs are composed of GN blocks in a sequential, encode-process-decode or recurrent manner. GN blocks contain three kinds of update functions,
namely $\phi^{e}(\cdot),\phi^{v}(\cdot),\phi^{u}(\cdot)$, and three kinds of aggregation functions, namely $\psi^{e\rightarrow v}(\cdot), \psi^{e\rightarrow u}(\cdot), \psi^{v\rightarrow u}(\cdot)$.
The iterations are described as follows.
\begin{equation}\label{GN-BLOCK}
\begin{array}{lll}
e'_k=\phi^{e}\left(e_k,v_{r_k},v_{s_k},u\right), & \bar{e}'_i=\psi^{e\rightarrow v}\left(E'_i\right), & v'_i=\phi^{v}\left(v_i,\bar{e}'_i,u\right), \\[2mm]
\bar{e}'=\psi^{e\rightarrow u}\left(E'\right), & u'=\phi^{u}\left(u,\bar{e}',\bar{v}'\right), & \bar{v}'=\psi^{v\rightarrow u}\left(V'\right)
\end{array}
\end{equation}
where $e_k$ is an arch from $v_{s_k}$ to $v_{r_k}$, $E'_i=\left\{(e'_k,s_k,r_k):r_k=i,k=1,\cdots,M\right\}$, $V'=\left\{v'_i:i=1,\cdots,N\right\}$ and $E'=\left\{(e'_k,s_k,r_k):k=1,\cdots,M\right\}$,
see Fig. \ref{Spatial GCNN}. It is noted that the aggregation functions should be invariant to any permutations of nodes or edges.
In practice, the GN framework can be used to implement a wide variety of architectures in accordance with three key design principles,
namely flexible representations, configuable within-block structure and flexible multi-block architectures.
Below, we introduce three prevalent variants of the GNs, namely Message Passing Neural Networks (MPNNs) \cite{93}, Non-local Neural Networks (NLNNs) \cite{102} and GraphSAGE \cite{104}.

\textbf{Variants of GNs------MPNNs.} MPNNs \cite{93} have two phases, a message passing phase and a readout phase. The message passing phase is defined by a message function $M_l$
(playing the role of the composition of the update function $\psi^{e\rightarrow v}(\cdot)$ and the update function $\phi^{e}(\cdot)$)
and a vertex update function $U_l$ (playing the role of the update function $\phi^{v}(\cdot)$). Specifically,
\begin{displaymath}
m_v^{(l+1)}=\displaystyle\sum_{u\in N_G(v)}M_l\left(x_v^{(l)},x_u^{(l)},e_{v,u}\right),\quad x_v^{(l+1)}=U_l\left(x_v^{(l)},m_v^{(l+1)}\right)
\end{displaymath}
where $e_{v,u}$ denotes the feature vector of the edge with two endpoints $v$ and $u$.
The readout phase computes a universal feature vector for the whole graph using a readout function $R(\cdot)$, i.e. $u=R\left(\left\{x_v^{(L)}:v\in V\right\}\right)$.
The readout function $R(\cdot)$ should be invariant to permutations of nodes.
A lot of GCNNs can be regarded as special forms of the MPNN, e.g. \cite{103,122,336,195}.

\textbf{Variants of GNs------NLNNs.} NLNNs \cite{102} give a general definition of non-local operations \cite{381} which is a flexible building block and can be easily integrated into convolutional/recurrent layers.
Specifically, the generic non-local operation is defined as
\begin{equation}\label{NLNN}
y_s=\frac{1}{\mathcal{C}(x_s)}\sum_{t}f(x_s,x_t)g(x_t),
\end{equation}
where $f(\cdot,\cdot)$ denotes the affinity between $x_s$ and $x_t$, and $\mathcal{C}(x_s)=\sum_{t}f(x_s,x_t)$ is a normalization factor.
The affinity function $f(\cdot,\cdot)$ is of the following form
\begin{enumerate}
  \item Gaussian: $f(x_s,x_t)=e^{x_s^Tx_t}$;
  \item Embedded Gaussian: $f(x_s,x_t)=e^{\theta(x_s)^T\eta(x_t)}$, where $\theta(x_s)=W_{\theta}x_s$ and $\eta(x_t)=W_{\eta}x_t$;
  \item Dot Product: $f(x_s,x_t)=\theta(x_s)^T\eta(x_t)$;
  \item Concatenation: $f(x_s,x_t)=\text{ReLU}(w_f^T\left[\theta(x_s),\eta(x_t)\right])$.
\end{enumerate}
The non-local building block is defined as $z_s=W_zy_s+x_s$ where "$+x_s$" denotes a residual connection.
It is noted that $f(\cdot,\cdot)$ and $g(\cdot)$ play the role of $\phi^e(e_k,v_{r_k},v_{s_k},u)$, and the summation in Eq. (\ref{NLNN}) plays the role of $\psi^{e\rightarrow v}(E'_i)$.

\textbf{Variants of GNs------GraphSAGE.} GraphSAGE (\textsc{SAmple} and \textsc{aggreGatE}) \cite{104} is a general inductive framework capitalizing on node feature information to efficiently generate node embedding vectors for previously unseen nodes.
Specifically, GraphSAGE is composed of an aggregation function $\text{\textsc{Aggregate}}^{(l)}(\cdot)$ and an update function $\text{\textsc{Update}}^{(l)}$, i.e.
\begin{displaymath}
\begin{array}{l}
x_{N_G(v)}^{(l)}=\text{\textsc{Aggregate}}^{(l)}\left(\left\{x_u^{(l-1)}: u\in N_G(v)\right\}\right) \\[5mm]
x_v^{(l)}=\text{\textsc{Update}}^{(l)}\left(\left\{x_v^{(l-1)},x_{N_G(v)}^{(l)}\right\}\right)
\end{array}
\end{displaymath}
where $N_G(v)$ denotes a fixed-size set of neighbors of $v$ uniformly sampling from its whole neighbors.
The aggregation function is of the following form
\begin{enumerate}
  \item Mean Aggregator: $x_v^{(l)}=\sigma\left(W\cdot\text{\textsc{Mean}}\left(\left\{x_v^{(l-1)}\right\}\cup\left\{x_u^{(l-1)}:u\in N_G(v)\right\}\right)\right)$;
  \item LSTM Aggregator: applying the LSTM \cite{160} to aggregate the neighbors of $v$;
  \item Pooling Aggregator: $\text{\textsc{Aggregate}}^{(l)}=\max\left(\left\{\sigma\left(Wx_u^{(l)}+b\right): u\in N_G(v)\right\}\right)$.
\end{enumerate}
Note that the aggregation function and update function play the role of $\psi^{e\rightarrow v}(E'_i)$ and $\phi^{v}(v_i,\overline{e}'_i,u)$ in formula (\ref{GN-BLOCK}) respectively.

\textbf{Variants of GNs------Hyperbolic GCNNs.} The Euclidean GCNNs aim to embed nodes in a graph into a Euclidean space. This will incur a large distortion
especially when embedding real-world graphs with scale-free and hierarchical structure. Hyperbolic GCNNs pave an alternative way of embedding with little distortion.
The $n\text{-dimensional}$ hyperbolic space \cite{69,193}, denoted as $\mathbb{H}_K^n$, is a unique, complete, simply connected $d\text{-dimensional}$ Riemannian manifold
with constant negative sectional curvature $-\frac{1}{K}$, i.e.
\begin{displaymath}
\mathbb{H}_K^n=\left\{x\in\mathbb{R}^{n+1}:\langle x,x\rangle_{\mathcal{M}}=-K,x_0>0\right\},
\end{displaymath}
where the Minkowski inner produce $\langle x,y\rangle_{\mathcal{M}}=-x_0y_0+\sum_{j=1}^dx_jy_j,\forall x,y\in\mathbb{R}^{n+1}$.
Its tangent space centered at point $x$ is denoted as $\mathcal{T}_x\mathbb{H}_K^n=\left\{v\in\mathbb{R}^{n+1}:\langle x,v\rangle_{\mathcal{M}}=0\right\}$.
Given $x\in\mathbb{H}_K^n$, let $u\in\mathcal{T}_x\mathbb{H}_K^n$ be unit-speed. The unique unit-speed geodesic $\gamma_{x\rightarrow u}(\cdot)$ such that $\gamma_{x\rightarrow u}(0)=x$
and $\dot{\gamma}_{x\rightarrow u}(0)=u$ is denoted as $\gamma_{x\rightarrow u}(t)=\cosh\left(\frac{t}{\sqrt{K}}\right)x+\sqrt{K}\sinh\left(\frac{t}{\sqrt{K}}\right), u, t>0.$
The intrinsic distance between two points $x,y\in\mathbb{H}_K^n$ is then equal to
\begin{displaymath}
d_{\mathcal{M}}^K\left(x,y\right)=\sqrt{K}\text{arcosh}\left(-\frac{\langle x,y\rangle_{\mathcal{M}}}{K}\right).
\end{displaymath}
Therefore, the above $n\text{-dimensional}$ hyperbolic space with constant negative sectional curvature $-\frac{1}{K}$ is usually denoted as $\left(\mathbb{H}_K^n,d_{\mathcal{M}}^K(\cdot,\cdot)\right)$.
In particular, $\mathbb{H}_1^n$, i.e. $K=1$, is called the hyperboloid model of the hyperbolic space.
Hyperbolic Graph Convolutional Networks (HGCN) \cite{15} benefit from the expressiveness of both GCNNs and hyperbolic embedding.
It employs the exponential and logarithmic maps of the hyperboloid model, respectively denoted as $\exp_x^K(\cdot)$ and $\log_x^K(\cdot)$, to realize the mutual transformation
between Euclidean features and hyperbolic ones. Let $\|v\|_{\mathcal{M}}=\langle v,v\rangle_{\mathcal{M}},v\in\mathcal{T}_x\mathbb{H}_K^n$. The $\exp_x^K(\cdot)$ and $\log_x^K(\cdot)$ are respectively defined to be
\begin{displaymath}
\begin{array}{l}
\exp_x^K(v)=\cosh\left(\frac{\|v\|_{\mathcal{M}}}{\sqrt{K}}\right)x+\sqrt{K}\sinh\left(\frac{\|v\|_{\mathcal{M}}}{\sqrt{K}}\right)\frac{v}{\|v\|_{\mathcal{M}}} \\[2mm]
\log_x^{K}(y)=d_{\mathcal{M}}^K(x,y)\dfrac{y+\frac{1}{K}\langle x,y\rangle_{\mathcal{M}}x}{\left\|y+\frac{1}{K}\langle x,y\rangle_{\mathcal{M}}x\right\|_{\mathcal{M}}},
\end{array}
\end{displaymath}
where $x\in\mathbb{H}_K^n$, $v\in\mathcal{T}_x\mathbb{H}_K^n$ and $y\in\mathbb{H}_K^n$ such that $y\neq 0$ and $y\neq x$.
The HGCN architecture is composed of three components: a Hyperbolic Feature Transform (HFT), an Attention-Based Aggregation (ABA) and a Non-Linear Activation with Different Curvatures (NLADC).
They are respectively defined as
\begin{displaymath}
\begin{array}{lr}
h_j^{H,l}=\left(W^{(l)}\otimes^{K_{l-1}}x_j^{H,l-1}\right)\oplus^{K_{l-1}}b^{(l)}    &\quad \text{(HFT)}, \\[2mm]
y_j^{H,l}=\text{\textsc{Aggregate}}^{K_{l-1}}(h^{H,l})_j                            &\quad \text{(ABA)}, \\[2mm]
x_j^{H,l}=\exp_o^{K_l}\left(\rho\left(\log_o^{K_{l-1}}\left(y_j^{H,l}\right)\right)\right) &\quad \text{(NLADC)},
\end{array}
\end{displaymath}
where $o=\left(\sqrt{K},0,\cdots,0\right)\in\mathbb{H}_K^n$, the subscript $j$ denotes the indices of nodes, the superscript $l$ denotes the layer of the HGCN.
The linear transform in hyperboloid manifold is defined to be $W\otimes^K x^H=\exp_o^K\left(W\log_o^K(x^H)\right)$ and $x^H\oplus^K b=\exp_{x^H}^K\left(P_{o\rightarrow x^H}^K(b)\right)$,
where $P_{o\rightarrow x^H}^K(b)$ is the parallel transport from $\mathcal{T}_o\mathbb{H}_K^n$ to $\mathcal{T}_{x^H}\mathbb{H}_K^{n}$.
The attention-based aggregation is defined to be $\text{\textsc{Aggregate}}^K(x^H)_j=\exp_{x_j^H}^K\left(\sum_{k\in N_G(j)}\omega_{j,k}\log_{x_j^H}^K\left(x_k^H\right)\right)$,
where the attention weight $\omega_{j,k}=\text{Softmax}_{k\in N_G(j)}\left(\text{MLP}\left(\log_o^K\left(x_j^H\right)\bowtie\log_o^K\left(x_k^H\right)\right)\right)$.

\textbf{Higher-Order Spatial GCNNs.} the aforementioned GCNN architectures are constructed from the microscopic perspective.
They only consider nodes and edges, yet overlook the higher-order substructures and their connections, i.e subgraphs consisting of at least 3 nodes.
Here, we introduce the studies on the $k\text{-dimensional}$ GCNNs \cite{88}. Specifically, they take higher-order graph structures at multiple scales into consideration by leveraging the $k\text{-Weisfeiler-Leman}$
($k\text{-WL}$) graph isomorphism test so that the message passing is performed directly between subgraph structures rather than individual nodes. Let $\{\hspace{-1.2mm}\{\cdots\}\hspace{-1.2mm}\}$
denote a multiset, $\text{\textsc{Hash}}(\cdot)$ a hashing function and $C_{l,k}^{(l)}(s)$ the node coloring (label) of $s=(s_1,\cdots,s_k)\in V^k$ at the $l\text{-th}$ time. Moreover, let $N_G^j(s)=\left\{(s_1,\cdots,s_{j-1},r,s_{j+1},\cdots,s_k):r\in V\right\}$.
The $k\text{-WL}$ is computed by
\begin{displaymath}
C_{l,k}^{(l+1)}(s)=\text{\textsc{Hash}}\left(C_{l,k}^{(l)}(s),\left(c_1^{(l+1)}(s),\cdots,c_k^{(l+1)}(s)\right)\right),
\end{displaymath}
where $c_{j}^{(l+1)}=\text{\textsc{Hash}}\left(\left\{\hspace{-1.2mm}\left\{C_{l,k}^{(l)}(s'):s'\in N_G^j(s)\right\}\hspace{-1.2mm}\right\}\right)$.
The $k\text{-GCNN}$ computes new features of $s\in V^k$ by multiple computational layers. Each layer is computed by
\begin{displaymath}
X_k^{(l+1)}[s,:]=\rho\left(X_k^{(l)}[s,:]W_1^{(l)}+\sum_{t\in N_G(s)}X_k^{(l)}[t,:]W_2^{(l)}\right).
\end{displaymath}
In practice, the local $k\text{-GCNNs}$ is often employed to learn the hierarchical representations of nodes in order to scale to larger graphs and mitigate the overfitting problem.

\textbf{Other Variants of GNs.} In addition to the aforementioned GNs and its variants, there are still many other spatial GCNNs which is defined from other perspectives, e.g.
Diffusion-Convolutional Neural Network (DCNN) \cite{20}, Position-aware Graph Neural Network (P-GNN) \cite{46}, Memory-based Graph Neural Network (MemGNN) and Graph Memory Network (GMN) \cite{84},
Graph Partition Neural Network (GPNN) \cite{35}, Edge-Conditioned Convolution (ECC) \cite{43}, DEMO-Net \cite{386}, Column network \cite{29}, Graph-CNN \cite{32}.

\textbf{Invariance and Equivariance.} Permutation-invariance refers to that a function $f:\mathbb{R}^{n^k}\rightarrow\mathbb{R}$ (e.g. the aggregation function)
is independent of any permutations of node/edge indices \cite{50,389}, i.e. $f(P^TA_GP)=f(A_G)$ where $P$ is a permutation matrix and $A_G\in\mathbb{R}^{n^k}$ is
a $k\text{-order}$ tensor of edges or multi-edges in the (hyper-)graph $G$. Permutation-equivariance refers to that a function $f:\mathbb{R}^{n^k}\rightarrow\mathbb{R}^{n^l}$
coincides with permutations of node/edge indices \cite{50,389}, i.e. $f(P^TA_GP)=P^Tf(A_G)P$ where $P$ and $A_G$ are defined as similarly as the permutation-invariance.
For permutation-invariant aggregation functions, a straightforward choice is to take $\text{sum}/\max/\text{average}/\text{concatenation}$ as heuristic aggregation schemes \cite{50}. Nevertheless,
these aggregation functions treat all the neighbors of a vertex equivalently so that they cannot precisely distinguish the structural effects of different neighbors to the target vertex.
That is, the aggregation functions should extract and filter graph signals aggregated from neighbors of different hops away and different importance.
GeniePath \cite{34} proposes a scalable approach for learning adaptive receptive fields of GCNNs. It is composed of two complementary functions,
namely adaptive breadth function and adaptive depth function. The former learns the importance of different sized neighborhoods,
whereas the latter extracts and filters graph signals aggregated from neighbors of different hops away. More specifically, the adaptive breadth function is defined as follows.
\begin{displaymath}
h_{v_j}^{\text{temp}}=\tanh\left((W^{(t)})^T\sum_{v_k\in N_G(v_j)\cup\{v_j\}}\alpha(h_{v_j}^{(t)},h_{v_k}^{(t)})\cdot h_{v_k}^{(t)}\right),
\end{displaymath}
where $\alpha(x,y)=\text{Softmax}_y\left(\alpha^T\tanh\left(W_x^Tx+W_y^Ty\right)\right)$.
The adaptive depth function is defined as a LSTM \cite{160}, i.e.
\begin{equation}\label{LSTM Formulation}
\begin{array}{ll}
i_{v_j}=\sigma\left(\left(W_i^{(t)}\right)^Th_{v_j}^{\text{temp}}\right) & \qquad f_{v_j}=\sigma\left(\left(W_f^{(t)}\right)^Th_{v_j}^{\text{temp}}\right)     \\
o_{v_j}=\sigma\left(\left(W_o^{(t)}\right)^Th_{v_j}^{temp}\right)        & \qquad \widetilde{C}_{v_j}=\tanh\left(\left(W_c^{(t)}\right)^Th_{v_j}^{(temp)}\right) \\
C_{v_j}^{(t+1)}=f_{v_j}\circledast C_{v_j}^{(t)}+i_{v_j}\circledast\widetilde{C}_{v_j} & \qquad h_{v_j}^{(t+1)}=o_{v_j}\circledast\tanh\left(C_{v_j}^{(t+1)}\right).
\end{array}
\end{equation}

\textsc{Geom-GCN} \cite{17} proposes a novel permutation-invariant geometric aggregation scheme consisting of three modules,
namely node embedding, structural neighborhood, and bi-level aggregation. This aggregation scheme does not lose structural information of nodes and fail to capture long-range dependencies in disassortative graphs.
For the permutation-invariant graph representations, PiNet \cite{41} proposes an end-to-end spatial GCNN architecture that utilizes the permutation equivariance of graph convolutions.
It is composed of a pair of double-stacked message passing layers, namely attention-oriented message passing layers and feature-oriented message passing layers.

\textbf{Depth Trap of Spatial GCNNs.} Similar to the spectral GCNNs, the spatial GCNNs is also confronted with the depth trap.
As stated previously, the depth trap results from oversmoothing, overfitting and gradient vanishing/explosion.
In order to escape from the depth trap, some studies propose some available strategies, e.g. DeepGCN \cite{13} and Jumping Knowledge Network \cite{89}.
The jumping knowledge networks \cite{89} adopt neighborhood aggregation with skip connections to integrate information from different layers.
The DeepGCN \cite{13} apply the residual/dense connections \cite{382,383} and dilated aggreagation \cite{384} in the CNNs to construct the spatial GCNN architecture.
They has three instantiations, namely ResGCN, DenseGCN and dilated graph convolution.
ResGCN is inspired by the ResNet \cite{382}, which is defined to be
\begin{displaymath}
\begin{array}{lcl}
G^{(l+1)} & \triangleq & \mathcal{H}(G^{(l)},\Theta^{(l)}) \\ [2mm]
          & = & \mathcal{F}(G^{(l)},\Theta^{(l)})+G^{(l)},
\end{array}
\end{displaymath}
where $\mathcal{F}(\cdot,\cdot)$ can be computed by spectral or spatial GCNNs.
DenseGCN collectively exploit information from different GCNN layers like the DenseNet \cite{383}, which is defined to be
\begin{displaymath}
\begin{array}{lcl}
G^{(l+1)} & \triangleq & \mathcal{H}(G^{(l)},\Theta^{(l)}) \\ [2mm]
          & =          & \text{Concat}\left(\mathcal{F}(G^{(l)},\Theta^{(l)}),G^{(l)}\right) \\ [2mm]
          & =          & \text{Concat}\left(\mathcal{F}(G^{(l)},\Theta^{(l)}),\cdots,\mathcal{F}(G^{(0)},\Theta^{(0)}),G^{(0)}\right).
\end{array}
\end{displaymath}
The dilated aggregation \cite{384} can magnify the receptive field of spatial GCNNs by a dilation rate $d$. More specifically, let $N_G^{(k,d)}(v)$ denote the set of $k$ $d\text{-dilated}$ neighbors of vertex $v$ in $G$.
If $\left(u_1,u_2,\cdots,u_{k\times d}\right)$ are the first sorted $k\times d$ nearest neighbors, then $N_G^{(k,d)}(v)=\left\{u_1,u_{1+d},\cdots,u_{1+(k-1)d}\right\}$.
Thereby, we can construct a new graph $G^{(k,d)}=\left(V^{(k,d)},E^{(k,d)}\right)$ where $V^{(k,d)}=V$ and $E^{(k,d)}=\left\{\left\langle v,u\right\rangle:v\in V,u\in N_G^{(k,d)}\right\}$.
The dilated graph convolution layer can be obtained by running the spatial GCNNs over $G^{(k,d)}$.

%%%%%%%%%%%%%%%%%%%%%%%%%%%%%%%%%%%%%%%%%%%%%%%%%%%%%%%%%%%%%%%%%%%%%%%%%%%%%%%%
\subsubsection{Graph Wavelet Neural Networks}
%%%%%%%%%%%%%%%%%%%%%%%%%%%%%%%%%%%%%%%%%%%%%%%%%%%%%%%%%%%%%%%%%%%%%%%%%%%%%%%%

As stated previously, the spectral and spatial GCNNs are respectively inspired by the graph Fourier transform and message-passing mechanism.
Here, we introduce a new GCNN architecture from the perspective of the Spectral Graph Wavelet Transform (SGWT) \cite{189}.
First of all, the SGWT is determined by a graph wavelet generating kernel $g:\mathbb{R}^+\rightarrow\mathbb{R}^+$ with the property $g(0)=0,g(+\infty)=\lim_{x\rightarrow\infty}g(x)=0$.
A feasible instance of $g(\cdot)$ is parameterized by two integers $\alpha$ and $\beta$, and two positive real numbers $x_1$ and $x_2$ determining the transition regions, i.e.
\begin{displaymath}
g(x;\alpha,\beta,x_1,x_2)=\left\{
\begin{array}{ll}
x_1^{-\alpha}x^{\alpha} & x<x_1 \\
s(x)                    & x_1\leq x\leq x_2 \\
x^{-\beta}x_2^{\beta}   & x>x_2,
\end{array}
\right.
\end{displaymath}
where $s(x)$ is a cubic polynomial whose coefficients can be determined by the continuity constraints $s(x_1)=s(x_2)=1$, $s'(x_1)=\frac{\alpha}{x_1}$ and $s'(x_2)=-\frac{\beta}{x_2}$.
Given the graph wavelet generating kernel $g(\cdot)$ and a scaling parameter $s\in\mathbb{R}^+$, the spectral graph wavelet operator $\Psi_g^s$ is defined to be $\Psi_g^s=Ug(s\Lambda)U^T$
where $g(s\Lambda)=g\left(\text{diag}\left(s\lambda_1,\cdots,s\lambda_N\right)\right)$.
A graph signal $x\in\mathbb{R}^N$ on $G$ can thereby be filtered by the spectral graph wavelet operator, i.e. $\mathcal{W}_{g,s}^x=\Psi_g^sx\in\mathbb{R}^N$.
The literature \cite{190} utilizes a special instance of the graph wavelet operator $\Psi_g^s$ to construct a graph scattering network,
and proves its covariance and approximate invariance to permutations and stability to graph operations.

\textbf{Graph Wavelet Neural Networks.} The above spectral graph wavelet operator $\Psi_g^s$ can be employed to construct a Graph Wavelet Neural Network (GWNN) \cite{191}.
Let $\Psi_g^{-s}\triangleq\left(\Psi_g^s\right)^{-1}$. The graph wavelet based convolution is defined to be
\begin{displaymath}
x*_Gy=\Psi_g^{-s}\left(\Psi_g^sx\circledast\Psi_g^sy\right).
\end{displaymath}
The GWNN is composed of multiple layers of the graph wavelet based convolution. The structure of the $l\text{-th}$ layer is defined as
\begin{equation}\label{Graph Wavelet based Convolution in vector form}
X^{(l+1)}[:,j]=\rho\left(\sum_{k=1}^{d^{(l)}}\Psi_g^{-s}\Theta_{j,k}^{(l)}\Psi_g^sX^{(l)}[:,k]\right),
\end{equation}
where $\Theta_{j,k}^{(l)}$ is a diagonal filter matrix learned in spectral domain.
Eq. (\ref{Graph Wavelet based Convolution in vector form}) can be rewritten as a matrix form, i.e. $X^{(l+1)}=\rho\left(\Psi_g^{-s}\Theta\Psi_g^{s}X^{(l)}W^{(l)}\right)$.
The learnable filter matrix $\Theta$ can be replaced with the $K\text{-localized}$ Chebyshev Polynomial so as to eschew the time-consuming eigendecomposition of $\overline{L}_G$.

\subsubsection{Summary}

The aforementioned GCNN architectures provide available ingredients of constructing the GNNs.
In practice, we can construct our own GCNNs by assembling different modules introduced above.
Additionally, some scholars also study the GCNNs from some novel perspectives,
e.g. the parallel computing framework of the GCNNs \cite{40}, the hierarchical covariant compositional networks \cite{213},
the transfer active learning for GCNNs \cite{45} and quantum walk based subgraph convolutional neural network \cite{51}.
They are closely related to the GCNNs, yet fairly different from the ones introduced above.

%%%%%%%%%%%%%%%%%%%%%%%%%%%%%%%%%%%%%%%%%%%%%%%%%%%%%%%%%%%%%%%%%%%%%%%%%%%%%%%%
\subsection{Graph Pooling Operators}
%%%%%%%%%%%%%%%%%%%%%%%%%%%%%%%%%%%%%%%%%%%%%%%%%%%%%%%%%%%%%%%%%%%%%%%%%%%%%%%%

Graph pooling operators are very important and useful modules of the GCNNs, especially for graph-level tasks such as the graph classification.
There are two kinds of graph pooling operators, namely global graph pooling operators and hierarchical graph pooling operators.
The former aims to obtain the universal representations of input graphs, and the latter aims to capture adequate structural information for node representations.

\subsubsection{Global Graph Pooling Operators.}
Global graph pooling operators pool all of representations of nodes into a universal graph representation.
Many literatures \cite{8,21,41} apply some simple global graph pooling operators, e.g. max/average/concatenate graph pooling, to performing graph-level classification tasks.
Here, we introduce some more sophisticated global graph pooling operators in contrast to the simple ones.
Relational pooling (RP) \cite{98} provides a novel framework for graph representation with maximal representation power.
Specifically, all node embeddings can be aggregated via a learnable function to form a global embedding of $G$.
Let $X^{(v)}\in\mathbb{R}^{N\times d_v}$ and $\underline{X}^{(e)}\in\mathbb{N}^{N\times N\times d_e}$ respectively denote node feature matrix and edge feature tensor.
Tensor $\underline{A}_G\in\mathbb{R}^{N\times N\times (1+d_e)}$ combines the adjacency matrix $A_G$ of $G$ with its edge feature tensor $\underline{X}^{(e)}$,
i.e. $\underline{A}_G[u,v,:]=\mathbb{I}_{(u,v)\in E_G}\bowtie\underline{X}^{(e)}[u,v,:]$.
After performing a permutation on $V_G$, the edge feature tensor $\underline{A}_G^{(\pi,\pi)}\left[\pi(r),\pi(c),d\right]=\underline{A}_G[r,c,d]$ and the node feature matrix $X_{\pi}^{(v)}[\pi(r),c]=X^{(v)}[r,c]$.
The joint RP permutation-invariant function for directed or undirected graphs is defined as
\begin{displaymath}
\bar{\bar{f}}(G)=\frac{1}{N!}\sum_{\pi\in\Pi_{|V|}}\overrightarrow{f}(\underline{A}_G^{\pi,\pi},X_{\pi}^{(v)}),
\end{displaymath}
where $\Pi_{|V|}$ is the set of all distinct permutations on $V_G$ and $\overrightarrow{f}(\cdot,\cdot)$ is an arbitrary (possibly permutation-sensitive) vector-valued function.
Specifically, $\overrightarrow{f}(\cdot,\cdot)$ can be denoted as Multi-Layer Perceptrons (MLPs), Recurrent Neural Networks (RNNs), Convolutional Neural Networks (CNNs) or Graph Neural Networks (GNNs).
The literature \cite{98} proves that $\bar{\bar{f}}(G)$ has the most expressive representation of $G$ under some mild conditions, and provides approximation approaches to making RP computationally tractable.
In addition, there are some other available global graph pooling operators, e.g. SortPooling \cite{92} and function space pooling \cite{97}.

\subsubsection{Hierarchical Graph Pooling Operators.}
Hierarchical graph pooling operators group a set of proximal nodes into a super-node via graph clustering methods.
Consequently, the original graph is coarsened to a new graph with coarse granularity. In practice, the hierarchical graph pooling operators are interleaved with the vanilla GCNN layers.
In general, there are three kinds of approaches to performing the graph coarsening operations, namely invoking the existing graph clustering algorithms (e.g. spectral clustering \cite{7} and Graclus \cite{99}),
learning a soft cluster assignment and selecting the first $k$ top-rank nodes.

\textbf{Invoking existing graph clustering algorithms.}
The graph clustering aims to assign proximal nodes to the same cluster and in-proximal nodes to different clusters.
The coarsened graph regards the resulting clusters as super-nodes and connections between two clusters as super-edges.
The hierarchical graph pooling operators aggregate the representations of nodes in super-nodes via aggregation functions such as max pooling and average pooling \cite{91}
to compute the representations of the super-nodes. The literature \cite{82} proposes the EigenPooling method and presents the relationship between the original and coarsened graph.
In order to construct a coarsened graph of $G$, a graph clustering method is employed to partition $G$ into $K$ disjoint clusters, namely $\left\{G_k:k=1,\cdots,K\right\}$.
Suppose each cluster $G_k$ has $N_k$ nodes, namely $\left\{v_{k,1},\cdots,v_{k,N_k}\right\}$, and its adjacency matrix is denoted as $A_{G_k}$.
The coarsened graph $G_{\text{coar}}$ of $G$ can be constructed by regarding the clusters $G_k,k=1,\cdots,K$ as super-nodes and connections between two super-nodes as edges.
For $G_k$, its sampling matrix $C_k$ of size $\left(N\times N_k\right)$ is defined by
\begin{displaymath}
C_k(s,t)=\left\{
\begin{array}{lll}
1 && \text{if node $v_{k,s}$ in $G_k$ is identical to vertex $v_{t}$ in $G$} \\ [2mm]
0 && \text{otherwise}
\end{array}
\right.
\end{displaymath}
On one hand, $C_k$ can be used to down-sample a $1\text{-dimensional}$ graph signal $x$ on $G$ to obtain an contracted graph signal $x_{G_k}$ on $G_k$, i.e. $x_{G_k}=C_k^Tx$.
On the other hand, $C_k$ can also be used to up-sample a graph signal $x_{G_k}$ on $G_k$ to obtain a dilated graph $G$, i.e. $x=C_kx_{G_k}$.
Furthermore, the adjacency matrix $A_{G_k}$ of $G_k$ can be computed by
\begin{displaymath}
A_{G_k}=C_k^TA_GC_k.
\end{displaymath}
The intra-subgraph adjacency matrix of $G$ is computed by $A_{\text{intra}}=\sum_{k=1}^KC_kA_{G_k}C_k^T$.
Thereby, the inter-subgraph adjacency matrix of $G$ can be computed by $A_{\text{inter}}=A_G-A_{\text{intra}}$.
Let $M_{\text{coar}}\in\mathbb{R}^{N\times K}$ denote the assignment matrix from $G$ to $G_{\text{coar}}$. Its $(j,k)\text{-th}$ entry is defined as
\begin{displaymath}
M_{\text{coar}}[j,k]=\left\{
\begin{array}{lll}
1 && \text{if $v_j$ in $G$ is grouped into $G_k$ in $G_{\text{coar}}$} \\ [2mm]
0 && \text{otherwise} \\
\end{array}
\right.
\end{displaymath}
As a result, the adjacency matrix $A_{\text{coar}}$ of the coarsened graph $G_{\text{coar}}$ is computed by $A_{\text{coar}}=M_{\text{coar}}^TA_{\text{inter}}M_{\text{coar}}$.
In fact, $A_{\text{coar}}$ can be written as $A_{\text{coar}}=f\left(M_{\text{coar}}^TA_GM_{\text{coar}}\right)$ as well,
where $f(\widetilde{a}_{i,j})=1$ if $\widetilde{a}_{i,j}>0$ and $f(\widetilde{a}_{i,j})=0$ otherwise.
As stated previously, $X$ is a $d\text{-dimensional}$ graph signal on $G$.
Then, a $d\text{-dimensional}$ graph signal $X_{\text{coar}}$ on $G_{\text{coar}}$ can be computed by $X_{\text{coar}}=M_{\text{coar}}^TX$.
EigenPooling \cite{82} employs spectral clustering to obtain the coarsened graph, and then up-sample
the Fourier basis of subgraphs $G_k,k=1,\cdots,K$. These Fourier basis are then organized into pooling operators with regard to ascending eigenvalues.
Consequently, the pooled node feature matrix is obtained via concatenating the pooled results.
The literature \cite{42} proposes a novel Hierarchical Graph Convolutional Network (H-GCN) consisting of graph coarsening layers and graph refining layers.
The former employs structural equivalence grouping and structural similarity grouping to construct the coarsened graph,
and the latter restores the original topological structure of the corresponding graph.

\textbf{Learning a soft cluster assignment.}
\textsc{StructPool} \cite{390}, as a structured graph pooling technique, regards the graph pooling as a graph clustering problem so as to learn a
 cluster assignment matrix
via the feature matrix $X$ and adjacency matrix $A_G$. Learning the cluster assignment matrix can formulated as a Conditional Random Field (CRF) \cite{391} based probabilistic inference.
Specifically, the input feature matrix $X$ is treated as global observation, and $Y=\left\{Y_1,\cdots,Y_N\right\}$ is a random field where $Y_i\in\{1,\cdots,K\}$ is a random variable
indicating which clusters the node $v_i$ is assigned to. As a result, $\left(Y,X\right)$ can be characterized by a CRF model, i.e.
\begin{displaymath}
\begin{array}{lll}
\mathbb{P}(Y|X) & = & \dfrac{1}{Z(X)}\exp\left(-\mathcal{E}(Y|X)\right) \\[2mm]
                & = & \dfrac{1}{Z(X)}\exp\left(\displaystyle\sum_{C\in\mathcal{C}_G}\psi_C(Y_C|X)\right)
\end{array}
\end{displaymath}
where $\mathcal{E}(Y|X)=-\sum_{C\in\mathcal{C}_G}\psi_C(Y_C|X)$ is called an energy function, $\mathcal{C}_G$ is a set of cliques, $\psi_C(Y_C|X)$ is a potential function and $Z(X)$ is a partition function.
The energy function $\mathcal{E}(Y|X)$ can be characterized by an unary energy $\psi_u(\cdot)$ and a pairwise energy $\psi_p(\cdot,\cdot)$, i.e.
\begin{displaymath}
\mathcal{E}(Y|X)=\sum_{s=1}^N\psi_u(y_s|X)+\sum_{s\neq t}\psi_p(y_s,y_t|X)a_{s,t}^l,
\end{displaymath}
where $a_{s,t}^l$ denotes the $(s,t)\text{-th}$ entry of the $l\text{-hop}$ adjacency matrix $A_G^l$.
The unary energy matrix $\Psi_u=\left(\psi_u(y_s|X)\right)_{N\times K}$ can be obtained by a GCNN taking the global observation $X$ and the adjacency $A_G$ as input.
The pairwise energy matrix $\Psi_p=\left(\psi_p(y_s,y_t|X)\right)_{K\times K}$ can be obtained by
\begin{displaymath}
\psi_p(y_s,y_t|X)=\mu(y_s,y_t)\frac{x_s^Tx_t}{\sum_{j\neq s}x_s^Ts_j},
\end{displaymath}
where $\mu(y_s,y_t)$ is a learnable compatibility function. Minimizing the energy function $\mathcal{E}(Y|X)$ via mean-field approximation
results in the most probable cluster assignment matrix $M$ for a give graph $G$. As a result, we obtain a new graph $A_{\text{coar}}=f\left(M^TA_GM\right)$ and $X_{\text{coar}}=M^TX$.
\textsc{DiffPool} \cite{94} is a differentiable graph pooling operator
which can generate hierarchical representations of graphs and can be incorporated into various GCNNs in an end-to-end fashion.
It maps an adjacency matrix $A_{G^{(l)}}$ and embedding matrix $Z^{(l)}$ at the $l\text{-th}$ layer
to a new adjacency matrix $A_{G^{(l+1)}}$ and a coarsened feature matrix $X^{(l+1)}$, i.e. $\left(A_{G^{(l+1)}},X^{(l+1)}\right)=\text{\textsc{DiffPool}}\left(A_{G^{(l)}},Z^{(l)}\right)$.
More specifically, $X^{(l+1)}=\left(M^{(l)}\right)^TZ^{(l)}$, $A_{G^{(l+1)}}=\left(M^{(l)}\right)^TA_{G^{(l)}}M^{(l)}$.
Note that the assignment matrix $M^{(l)}$ and embedding matrix $Z^{(l)}$ are respectively computed by two separate GCNNs, namely embedding GCNN and pooling GCNN, i.e.
$Z^{(l)}=\text{GCNN}_{\text{embed}}\left(A_{G^{(l)}},X^{(l)}\right)$, $M^{(l)}=\text{Softmax}\left(\text{GCNN}_{\text{pool}}\left(A_{G^{(l)}},X^{(l)}\right)\right)$.

\textbf{Selecting the first $k$ top-rank nodes.}
The literature \cite{86} proposes a novel Self-Attention Graph Pooling operator (abbreviated as SAGPool).
Specifically, SAGPool firstly employs the GCN \cite{6} to calculate the self-attention scores, and then invokes the top-rank function to select the top $\lceil kN\rceil$ node indices, i.e.
\begin{displaymath}
\begin{array}{lll}
Z=\rho\left(\widetilde{D}_G^{-\frac{1}{2}}\widetilde{A}_G\widetilde{D}_G^{-\frac{1}{2}}X\Theta\right), &  \text{idx}=\text{top-rank}\left(Z,\lceil kN\rceil\right), & Z_{\text{mask}}=Z_{\text{idx}}  \\
X'=X_{\text{idx}}, & X_{\text{out}}=X'\circledast Z_{\text{mask}}, & A_{\text{out}}=A_{\text{idx},\text{idx}}.
\end{array}
\end{displaymath}
As a result, the selected $\text{top-}\lceil kN\rceil$ node indices are employed to extract the output adjacency matrix $A_{\text{out}}$ and feature matrix $X_{\text{out}}$.
In order to exploit the expressive power of an encoder-decoder architecture like U-Net \cite{72}, the literature \cite{95} proposes a novel graph pooling (gPool) layer and a graph unpooling (gUnpool) layer.
The gPool adaptively selects $\text{top-}k$ ranked node indices by the down-sampling technique to form a coarsened graph ($A_{\text{coar}}\in\mathbb{R}^{N\times N}$ and $X_{\text{coar}}\in\mathbb{R}^{N\times d}$)
based on scalar projection values on a learnable projection vector, i.e.
\begin{displaymath}
\begin{array}{lll}
y=\frac{Xp}{\|p\|}, & \text{idx}=\text{top-rank}(y,k), & \widetilde{y}=\tanh(y_{\text{idx}}) \\
\widetilde{X}_{\text{coar}}=X_{\text{idx},:}, & A_{\text{coar}}=A_{\text{idx},\text{idx}}, & X_{\text{coar}}=\widetilde{X}_{\text{coar}}\circledast(\widetilde{y}\mathbf{1}_C^T),
\end{array}
\end{displaymath}
where $y\in\mathbb{R}^{d}$. The gUnpool performs the inverse operation of the gPool layer so as to restore the coarsened graph into its original structure.
To this end, gUnpool records the locations of nodes selected in the corresponding gPool layer, and then restores the selected nodes to their original positions in the graph.
Specifically, let $X_{\text{refine}}=\text{Distribute}(0_{N\times d},X_{\text{coar}},\text{idx})$,
where the function $\text{Distribute}(\cdot,\cdot,\cdot)$ distributes row vectors in $X_{\text{coar}}$ into $0_{N\times d}$ feature matrix according to the indices $\text{idx}$.
Note that row vectors of $X_{\text{refine}}$ with indices in $idx$  are updated by the ones in $X_{\text{coar}}$, whereas other row vectors remain zero.
It is worth noting that the literature \cite{96} adopts the similar pooling strategy as gPool to learn the hierarchical representations of nodes.

%%%%%%%%%%%%%%%%%%%%%%%%%%%%%%%%%%%%%%%%%%%%%%%%%%%%%%%%%%%%%%%%%%%%%%%%%%%%%%%%
\subsection{Graph Attention Mechanisms}
%%%%%%%%%%%%%%%%%%%%%%%%%%%%%%%%%%%%%%%%%%%%%%%%%%%%%%%%%%%%%%%%%%%%%%%%%%%%%%%%

Attention mechanisms, firstly introduced in the deep learning community, guide deep learning models to focus on the task-relevant part of its inputs so as to make precise predictions or inferences \cite{73,306,393}.
Recently, applying the attention mechanisms to GCNNs has gained considerable attentions so that various attention techniques have been proposed.
Below, we summarize the graph attention mechanisms on graphs from the next 4 perspectives \cite{394}, namely softmax-based graph attention, similarity-based graph attention,
spectral graph attention and attention-guided walk. Without loss of generality, the neighbors of a given node $v_0$ in $G$ are denoted as $v_1,\cdots,v_{d_0}$,
and their current feature vectors are respectively denoted as $x_0,x_1,\cdots,x_{d_0}$, where $d_0=d_{G}(v_0)$.

\begin{figure}[htb]
  \centering
  \includegraphics[width=0.8\textwidth]{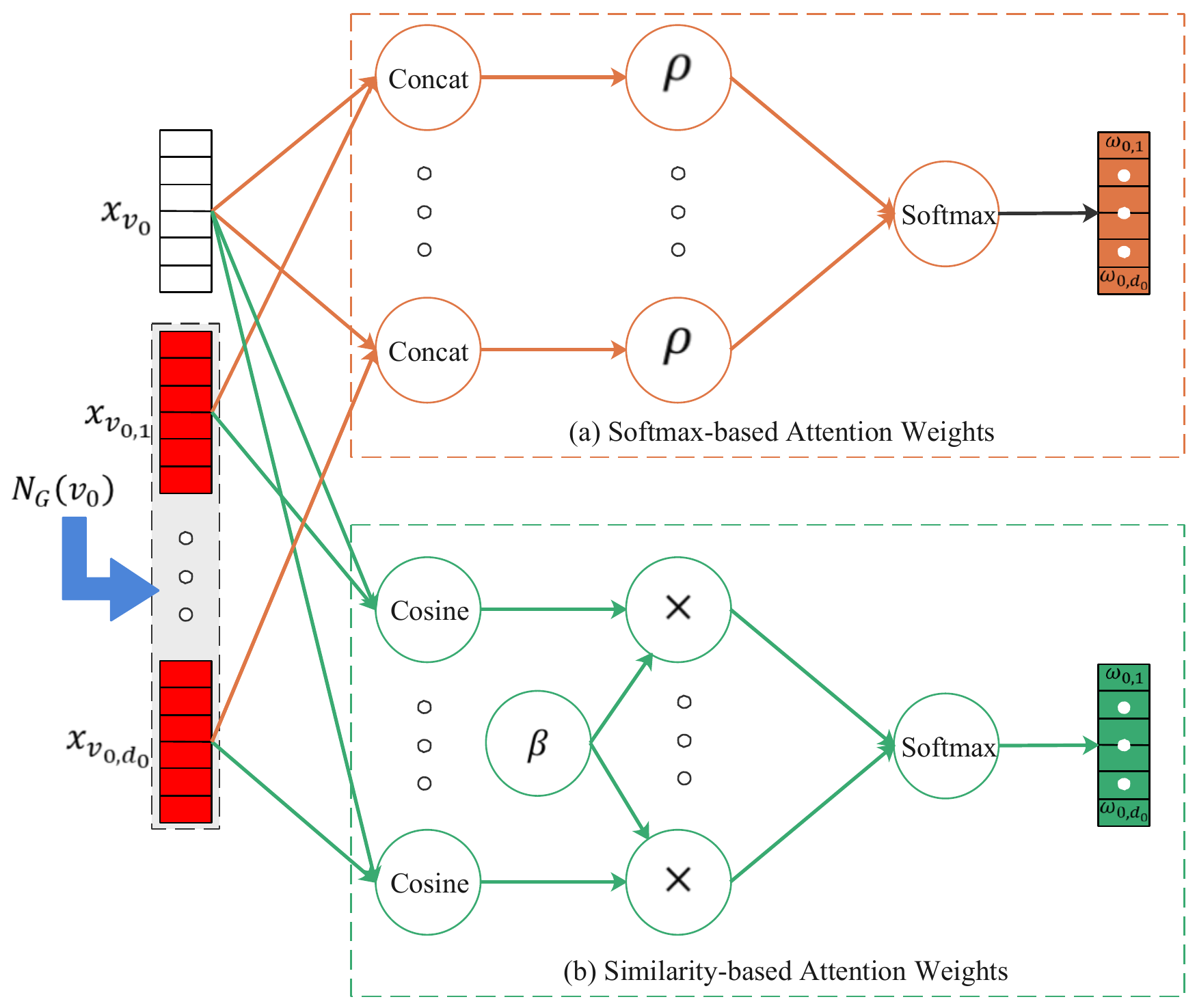}
  \caption{Two kinds of graph attention mechanisms.}
  \label{GAT}
\end{figure}

\subsubsection{Concatenation-based Graph Attention.}
The softmax-based graph attention is typically implemented by employing a softmax with learnable weights \cite{66,110}
to measure the relevance of $v_j,j=1,\cdots,d_G(v_0)$ to $v_0$. More specifically, the softmax-based attention weights between $v_0$ and $v_j$ can be defined as
\begin{equation}\label{Softmax-based Attention}
\begin{array}{lll}
\omega_{0,j} & = & \text{Softmax}\left(\left[e_{0,1},\cdots,e_{0,d_G(v_0)}\right]\right) \\[2mm]
             & = & \dfrac{\exp\left(\rho\left(a^T(Wx_0\bowtie Wx_j)\right)\right)}{\sum_{k=1}^{d_G(v_0)}\exp\left(\rho\left(a^T(Wx_0\bowtie W_k)\right)\right)},
\end{array}
\end{equation}
where $e_{0,j}=\exp\left(\rho\left(a^T(Wx_0\bowtie Wx_j)\right)\right)$, $a$ is a learnable attention vector and $W$ is a learnable weight matrix, see Fig. \ref{GAT}(a).
As a result, the new feature vector of $v_0$ can be updated by
\begin{equation} \label{Single_Graph_Attention}
x'_{0}=\rho\left(\sum_{j=1}^{d_G(v_0)}\omega_{0,j}Wx_j\right).
\end{equation}
In practice, multi-head attention mechanisms are usually employed to stabilize the learning process of the single-head attention \cite{66}.
For the multi-head attention, assume that the feature vector of each head is $x_0^{(h)}=\rho\left(\sum_{j=1}^{d_G(v_0)}\omega_{0,j}^{(h)}W^{(h)}x_j\right)$.
The concatenation based multi-head attention is computed by
$x'_0=\bowtie_{h=1}^H x_0^{(h)}=\bowtie_{h=1}^H \rho\left(\sum_{j=1}^{d_G(v_0)}\omega_{0,j}^{(h)}W^{(h)}x_j\right)$.
The average based multi-head attention is computed by $x'_0=\rho\left(\frac{1}{H}\sum_{h=1}^H\sum_{j=1}^{d_G(v_0)}\omega_{0,j}^{(h)}W^{(h)}x_j\right)$.

The conventional multi-head attention mechanism treats all the attention heads equally so that feeding the output of an attention that captures a useless representation maybe mislead the final prediction of the model.
The literature \cite{74} computes an additional soft gate to assign different weights to heads, and gets the formulation of the gated multi-head attention mechanism.
The Graph Transformer (GTR) \cite{110} can capture long-range dependencies of dynamic graphs with softmax-based attention mechanism by propagating features within the same graph structure via an intra-graph message passing.
The Graph-BERT \cite{367} is essentially a pre-training method only based on the graph attention mechanism without any graph convolution or aggregation operators.
Its key component is called a graph transformer based encoder, i.e. $X^{(l+1)}=\text{Softmax}\left(\frac{QK^T}{\sqrt{d_h}}\right)V$,
where $Q=X^{(l)}W_Q^{(l+1)}$, $K=X^{(l)}W_K^{(l+1)}$ and $V=X^{(l)}W_V^{(l+1)}$.
The Graph2Seq \cite{106} is a general end-to-end graph-to-sequence neural encoder-decoder model converting an input graph to a sequence of vectors with the attention based LSTM model.
It is composed of a graph encoder, a sequence decoder and a node attention mechanism. The sequence decoder takes outputs (node and graph representations) of the graph encoder as input, and employs the softmax-based
attention to compute the context vector sequence.

\subsubsection{Similarity-based Graph Attention.}
The similarity-based graph attention depends on the cosine similarities of the given node $v_0$ and its neighbors $v_j,j=1,\cdots,d_G(v_0)$.
More specifically, the similarity-based attention weights are computed by
\begin{equation}\label{Similarity-based Attention}
\omega_{0,j}=\dfrac{\exp\left(\beta\cdot\cos\left(Wx_0,Wx_j\right)\right)}{\sum_{k=1}^{d_G(v_0)}\exp\left(\beta\cdot\cos\left(Wx_0,Wx_k\right)\right)},
\end{equation}
where $\beta$ is learnable bias and $W$ is a learnable weight matrix, see Fig. \ref{GAT}(b). It is well known that $\cos\left(x,y\right)=\frac{\langle x,y\rangle}{\|x\|_2\|y\|_2}$.
Attention-based Graph Neural Network (AGNN) \cite{67} adopts the similarity-based attention to construct the propagation matrix $P^{(l)}$ capturing the relevance of $v_j$ to $v_i$.
As a result, the output hidden representation $X^{(l+1)}$ at the $(l+1)\text{-th}$ layer is computed by
\begin{displaymath}
X^{(l+1)}=\rho\left(P^{(l)}X^{(l)}W^{(l)}\right),
\end{displaymath}
where $P^{(l)}(i,j)\triangleq\omega_{i,j}$ is defined in formula (\ref{Similarity-based Attention}).

\subsubsection{Spectral Graph Attention.}
The Spectral Graph Attention Networks (SpGAT) aims to learn representations for different frequency components \cite{362}.
The eigenvalues of the normalized graph Laplacian $\overline{L}_G$ can be treated as frequencies on the graph $G$. As stated in the Preliminary section, $0=\lambda_1\leq\lambda_2\leq\cdots\leq\lambda_N=\lambda_{\max}$.
The SpGAT firstly extracts the low-frequency component $B_L=\left\{u_1,\cdots,u_{n}\right\}$ and the high-frequency component $B_H=\left\{u_{N-n+1},\cdots,u_N\right\}$
from the graph Fourier bases $\left\{u_1,u_2,\cdots,u_N\right\}$.
So, we have
\begin{equation}\label{Spectral Graph Attention Networks}
\begin{array}{l}
X_L=X^{(l)}\Theta_L,\quad X_H=X^{(l)}\Theta_H \\
X^{(l+1)}=\rho\left(\text{\textsc{Aggregate}}\left(B_LF_LB_L^TX_L,B_HF_HB_H^TX_H\right)\right),
\end{array}
\end{equation}
where $F_L$ and $F_H$ respectively measure the importance of the low- and high-frequency.
In practice, we exploit a re-parameterization trick to accelerate the training. More specifically, we replace $F_L$ and $F_H$ respectively with the learnable attention weights
$\Omega_L=\text{diag}\left(\omega_L,\cdots,\omega_L\right)$ and $\Omega_H=\text{diag}\left(\omega_H,\cdots,\omega_H\right)$ so as to reduce the number of learnable parameters.
To ensure that  $\omega_L$ and $\omega_H$ are positive and comparable, we normalize them by the softmax function, i.e.
$\omega_L=\frac{\exp\left(\omega_L\right)}{\exp\left(\omega_L\right)+\exp\left(\omega_H\right)},\quad \omega_H=\frac{\exp\left(\omega_H\right)}{\exp\left(\omega_L\right)+\exp\left(\omega_H\right)}$.
In addition to the attention weights, another important issue is how to choose the low- and high-frequency components $B_L$ and $B_H$.
A natural choice is to use the graph Fourier bases, yet the literatures \cite{191,194} conclude that utilizing the spectral graph wavelet operators can achieve better embedding results than the graph Fourier bases.
Therefore, we substitute $B_L$ and $B_H$ in Eq. (\ref{Spectral Graph Attention Networks}) for the spectral graph wavelet operator $\Psi_{L,g}^s$ and $\Psi_{H,g}^s$, i.e.
\begin{displaymath}
X^{(l+1)}=\rho\left(\text{\textsc{Aggregate}}\left(\left(\Psi_{L,g}^s\right)F_L\left(\Psi_{L,g}^s\right)^{-1}X_L,\left(\Psi_{H,g}^s\right)F_H\left(\Psi_{H,g}^s\right)^{-1}X_H\right)\right).
\end{displaymath}

\subsubsection{Attention-guided Walk.}
\begin{figure}[htb]
  \centering
  \includegraphics[width=0.8\textwidth]{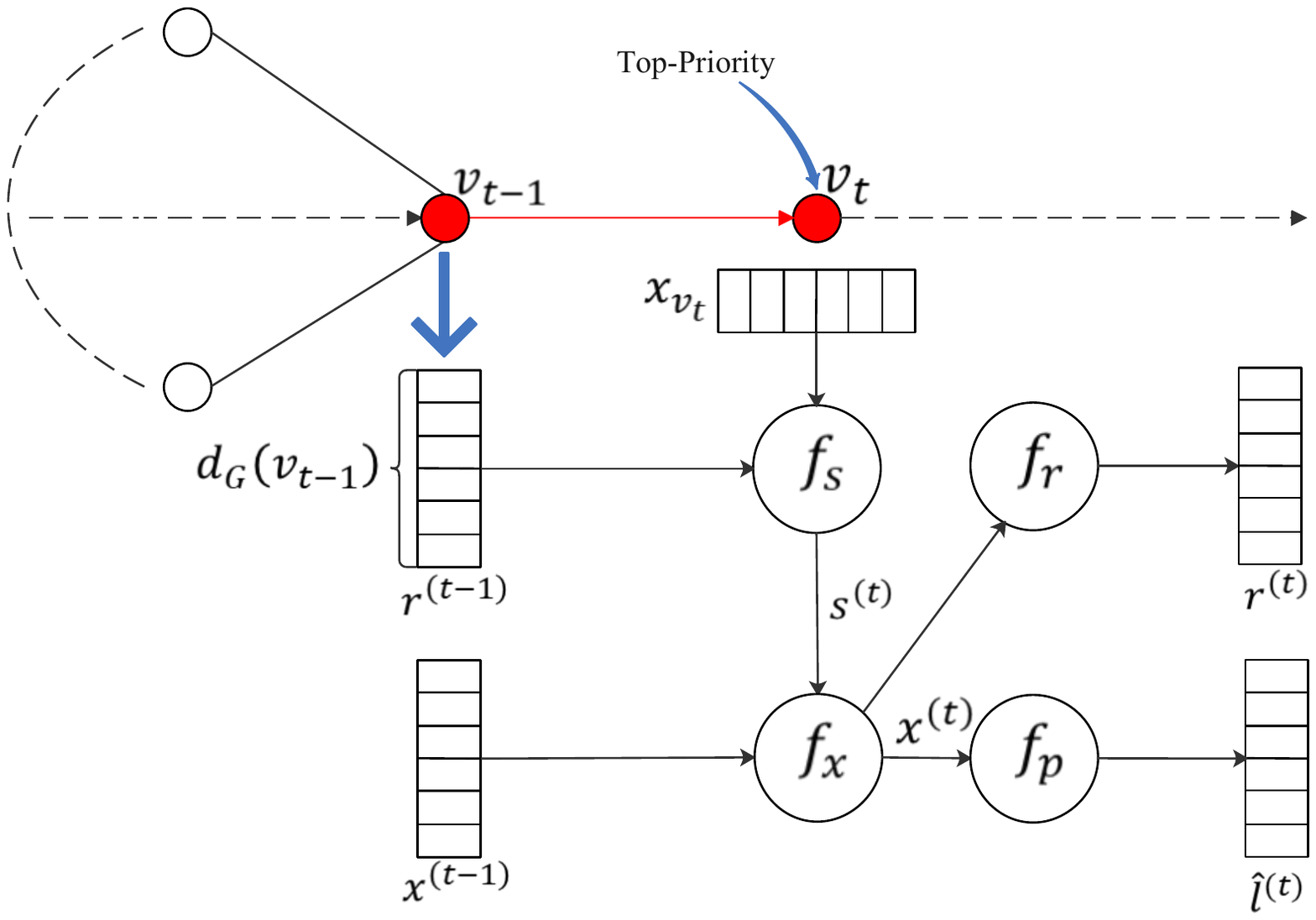}
  \caption{The computational framework of the attention-guided walk.}
  \label{attention-guided-walk}
\end{figure}

The two aforementioned kinds of attention mechanisms focus on incorporating task-relevant information from the neighbors of a given node into the updated representations of the pivot.
Here, we introduce a new attention mechanism, namely attention-guided walk \cite{81}, which has different purpose from the softmax- and similarity-based attention mechanisms.
Suppose a walker walks along the edges of the graph $G$ and he currently locates at the node $v_t$.
The hidden representation $x^{(t)}$ of $v_t$ is computed by a recurrent neural network $f_x\left(\cdot\right)$
taking the step embedding $s^{(t)}$ and internal representation of the historical information from the previous step $x^{(t-1)}$ as input, i.e.
\begin{displaymath}
x^{(t)}=f_x\left(s^{(t)},x^{(t-1)};\Theta_x\right).
\end{displaymath}
The step embedding $s^{(t)}$ is computed by a step network $f_s\left(r^{(t-1)},x_{v_t};\Theta_s\right)$ taking the ranking vector $r^{(t-1)}$
and the input feature vector $x_{v_t}$ of the top-priority node $v_t$ as input, i.e.
\begin{displaymath}
s^{(t)}=f_s\left(r^{(t-1)},x_{v_t};\Theta_s\right).
\end{displaymath}
The hidden representation $x^{(t)}$ is then feeded into a ranking network $f_r\left(x^{(t)};\Theta_r\right)$ and a predicting network $f_p\left(x^{(t)};\Theta_p\right)$, i.e.
\begin{displaymath}
r^{(t)}=f_r\left(x^{(t)};\Theta_r\right), \quad\hat{l}^{(t)}=f_p\left(x^{(t)};\Theta_p\right).
\end{displaymath}
The ranking network $f_r\left(x^{(t)};\Theta_r\right)$ determines which neighbors of $v_t$ should be prioritized in the next step, and the predicting network $f_p\left(x^{(t)};\Theta_p\right)$
makes a prediction on graph labels. Now, $x^{(t)}$ and $r^{(t)}$ are feeded into the next node to compute its hidden representations.
Fig. \ref{attention-guided-walk} shows the computational framework of the attention-guided walk.

%%%%%%%%%%%%%%%%%%%%%%%%%%%%%%%%%%%%%%%%%%%%%%%%%%%%%%%%%%%%%%%%%%%%%%%%%%%%%%%%
\subsection{Graph Recurrent Neural Networks}
%%%%%%%%%%%%%%%%%%%%%%%%%%%%%%%%%%%%%%%%%%%%%%%%%%%%%%%%%%%%%%%%%%%%%%%%%%%%%%%%

The Graph Recurrent Neural Networks (GRNNs) generalize the Recurrent Neural Networks (RNNs) to process the graph-structured data.
In general, the GRNN can be formulated as $h_{v_j}^{(l+1)}=\text{GRNN}\left(x_{v_j}^{(l)},\left\{h_{v_k}^{(l)}:v_k\in N_G(v_j)\cup\left\{v_j\right\}\right\}\right)$.
Below, we introduce some available GRNN architectures.

\subsubsection{Graph LSTM}
\begin{figure}[htb]
  \centering
  \includegraphics[width=0.8\textwidth]{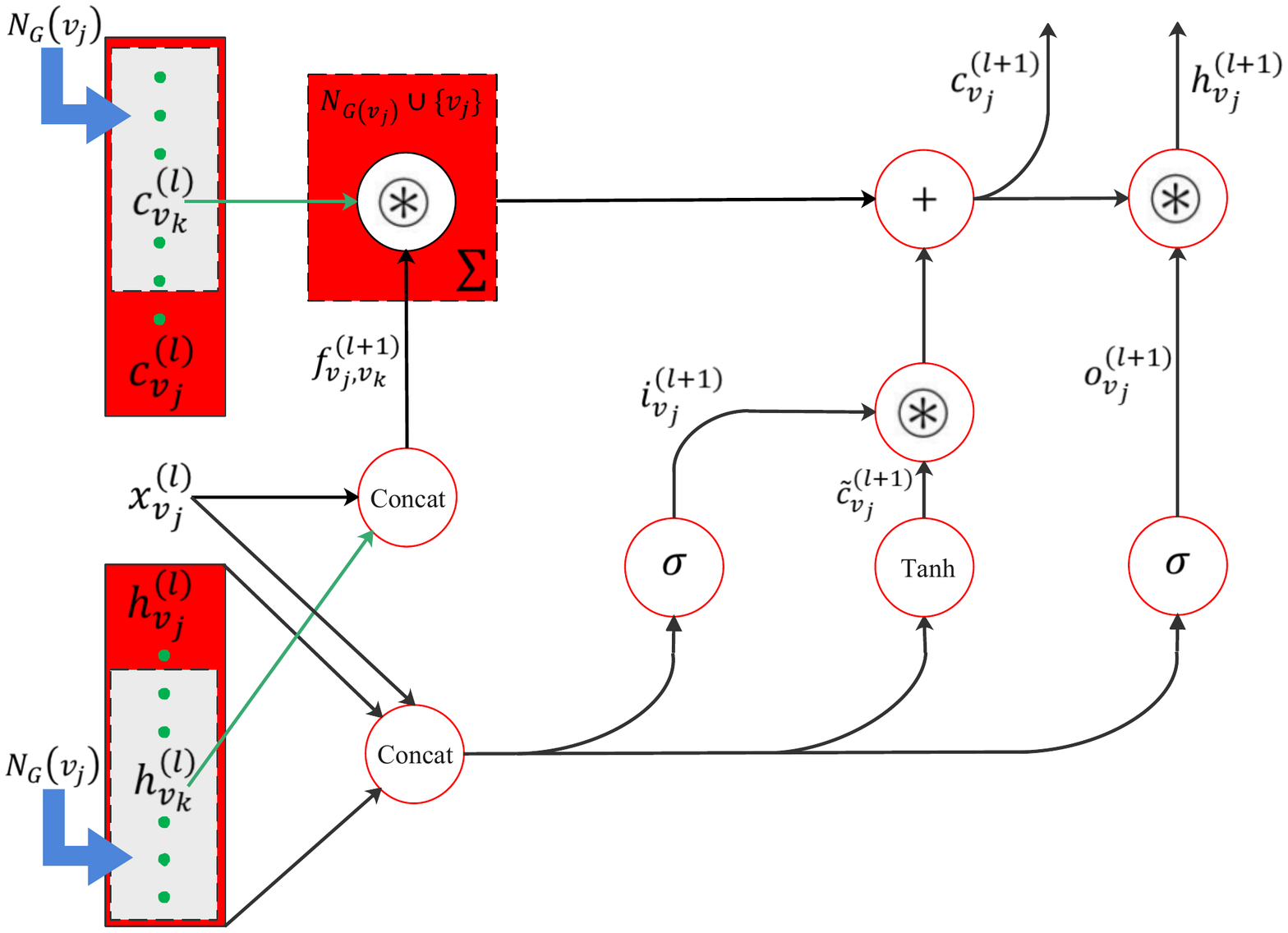}
  \caption{Computational framework of the Graph LSTM.}
  \label{Graph LSTM}
\end{figure}

The Graph Long Short Term Memroy (Graph LSTM) \cite{117,118,119,120,121,123,143} generalizes the vanilla LSTM for the sequential data to the ones for general graph-structured data.
Specifically, the graph LSTM updates the hidden states and cell states of nodes by the following formula,
\begin{displaymath}
\begin{array}{ll}
i_{v_j}^{(l+1)}=\sigma\left(W_ix_{v_j}^{(l)}+\sum_{v_k\in N_G(v_j)\cup\left\{v_j\right\}}U_ih_{v_k}^{(l)}+b_i\right), \\ [2mm]
o_{v_j}^{(l+1)}=\sigma\left(W_ox_{v_j}^{(l)}+\sum_{v_k\in N_G(v_j)\cup\left\{v_j\right\}}U_oh_{v_k}^{(l)}+b_o\right) \\ [2mm]
\widetilde{c}_{v_j}^{(l+1)}=\tanh\left(W_cx_{v_j}^{(l)}+\sum_{v_k\in N_G(v_j)\cup\left\{v_j\right\}}U_ch_{v_k}^{(l)}+b_c\right), \\ [2mm]
f_{v_j,v_k}^{(l+1)}=\sigma\left(W_fx_{v_j}^{(l)}+U_fh_{v_k}^{(l)}+b_f\right) \\ [2mm]
c_{v_j}^{(l+1)}=i_{v_j}^{(l+1)}\circledast\widetilde{c}_{v_j}^{(l+1)}+\sum_{v_k\in N_G{v_j}\cup\left\{v_j\right\}}f_{v_j,v_k}^{(l+1)}\circledast c_{v_k}^{(l)}, \\ [2mm] h_{v_j}^{(l+1)}=o_{v_j}^{(l+1)}\circledast\tanh(c_{v_j}^{(l+1)}).
\end{array}
\end{displaymath}
see the Fig. \ref{Graph LSTM}.
The literature \cite{267} develops a general framework, named structure-evolving LSTM, for learning interpretable data representations via the graph LSTM.
It progressively evolves the multi-level node representations by stochastically merging two adjacent nodes with high compatibilities estimated by the adaptive forget gate of the graph LSTM.
As a result, the new graph is produced with a Metropolis-Hastings sampling method.
The Gated Graph Sequence Neural Networks (GGS-NNs) \cite{122} employs the Gated Recurrent Unit (GRU) \cite{115} to modify the vanilla GCNNs so that it can be extended to process the sequential data.

\subsubsection{GRNNs for Dynamic Graphs}
A dynamic graph is the one whose structure (i.e. adding a node, removing a node, adding an edge, removing an edge) or features on nodes/edges evolve over time.
The GRNNs are a straightforward approach to tackling dynamic graphs. Below, we introduce some studies on the GRNNs for dynamic graphs.

The literature \cite{107} proposes a Dynamic Graph Neural Network (DGNN) concerning the dynamic graph only with  nodes or edges adding.
More specifically, the DGNN is composed of two key components: an update component and a propagation component. Suppose an edge $(v_s,v_g,t)$ is added to the input dynamic graph at time $t$.
Let $t-$ denotes a time before the time $t$. The update component consists of three sequential units: the interacting unit, the S- or G-update unit and the merging unit.
The interacting unit takes the source and target representations before the time $t$ as input, and outputs the joint representation of the interaction, i.e.
$x_e^{(t)}=\rho\left(W_sx_s^{(t-)}+W_gx_g^{(t-)}+b_e\right)$. The S- and G-update units employ the LSTM \cite{160} to respectively update the cell states and hidden states of
the source and target, i.e.
\begin{displaymath}
\begin{array}{l}
\left(C_{v_s}^{(t)},h_{v_s}^{(t)}\right)=\text{LSTM}_s\left(C_{v_s}^{(t-)},h_{v_s}^{(t-)},\Delta t_s\right),\quad \left(C_{v_g}^{(t)},h_{v_g}^{(t)}\right)=\text{LSTM}_g\left(C_{v_g}^{(t-)},h_{v_g}^{(t-)},\Delta t_g\right).
\end{array}
\end{displaymath}
The merging unit adopts the similar functions to the interacting unit to respectively merge $h_{v_s}^{(t)}$ and $h_{v_s}^{t-}$, and $h_{v_g}^{(t)}$ and $h_{v_g}^{t-}$.
The propagation component can propagate information from two interacting nodes ($v_s$ and $v_g$) to influenced nodes (i.e. their neighbors). It also consists of three units: the interacting unit,
the propagation unit and the merge unit, which are defined similarly to the update component except that they have different learnable parameters.
The literature \cite{105} addresses the vertex- and graph-focused prediction tasks on dynamic graphs with a fixed node set by combining GCNs, LSTMs and fully connected layers.
The Variational Graph Recurrent Neural Networks (VGRNNs) \cite{113} is essentially a variational graph auto-encoder whose encoder integrates the GCN and RNN into a graph RNN (GRNN) framework
and decoder is a joint probability distribution of a multi-variate Gaussian distribution and Bernoulli Distribution. The semi-implicit variational inference is employed to approximate the posterior
so as to generate the node embedding.

\subsubsection{GRNNs based on Vanilla RNNs}
The GRNNs based on vanilla RNNs firstly employ random walk techniques or traversal methods, e.g. Breadth-First Search (BFS) and Depth-First Search (DFS),
to obtain a collection of node sequences, and then leverage a RNN, e.g. LSTM and GRU, to capture long short-term dependencies.
The literature \cite{116} performs joint random walks on attributed networks, and utilizes them to boost the deep node representation learning.
The proposed GraphRNA in \cite{116} consists of two key components, namely a collaborative walking mechanism AttriWalk and a tailored deep embedding architecture for joint random walks GRN.
Suppose $\mathcal{A}_G$ denotes the node-attribute matrix of size $N\times M$. The AtriWalk admits the transition matrix of size $\mathbb{R}_+^{(N+M)\times(N+M)}$ which is written as
\begin{displaymath}
\mathcal{T}=
\left[
\begin{array}{cc}
\alpha A_G & (1-\alpha)\mathcal{A}_G \\
(1-\alpha)\mathcal{A}_G^T & 0
\end{array}
\right].
\end{displaymath}
After obtaining the sequences via the collaborative random walk, the bi-directional GRU \cite{115} and pooling operator are employed to learn the global representations of sequences.
The literature \cite{111} leverages the BFS node ordering and truncation strategy to obtain a collection of node representation sequences,
and then uses the GRU model and variational auto-regression regularization to perform the graph classification.

%%%%%%%%%%%%%%%%%%%%%%%%%%%%%%%%%%%%%%%%%%%%%%%%%%%%%%%%%%%%%%%%%%%%%%%%%%%%%%%%
\section{Extensions and Applications}
%%%%%%%%%%%%%%%%%%%%%%%%%%%%%%%%%%%%%%%%%%%%%%%%%%%%%%%%%%%%%%%%%%%%%%%%%%%%%%%%

The aforementioned architectures essentially provide ingredients of constructing the GNNs for us. Below, we investigate the extensions of the GNNs from the next 8 aspects:
GCNNs on spectral graphs, capability and interpretability, deep graph representation learning, deep graph generative models, combinations of the PI and GNNs,
adversarial attacks for the GNNs, graph neural architecture search and graph reinforcement learning, and briefly summarize the applications of the GNNs at last.

\subsection{GCNNs on Special Graphs}

The vanilla GCNNs aims at learning the representations of input graphs (directed or undirected, weighted or unweighted).
The real-world graphs may have more additional characteristics, e.g. spatial-temporal graphs, heterogeneous graphs, hyper-graphs, signed graphs and so all.
The GCNN for signed graphs \cite{28} leverage the balance theory to aggregate and propagate information through positive and negative links.

\subsubsection{Heterogeneous Graphs.}
Heterogeneous Graphs are composed of nodes and edges of different types, and each type of edges is called a relation between two types of nodes.
For example, a bibliographic information network contains at least 4 types of nodes, namely Author, Paper, Venue and Term, and at least 3 types of edges, namely Author-Paper, Term-Paper and Venue-Paper \cite{395}.
The heterogeneity and rich semantic information brings great challenges for designing heterogeneous graph convolutional neural networks. In general, a heterogeneous graph can be denoted as $H=(V,E,\nu,\zeta)$,
where $\nu(v)$ denotes the type of node $v\in V$ and $\zeta(e)$ denotes the type of edge $e\in E$. Let $\mathcal{T}^v$ and $\mathcal{T}^e$ respectively denote the set of node types and edge types.
Below, we summarize the vanilla heterogeneous GCNNs, namely HetGNNs \cite{61}.

\textbf{Vanilla Heterogeneous GCNNs.}
The Heterogeneous Graph Neural Networks (HetGNNs) \cite{61} aims to resolve the issue of jointly considering heterogeneous structural information
as well as heterogeneous content information of nodes. It firstly samples a fixed size of strongly correlated heterogeneous neighbors for each node via a Random Walk with Restart (RWR)
and groups them into different node types. Then, it aggregates feature information of those sampled neighboring nodes via a bi-directional Long Short Term Memory (LSTM) and attention mechanism.
Running RWR with a restart probability $p$ from node $v$ will yield a collection of a fixed number of nodes, denoted as $\text{RWR}(v)$. For each node type $t$,
the $t\text{-type}$ neighbors $N_G^t(v)$ of node $v$ denotes the set of $\text{top-}k_t$ nodes from $\text{RWR}(v)$ with regard to frequency.
Let $\mathcal{C}_v$ denote the heterogeneous contents of node $v$, which can be encoded as a fixed size embedding via a function $f_1(v)$, i.e.
\begin{displaymath}
f_1(v)=\dfrac{\sum_{j\in\mathcal{C}_v}\left[\overrightarrow{\text{LSTM}}\left(\mathcal{FC}_{\theta_x}(x_j)\right)\bowtie\overleftarrow{\text{LSTM}}\left(\mathcal{FC}_{\theta_x}(x_j)\right)\right]}{|\mathcal{C}_v|},
\end{displaymath}
where $\mathcal{FC}_{\theta_x}(\cdot)$ denotes feature transformer, e.g. identity or fully connected neural networks with parameter $\theta_x$, and
$\overrightarrow{\text{LSTM}}\left(\cdot\right)$ and $\overleftarrow{\text{LSTM}}\left(.\right)$ is defined by the Eq. (\ref{LSTM Formulation}).
The content embedding of the $t\text{-type}$ neighbors of node $v$ can be aggregated as follows,
\begin{displaymath}
\begin{array}{lcl}
f_2^t(v) & = & \text{\textsc{Aggregate}}^T\left(\left\{f_1(v'):v'\in N_G^t(v)\right\}\right) \\[2mm]
         & = & \dfrac{\sum_{v'\in N_G^t(v)}\left[\overrightarrow{\text{LSTM}}(f_1(v'))\bowtie\overleftarrow{\text{LSTM}}(f_1(v'))\right]}{|N_G^t(v)|}.
\end{array}
\end{displaymath}
Let $\mathcal{F}(v)=\{f_1(v)\}\cup\{f_2^t(v):t\in\mathcal{T}^v\}$.
As a result, the output embedding of node $v$ can be obtained via the attention mechanism, i.e. $\mathcal{E}_v=\sum_{f_j(v)\in\mathcal{F}(v)}\omega_{v,j}f_j(v)$,
where the attention weights is computed by
$\omega_{v,j}=\dfrac{\exp\left(\rho\left(u^t\left[f_j(v)\bowtie f_1(v)\right]\right)\right)}{\sum_{f_j(v)\in\mathcal{F}(v)}\exp\left(\rho\left(u^t\left[f_j(v)\bowtie f_1(v)\right]\right)\right)}$.
In addition, the GraphInception \cite{397} can be employed to learn the hierarchical relational features on heterogeneous graphs
by converting the input graph into a multi-channel graph (each meta path as a channel) \cite{62}.

\textbf{Heterogeneous Graph Attention Mechanism.} The literature \cite{75} firstly proposes a hierarchical attention based
heterogeneous GCNNs consisting of node-level and semantic-level attentions. The node-level attention aims to learn the attention weights of a node and its meta-path-based neighbors,
and the semantic-level attention aims to learn the importance of different meta-paths. More specifically, given a meta path $\Phi$, the node-level attention weight of a node $v_i$
and its meta-path-based neighbors $v_j\in N_G^{\Phi}(v_i)$ is defined to be
\begin{displaymath}
\omega_{j,k}^{\Phi}=\dfrac{\exp\left(\rho\left(a_{\Phi}^T(M_{\nu(v_j)}x_j\bowtie M_{\nu(v_k)}x_k)\right)\right)}{\sum_{v_k\in N_G^{\Phi}(v_j)}\exp\left(\rho\left(a_{\Phi}^T(M_{\nu(v_j)}x_j\bowtie M_{\nu(v_k)}x_k)\right)\right)},
\end{displaymath}
where $M_{\nu(v_j)}$ transforms the feature vectors of nodes of type $\nu(v_j)$ in different vector spaces into a unified vector space.
The embedding of node $v_i$ under the meta path $\Phi$ can be computed by
\begin{displaymath}
x_j^{\Phi,l+1}=\bowtie_{k=1}^K\rho\left(\sum_{k\in N_G^{\Phi}(v_j)}\omega_{j,k}^{\Phi}M_{\nu(v_k)}x_k^{\Phi,l}\right).
\end{displaymath}
Given a meta-path set $\left\{\Phi_0,\cdots,\Phi_P\right\}$, performing the node-level attention layers under each meta path will yield a set of semantic-specific node representations,
namely $\left\{X^{\Phi_0},\cdots,X^{\Phi_P}\right\}$. The semantic-level attention weight of the meta path $\Phi_j$ is defined as
\begin{displaymath}
\beta^{\Phi_j}=\dfrac{\exp\left(\omega^{\Phi_j}\right)}{\sum_{p=1}^P\exp\left(\omega^{\Phi_p}\right)},
\end{displaymath}
where $\omega^{\Phi_p}=\frac{1}{|V|}\sum_{v_k\in V}q^T\tanh\left(Wx_k^{\Phi_p}+b\right)$. As a result, the embedding matrix $X=\sum_{p=1}^P\beta^{\Phi_p}X^{\Phi_p}$.
In addition, there are some available studies on the GCNNs for multi-relational graphs \cite{36,68,361} and the transformer for dynamic heterogeneous graphs \cite{60,114}.

\subsubsection{Spatio-Temporal Graphs.}
Spatio-temporal graphs can be used to model traffic networks \cite{59,71} and skeleton networks \cite{55,57}.
In general, a spatio-temporal graph is denoted as $G_{ST}=(V_{ST},E_{ST})$ where $V_{ST}=\{v_{t,j}:t=1,\cdots,T,j=1,\cdots,N_{ST}\}$.
The edge set $E_{ST}$ is composed of two types of edges, namely spatial edges and temporal edges. All spatial edges $(v_{t,j},v_{t,k})\in E_{ST}$ are collected in the intra-frame edge set $E_S$,
and all temporal edges $(v_{t,j},v_{t+1,j})\in E_{ST}$ are collected in the inter-frame edge set $E_T$.
The literature \cite{59} proposes a novel deep learning framework, namely Spatio-Temporal Graph Convolutional Networks (STGCN), to tackle the traffic forecasting problem.
Specifically, STGCN consists of several layers of spatio-temporal convolutional blocks, each of which has a "sandwich" structure with two temporal gated convolution layers (abbreviated as Temporal Gated-Conv)
and a spatial graph convolution layer in between (abbreviated as Spatial Graph-Conv). The Spatial Graph-Conv exploits the conventional GCNNs to extract the spatial features,
whereas the Temporal Gated-Conv the temporal gated convolution operator to extract temporal features. Suppose that the input of the temporal gated convolution for each node is a $\text{length-}M$ sequence
with $C_{\text{in}}$ channels, i.e. $X\in\mathbb{R}^{M\times C_{\text{in}}}$. The temporal gated convolution kernel $\Gamma\in\mathbb{R}^{K\times C_{\text{in}}\times 2C_{\text{out}}}$ is used to
filter the input $Y$, i.e.
\begin{displaymath}
\Gamma\ast_T Y=P\circledast\sigma(Q)\in\mathbb{R}^{(M-K+1)\times C_{\text{out}}},
\end{displaymath}
to yield an output $P\bowtie Q\in\mathbb{R}^{(M-K+1)\times(2C_{\text{out}})}$.
The Spatial Graph-Conv takes a tensor $\underline{X}^{(l)}\in\mathbb{R}^{M\times N_{ST}\times C^{(l)}}$ as input, and outputs a tensor $\underline{X}^{(l+1)}\in\mathbb{R}^{(M-2(K-1))\times N_{ST}\times C^{(l+1)}}$, i.e.
\begin{displaymath}
\underline{X}^{(l+1)}=\Gamma_1^{(l)}\ast_T\rho\left(\Theta^{(l)}\ast_G\left(\Gamma_0^{(l)}\ast_T X^{(l)}\right)\right),
\end{displaymath}
where $\Gamma_{0}^{(l)},\Gamma_1^{(l)}$ are the upper and lower temporal kernel and $\Theta^{(l)}$ is the spectral kernel of the graph convolution.
In addition, some other studies pay attention to the GCNNs on the spatio-temporal graphs from other perspectives, e.g. Structural-RNN \cite{56} via a factor graph representation of the spatio-temporal graph
and GCRNN \cite{108,109} combining the vanilla GCNN  and RNN.

\subsubsection{Hypergraphs.}
The aforementioned GCNN architectures are concerned with the conventional graphs consisting of pairwise connectivity between two nodes.
However, there could be even more complicated connections between nodes beyond the pairwise connectivity, e.g. co-authorship networks. Under such circumstances, a hypergraph, as a generalization to the convectional graph,
provides a flexible and elegant modeling tools to represent these complicated connections between nodes. A hypergraph is usually denoted as $\mathcal{G}=(\mathcal{V},\mathcal{E},\omega)$,
where $\mathcal{V}=\{v_1,\cdots,v_N\}$ like the conventional graph, $\mathcal{E}=\{e_1,\cdots,e_M\}$ is a set of $M$ hyperedges.
$\omega(e_k)$ denote weights of hyperedges $e_k\in\mathcal{E}$. A non-trivial hyperedge is a subset of $\mathcal{V}$ with at least 2 nodes.
In particular, a trivial hyperedge, called a self-loop, is composed of a single node. The hypergraph $\mathcal{G}$ can also be denoted by an incidence matrix $\mathcal{H}_{\mathcal{G}}\in\mathbb{R}^{N\times M}$,
i.e.
\begin{displaymath}
\mathcal{H}_{\mathcal{G}}[j,k]=
\left\{
\begin{array}{ll}
0, & v_j\notin e_k \\
1, & v_j\in e_k.
\end{array}
\right.
\end{displaymath}
For a node $v_j\in\mathcal{V}$, its degree $\text{deg}_{\mathcal{V}}(v_j)=\sum_{e_k\in\mathcal{E}}\omega(e_k)\mathcal{H}_{\mathcal{G}}[j,k]$.
For a hyperedge $e_k\in\mathcal{E}$, its degree $\text{deg}_{\mathcal{E}}(e_k)=\sum_{v_j\in e_k}\mathcal{H}_{\mathcal{G}}[j,k]$.
Let $\mathcal{D}_{\mathcal{V}}=\text{diag}\left(\text{deg}_{\mathcal{V}}(v_1),\cdots,\text{deg}_{\mathcal{V}}(v_N)\right)$,
$\mathcal{D}_{\mathcal{E}}=\text{diag}\left(\text{deg}_{\mathcal{E}}(e_1),\cdots,\text{deg}_{\mathcal{E}}(e_M)\right)$, and
$\mathcal{W}_{\mathcal{G}}=\text{diag}\left(\omega(e_1),\cdots,\omega(e_M)\right)$.
The hypergraph Laplacian \cite{65} $\mathcal{L}_{\mathcal{G}}$ of $\mathcal{G}$ is defined to be
\begin{displaymath}
\mathcal{L}_{\mathcal{G}}=\mathcal{I}_N-\mathcal{D}_{\mathcal{V}}^{-\frac{1}{2}}\mathcal{H}_{\mathcal{G}}\mathcal{W}_{\mathcal{G}}\mathcal{D}_{\mathcal{E}}^{-1}\mathcal{H}_{\mathcal{G}}^T\mathcal{D}_{\mathcal{V}}^{-\frac{1}{2}}.
\end{displaymath}
It can also be factorized by the eigendecomposition, i.e. $\mathcal{L}_{\mathcal{G}}=\mathcal{U}\Lambda\mathcal{U}^T$.
The spectral hypergraph convolution operator, the Chebyshev hypergraph convolutional neural network and the hypergraph convolutional network
can be defined in analogy to the Eqs (\ref{Spectral Graph Convolution},\ref{Chebyshev Spectral Filter},\ref{GCN-Layer}).
The HyperGraph Neural Network (HGNN) architecture proposed in the literature \cite{65} is composed of multiple layers of the hyperedge convolution, and each layer is defined as
\begin{displaymath}
X^{(l+1)}=\rho\left(\mathcal{D}_{\mathcal{V}}^{-\frac{1}{2}}\mathcal{H}_{\mathcal{G}}\mathcal{W}_{\mathcal{G}}\mathcal{D}_{\mathcal{E}}^{-1}\mathcal{H}_{\mathcal{G}}^T\mathcal{D}_{\mathcal{V}}^{-\frac{1}{2}}X^{(l)}\Theta^{(l)}\right).
\end{displaymath}
In essence, the HGNN essentially views each hyperedge as a complete graph so that the hypergraph is converted into a conventional graph.
Treating each hyperedge as a complete graph obviously incurs expensive computational cost. Hence, some studies \cite{63,64} propose various approaches to approximate the hyperedges.
The HNHN \cite{174} interleaves updating the node representations with the hyperedge representations by the following formulas,
\begin{displaymath}
\mathcal{X}_{\mathcal{V}}^{(l+1)}=\rho\left(\mathcal{D}_{\mathcal{V}}^{-1}\mathcal{H}_{\mathcal{G}}\mathcal{X}_{\mathcal{E}}^{(l)}\Theta_{\mathcal{V}}^{(l)}\right),\quad
\mathcal{X}_{\mathcal{E}}^{(l+1)}=\rho\left(\mathcal{D}_{\mathcal{E}}^{-1}\mathcal{H}_{\mathcal{G}}^{T}\mathcal{X}_{\mathcal{V}}^{(l)}\Theta_{\mathcal{E}}^{(l)}\right).
\end{displaymath}

%%%%%%%%%%%%%%%%%%%%%%%%%%%%%%%%%%%%%%%%%%%%%%%%%%%%%%%%%%%%%%%%%%%%%%%%%%%%%%%%
\subsection{Capability and Interpretability}
%%%%%%%%%%%%%%%%%%%%%%%%%%%%%%%%%%%%%%%%%%%%%%%%%%%%%%%%%%%%%%%%%%%%%%%%%%%%%%%%

The GCNNs have achieved tremendous empirical successes over the supervised, semi-supervised and unsupervised learning on graphs. Recently,
many studies start to put their eyes on the capability and interpretability of the GCNNs.

\subsubsection{Capability}
The capability of the GCNNs refers to their expressive power. If two graphs are isomorphic, they will obviously output the same representations of nodes/edges/graph.
Otherwise, they should output different representations. However, two non-isomorphic graphs maybe output the same representations in practice.
This is the theoretical limitations of the GCNNs.
As described in the literatures \cite{83,88,360}, The $1\text{-hop}$ spatial GCNNs (1-GCNNs) have the same expressive power as the $1\text{-dimensional}$ Weisfeiler-Leman (1-WL)
graph isomorphism test in terms of distinguishing non-isomorphic graphs.
The 1-WL iteratively update the colors of nodes according to the following formula
\begin{displaymath}
C_{l}^{(l+1)}(v)=\text{\textsc{Hash}}\left(C_{l}^{(l)}(v),\left\{\hspace{-1.5mm}\left\{C_{l}^{(l)}(u):u\in N_G(v)\right\}\hspace{-1.5mm}\right\}\right).
\end{displaymath}
According to the literatures \cite{83,88}, we have that the 1-GCNN architectures do not have more power in terms of distinguishing two non-isomorphic graphs than the 1-WL heuristic. Nevertheless,
they have equivalent power if the aggregation and update functions are injective. In order to overcome
the theoretical limitations of the GCNNs, the literature \cite{83} proposes a Graph Isomorphism Network (GIN) architecture, i.e.
\begin{displaymath}
\begin{array}{rl}
A_v^{(l+1)}& =\text{\textsc{Aggregate}}^{(l+1)}\left(\left\{\left\{X^{(l)}[u,:]:u\in N_G(v)\right\}\right\}\right) \\[2mm]
           & \triangleq\sum_{u\in N_G(v)}X^{(l)}[u,:] \\[2mm]
X^{(l+1)}[v,:] & =\text{\textsc{Update}}^{(l+1)}\left(X^{(l)}[v,:],A_v^{(l+1)}\right) \\ [2mm]
               & \triangleq\text{MLP}\left((1+\epsilon^{(l+1)})X^{(l)}[v,:]+A_v^{(l+1)}\right),
\end{array}
\end{displaymath}
where $\epsilon^{(l)}$ is a scalar parameter. The literature \cite{163} studies the expressive power of the spatial GCNNs, and presents two results:
(1) The spatial GCNNs are shown to be a universal approximator under sufficient conditions on their depth, width, initial node features and layer expressiveness;
(2) The power of the spatial GCNNs is limited when their depth and width is restricted.
In addition, there are some other studies on the capability of the GCNNs from different perspectives, e.g. the first order logic \cite{164},
$p\text{-order}$ graph moments \cite{167}, algorithmic alignment with the dynamic programming \cite{401}, generalization and representational limits of the GNNs \cite{407}.

\subsubsection{Interpretability}
Interpretability plays a vital role in constructing a reliable and intelligent learning systems.
Although some studies have started to explore the interpretability of the conventional deep learning models, few of studies put their eyes on the interpretability of the GNs \cite{85}.
The literature \cite{165} bridges the gap between the empirical success of the GNs and lack of theoretical interpretations. More specifically, it considers two classes of techniques:
(1) gradient based explanations, e.g. sensitivity analysis and guided back-propagation; (2) decomposition based explanations, e.g. layer-wise relevance propagation and Taylor decomposition.
The GNNExplainer \cite{166} is a general and model-agnostic approach for providing interpretable explanations for any spatial GCNN based model in terms of graph machine learning tasks.
Given a trained spatial GCNN model $\Phi$ and a set of predictions, the GNNExplainer will generate a single-instance explanation by identifying a subgraph
of the computation graph and a subset of initial node features, which are the most vital for the prediction of the model $\Phi$. In general, the GNNExplainer can be formulated as an optimization problem
\begin{equation}\label{GNNExplainer optimization framework}
\max_{G_S,X_S^F} I\left(Y,(G_S,X_S^F)\right)=H\left(Y\right)-H\left(Y|G=G_S,X=X_S^F\right),
\end{equation}
where $I(\cdot,\cdot)$ denotes the mutual information of two random variables,
$G_S$ is a small subgraph of the computation graph and $X_S^F$ is a small subset of node features $\left\{X^F[j,:]:v_j\in G_S\right\}$.
The entropy term $H(Y)$ is constant because the spatial GCNN model $\Phi$ is fixed.
In order to improve the tractability and computational efficiency of the GNNExplainer, the final optimization framework is reformulated as
\begin{displaymath}
\min_{M,F} -\sum_{c=1}^C\mathbb{I}[y=c]\log P_{\Phi}\left(Y=y|G=A_G\circledast\sigma(M),X=X_S^F\right).
\end{displaymath}
In addition, the GNNExplainer also provides multi-instances explanations based on graph alignments and prototypes so as to answer questions like "How did a GCNN predict that a given set of nodes all have label $c$?".

%%%%%%%%%%%%%%%%%%%%%%%%%%%%%%%%%%%%%%%%%%%%%%%%%%%%%%%%%%%%%%%%%%%%%%%%%%%%%%%%
\subsection{Deep Graph Representation Learning}
%%%%%%%%%%%%%%%%%%%%%%%%%%%%%%%%%%%%%%%%%%%%%%%%%%%%%%%%%%%%%%%%%%%%%%%%%%%%%%%%

Graph representation learning (or called network embedding) is a paradigm of unsupervised learning on graphs. It gains a large amount of popularity since the DeepWalk \cite{154}.
Subsequently, many studies exploit deep learning techniques to learn low-dimensional representations of nodes \cite{139}. In general, the network embedding via the vanilla deep learning
techniques learn low-dimensional feature vectors of nodes by utilizing either stacked auto-encoders to reconstruct the adjacent or positive point-wise mutual information features \cite{124,132,133,136,141,142}
or RNNs to capture long and short-term dependencies of node sequences yielded by random walks \cite{134,135}. In the following, we introduce the network embedding approaches based on GNNs.

\subsubsection{Network Embedding based on GNNs}
In essence, the GNNs provides an elegant and powerful framework for learning node/edge/graph representations. The majority of the GCNNs and GRNNs are concerned with
semi-supervised learning (i.e. node-focused tasks) or supervised learning (i.e. graph-focused) tasks. Here, we review the GNN based unsupervised learning on graphs.
In general, the network embedding based on GNNs firstly utilize the GCNNs and variational auto-encoder to generate gaussian-distributed hidden states of nodes, and then reconstruct the adjacency matrix
and/or the feature matrix of the input graph \cite{87,127,129,131,138,148}. A representative approach among these ones is the Variational Graph Auto-Encoder (VGAE) \cite{87}
consisting of a GCN based encoder and an inner product decoder.
The GCN based encoder is defined to be
\begin{displaymath}
q(Z|X,A_G)=\prod_{j=1}^Nq(Z[j,:]|X,A_G)=\prod_{j=1}^N\mathcal{N}\left(Z[j,:]|\mu_j,\text{diag}(\sigma_j^2)\right),
\end{displaymath}
where $\mu_j=\text{GCN}_{\mu}\left(X,A_G\right)$ and $\log\sigma_j=\text{GCN}_{\sigma}(X,A_G)$.
The inner product decoder is defined to be

\begin{displaymath}
p(A_G|Z)=\prod_{j=1}^N\prod_{k=1}^Np\left(A_G[j,k]|Z[j,:],Z[k,:]\right)=\prod_{j=1}^N\prod_{k=1}^N\sigma\left(Z[j,:]Z[k,:]^T\right).
\end{displaymath}
They adopt the evidence lower bound \cite{151} as their objective function.
The adversarially regularized (variational) graph autoencoder \cite{129} extends the VGAE by adding an adversarial regularization term to the evidence lower bound.
The literature \cite{128} proposes a symmetric graph convolutional autoencoder which produces a low-dimensional latent nodes representations. Its encoder employs
the Laplacian smoothing \cite{16} to jointly encode the structural and attributed information,
and its decoder is designed based on Laplacian sharpening as the counterpart of the Laplacian smoothing of the encoder. The Laplacian sharpening is defined to be
$X^{(l+1)}=(1+\gamma)X^{(l)}-\gamma D^{-1}AX^{(l)}=X^{(l)}+\gamma(\mathbb{I}_N-D^{-1}A)X^{(l)}$,
which allows utilizing the graph structure in the whole processes of the proposed autoencoder architecture.
In addition, there are some other methods to perform the unsupervised learning on graphs, which do not rely on the reconstruction of the adjacency and/or feature matrix,
e.g. the graph auto-encoder on directed acyclic graphs \cite{79}, pre-training GNNs via context prediction and attribute masking strategies \cite{130},
and deep graph Infomax using a noise-contrastive type objective with a standard binary cross-entropy loss between positive examples and negative examples \cite{140}.

%%%%%%%%%%%%%%%%%%%%%%%%%%%%%%%%%%%%%%%%%%%%%%%%%%%%%%%%%%%%%%%%%%%%%%%%%%%%%%%%
\subsection{Deep Graph Generative Models}
%%%%%%%%%%%%%%%%%%%%%%%%%%%%%%%%%%%%%%%%%%%%%%%%%%%%%%%%%%%%%%%%%%%%%%%%%%%%%%%%

The aforementioned work concentrates on embedding an input graph into a low-dimensional vector space so as to perform semi-supervised/supervised/unsupervised learning tasks on graphs.
This subsection introduces deep graph generative models aiming to mimic real-world complex graphs.
Generating complex graphs from latent representations is confronted with great challenges due to high nonlinearity and arbitrary connectivity of graphs.
Note that graph translation \cite{145} is akin to graph generation. However, their difference lies in that the former takes two graphs, i.e. input graph and target graph, as input,
and the latter only takes a single graph as input. The NetGAN \cite{154} utilizes the generative adversarial network \cite{149} to mimic the input real-world graphs.
More specifically, it is composed of two components, i.e. a generator $G$ and a discriminator $D$, as well. The discriminator $D$ is modeled as a LSTM
in order to distinguish real node sequences, which are yielded by the second-order random walks scheme node2vec \cite{155}, from faked ones.
The generator $G$ aims to generate faked node sequences via another LSTM, whose generating process is as follows.
\begin{displaymath}
\begin{array}{l}
v_0=0, \quad z\sim N_m(0,\mathbb{I}_m), \quad (C_0,h_0)=g_{\theta'}(z), \\ [2mm]
(C_j,h_j,o_j)=\text{LSTM}_G(C_{j-1},h_{j-1},v_{j-1}),\quad v_j\sim\text{Cat}\left(\text{Softmax}(o_j)\right).
\end{array}
\end{displaymath}
The motif-targeted GAN \cite{147} generalizes random walk based architecture of the NetGAN to characterize mesoscopic context of nodes.
Different from \cite{147,154}, the GraphVAE \cite{146} adopts a probabilistic graph decoder to generate a probabilistic fully-connected graph, and then employs approximate graph matching to reconstruct the input graph.
Its reconstruction loss is the cross entropy between the input and reconstructed graphs.
The literature \cite{150} defines a sequential decision-making process to add a node/edge via the graph network \cite{85}, readout operator and softmax function.
The GraphRNN \cite{152} is a deep autoregressive model, which generates graphs by training on a representative set of graphs and decomposes the graph generation process into
a sequence of node and edge formations conditioned on the current generated graph.

%%%%%%%%%%%%%%%%%%%%%%%%%%%%%%%%%%%%%%%%%%%%%%%%%%%%%%%%%%%%%%%%%%%%%%%%%%%%%%%%
\subsection{Combinations of the PI and GNNs}
%%%%%%%%%%%%%%%%%%%%%%%%%%%%%%%%%%%%%%%%%%%%%%%%%%%%%%%%%%%%%%%%%%%%%%%%%%%%%%%%

The GNNs and PI are two different learning paradigms for complicated real-world data.
The former specializes in learning hierarchical representations based on local and global structural information, and the latter learning the dependencies between random variables.
This subsection provides a summarization of studies of combining these two paradigms.

\subsubsection{Conditional Random Field Layer Preserving Similarities between Nodes}
The literature \cite{162} proposes a CRF layer for the GCNNs to enforce hidden layers to preserve similarities between nodes.
Specifically, the input $X^{(l)}$ to $(l+1)\text{-th}$ layer is a random vector around the output $B^{(l)}=\text{GCNN}(X^{(l-1)},A_G, X)$ of the $(l-1)\text{-th}$ layer.
The objective function for the GCNN with a CRF layer can be reformulated as
\begin{displaymath}
\begin{array}{l}
J\left(W;X,A_G,Y\right) =\mathcal{L}\left(Y;B^{(L)}\right)+\displaystyle\sum_{l=1}^{L-1}\left(\frac{\gamma}{2}\|X^{(l)}-B^{(l)}\|_F^2+\mathcal{R}(X^{(l)})\right),
\end{array}
\end{displaymath}
where the first term after "=" is the conventional loss function for semi-supervised node classification problem, and
the last term is a regularization one implementing similarity constraint. The similarity constraint $\mathcal{R}(X^{(l)})$ is modeled as a CRF, i.e.
$p\left(X^{(l)}|B^{(l)}\right)=\frac{1}{Z(B^{(l)})}\exp\left(-E\left(X^{(l)}|B^{(l)}\right)\right)$
where the energy function $E\left(X^{(l)}|B^{(l)}\right)$ is formulated as
\begin{displaymath}
\begin{array}{ll}
 & E\left(X^{(l)}|B^{(l)}\right) \\
 = & \displaystyle\sum_{v\in V}\varphi_v(X^{(l)}[v,:],B^{(l)}[v,:])+\displaystyle\sum_{(u,v)\in E}\varphi_{u,v}\left(X^{(l)}[u,:],X^{(l)}[v,:],B^{(l)}[u,:],B^{(l)}[v,:]\right).
\end{array}
\end{displaymath}
Let $s_{u,v}$ denote the similarity between $u$ and $v$. The unary energy component $\varphi_v\left(\cdot,\cdot\right)$ and pairwise energy component
$\varphi_{u,v}\left(\cdot,\cdot,\cdot,\cdot\right)$ for implementing the similarity constraint are respectively formulated as
\begin{displaymath}
\begin{array}{l}
\varphi_v\left(X^{(l)}[v,:],B^{(l)}[v,:]\right)=\|X^{(l)}[v,:]-B^{(l)}[v,:]\|_2^2, \\[1.5mm]
\varphi_{u,v}\left(X^{(l)}[u,:],X^{(l)}[v,:],B^{(l)}[u,:],B^{(l)}[v,:]\right)=s_{u,v}\|X^{(l)}[u,:]-X^{(l)}[v,:]\|_2^2.
\end{array}
\end{displaymath}
The mean-field variational Bayesian inference is employed to approximate the posterior $p(B^{(l)}|X^{(l)})$. Consequently, the CRF layer is defined as
\begin{displaymath}
\left(X^{(l)}[v,:]\right)^{(k+1)}=\frac{\alpha B^{(l)}[v,:]+\beta\sum_{u\in N_G(v)}s_{u,v}\left(X^{(l)}[u,:]\right)^{(k)}}{\alpha+\beta\sum_{u\in N_G(v)}s_{u,v}}.
\end{displaymath}

\subsubsection{Conditional GCNNs for Semi-supervised Node Classification}
The conditional GCNNs incorporate the Conditional Random Field (CRF) into the conventional GCNNs so that the semi-supervised node classification can be enhanced by
both the powerful node representations and the dependencies of node labels.
The GMNN \cite{80} performs the semi-supervised node classification by incorporating the GCNN into the Statistical Relational Learning (SRL).
Specifically, the SRL usually models the conditional probability $p(Y_V|X_V)$ with the CRF, i.e.
$p(Y_V|X_V)=\frac{1}{Z(X_V)}\prod_{(i,j)\in E}\varphi_{i,j}\left(y_i,y_j,X_V\right)$,
where $y_*=Y_V[*,:], *=i,j$. Note that $Y_V$ is composed of the observed node labels $Y_L$ and hidden node labels $Y_U$.
The variational Bayesian inference is employed to estimate the posterior $p(Y_U|Y_L,X_V)$.
The objective function is defined as
\begin{displaymath}
\text{ELBO}\left(q_{\theta_v}(Y_U|X_V)\right)=\mathbb{E}_{q_{\theta_v}(Y_U|X_V)}\left[\log\left(p_{\theta_l}(Y_L,Y_U|X_V)\right)-\log\left(q_{\theta_v}(Y_U|X_V)\right)\right].
\end{displaymath}
This objective can be optimized by the variational Bayesian EM algorithm \cite{161}, which iteratively updates the variational distribution $q_{\theta_v}(Y_U|X_V)$ and the likelihood $p_{\theta_l}(Y_U|Y_L,X_V)$.
In the VBE stage, $q_{\theta_v}(Y_U|X_V)=\prod_{v\in U}q_{\theta_v}(y_v|X_V)$, and $q_{\theta_v}(y_v|X_V)$ is approximated by a GCNN.
In the VBM stage, the pseudo-likelihood is employed to approximate
\begin{displaymath}
\mathbb{E}_{q_{\theta_v}(Y_U|X_V)}\left[\sum_{v\in V}\log p_{\theta_l}\left(y_n|Y_{V\backslash v},X_V\right)\right]
=\mathbb{E}_{q_{\theta_v}(Y_U|X_V)}\left[\sum_{v\in V}\log p_{\theta_l}\left(y_n|Y_{N_G(v)},X_V\right)\right],
\end{displaymath}
and $p_{\theta_l}\left(y_n|Y_{N_G(v)},X_V\right)$ is approximated by another GCNN. The literature \cite{159} adopts the similar idea to the GMNN.
Its posterior is modeled as a CRF with unary energy components and pairwise energy components whose condition is the outputs of the prescribed GCNN.
The maximum likelihood estimation employed to estimate the model parameters.

\subsubsection{GCNN-based Gaussian Process Inference}
A Gaussian Process (GP) defines a distribution over a function space and assumes any finite collection of marginal distributions follows a multivariate Gaussian distribution.
A function $f:\mathbb{R}^d\rightarrow\mathbb{R}$ follows a Gaussian Process $\text{GP}(m(\cdot),\kappa(\cdot,\cdot))$ iff $\left(f(X_1),\cdots,f(X_N)\right)^T$ for any $N$ $d\text{-dimensional}$ random vectors.
follows a $N\text{-dimensional}$ Gaussian distribution $\mathcal{N}_N\left(\mu,\Sigma\right)$, where $\mu=\left(m(X_1),\cdots,m(X_N)\right)^T$ and $\Sigma=\left[\kappa(X_j,X_k)\right]_{N\times N}$,
For two $d\text{-dimensional}$ random vectors $X$ and $X'$, we have $\mathbb{E}\left[f(X)\right]=m(X)$ and $\text{Cov}(f(X),f(X'))=\kappa(X,X')$.
Given a collection of $N$ samples $\mathcal{D}=\left\{(X_j,y_j):j=1,\cdots N\right\}$, the GP inference aims to calculate the probability $p(y|X)$ for predictions, i.e.
\begin{displaymath}
f\sim\text{GP}(0(\cdot),\kappa(\cdot,\cdot)), y_j\sim\text{DIST}(\lambda(f(X_j))),
\end{displaymath}
where $\lambda(\cdot)$ is a link function and $\text{DIST}(\cdot)$ denotes an arbitrary feasible noise distribution. To this end, the posterior $p(f|\mathcal{D})$ needs to be calculated out firstly.
The literature \cite{156} employs amortized variational Bayesian inference to approximate $p(f|\mathcal{D})$, i.e. $f=\mu+L\epsilon$, and the GCNNs to estimate $\mu$ and $L$.

\subsubsection{Other GCNN-based Probabilistic Inference}
The literature \cite{157} combines the GCNNs and variational Bayesian inference to infer the input graph structure.
The literature \cite{158} infers marginal probabilities in probabilistic graphical models by incorporating the GCNNs to the conventional message-passing inference algorithm.
The literature \cite{168} approximates the posterior in Markov logic networks with the GCNN-enhanced variational Bayesian inference.

%%%%%%%%%%%%%%%%%%%%%%%%%%%%%%%%%%%%%%%%%%%%%%%%%%%%%%%%%%%%%%%%%%%%%%%%%%%%%%%%
\subsection{Adversarial Attacks for the GNNs}
%%%%%%%%%%%%%%%%%%%%%%%%%%%%%%%%%%%%%%%%%%%%%%%%%%%%%%%%%%%%%%%%%%%%%%%%%%%%%%%%

In many circumstances where classifiers are deployed, adversaries deliberately contaminate data in order to fake the classifiers \cite{149,364}.
This is the so-called adversarial attacks for the classification problems. As stated previously, the GNNs can solve semi-supervised node classification problems and supervised graph classification tasks.
Therefore, it is inevitable to study the adversarial attacks for GNNs and defense \cite{365}.

\subsubsection{Adversarial Attacks on Graphs}
The literature \cite{182} firstly proposes a reinforcement learning based attack method, which can learn a generalizable attack policy, on graphs.
This paper provides a definition for a graph adversarial attacker.
Given a sample $(G,c,y)\in\{(G_j,c_j,y_j):j=1,\cdots,M\}$ and a classifier $f\in\mathcal{F}$,
the graph adversarial attacker $g:\mathcal{F}\times\mathcal{G}\rightarrow\mathcal{G}$ attempts to modify a graph $G=(V,E)$ into $\widetilde{G}=(\widetilde{V},\widetilde{E})$, such that
\begin{displaymath}
\begin{array}{ll}
\max_{g} & \mathbb{I}(f(\widetilde{G},c)\neq y) \\ [1.5mm]
\text{s.t.}          & \widetilde{G}=g(f,(G,c,y)), \\ [1.5mm]
                     & \mathcal{I}(G,\widetilde{G},c)=1,
\end{array}
\end{displaymath}
where $\mathcal{I}: \mathcal{G}\times\mathcal{G}\times V\rightarrow\{0,1\}$, named an equivalence indicator,
checks whether two graph $G$ and $\widetilde{G}$ are equivalent under the classification semantics.
The equivalence indicator are usually defined in two fashions, namely explicit semantics and small modifications.
The explicit semantics are defined as $\mathcal{I}\left(G,\widetilde{G},c\right)=\mathbb{I}\left(f^*(G,c)=f^*(\widetilde{G},c)\right)$ where $f^*$ is a gold standard classifier,
and the small modifications
are defined as
$\mathcal{I}\left(G,\widetilde{G},c\right)=\mathbb{I}\left(|(E-\widetilde{E})\cup(\widetilde{E}-E)|<m\right)\cdot\mathbb{I}\left(\widetilde{E}\subseteq N(G,b)\right)$
where $N(G,b)=\left\{(u,v):u,v\in V,d_G(u,v)<=b\right\}$. In order to learn an attack policy,
the attack procedure is modeled as a finite horizon Markov Decision Process (MDP) $\mathcal{M}_m(f,G,c,y)$ and is therefore optimized by Q-learning with a hierarchical Q-function.
For the MDP $\mathcal{M}_m(f,G,c,y)$, its action $a_t\in\mathcal{A}\subseteq V\times V$ at time step $t$ is defined to add or delete edges in the graph,
its state $(\widehat{G}_t,c)$ at time step $t$ is a partially modified graph with some of the edges added/deleted from $G$,
and the reward function is defined as
\begin{displaymath}
R\left(\widetilde{G},c\right)=\left\{
\begin{array}{ll}
1 & f(\widetilde{G},c)\neq y \\ [2mm]
-1 & f(\widetilde{G},c)=y.
\end{array}
\right.
\end{displaymath}
Note that the GCNNs are employed to parameterize the Q-function.
The \textsc{Nettack} \cite{181} considers attacker nodes in $\mathcal{A}$ to satisfy a feature attack constraint $X'_{u,j}\neq X_{u,j}^{(0)}\Longrightarrow u\in\mathcal{A}$,
a structure attack constraint $A'_{u,v}\neq A_{u,v}^{(0)}\Longrightarrow u\in\mathcal{A}\vee v\in\mathcal{A}$ and an equivalence indicator constraint
$\sum_{u}\sum_{j}|X_{u,j}^{(0)}-X'_{u,j}|+\sum_{u<v}|A_{u,v}^{(0)}-A'_{u,v}|\leq\Delta$, where $G'$ is derived by perturbing $G^{(0)}$.
Let $\mathcal{P}_{\Delta,\mathcal{A}}^{G0}$ denote the set of all perturbed graphs $G'$ satisfying these three constraints.
The goal is to find a perturbed graph $G'=(A',X')$ that classifies a target node $v_0$ as $c_{\text{new}}$
and maximizes the log-probability/logit to $c_{\text{old}}$, i.e.
$\max_{(A',X')\in\mathcal{P}_{\delta,\mathcal{A}}^{G0}}\max_{c_{\text{new}}\neq c_{\text{old}}}\log Z_{v_0,c_{\text{new}}}^*-\log Z_{v_0,c_{\text{old}}}^*$
where $Z^*=f_{\theta^*}(A',X')$ with $\theta^*=\arg\min_{\theta}\mathcal{L}(\theta;A',X')$. The \textsc{Nettack} employs the GCNNs to model the classifier.
The literature \cite{183} adopts the similar equivalence indicator, and poisoning attacks are mathematically formulated as a bilevel optimization problem, i.e.
\begin{equation}\label{bilevel optimization}
\begin{array}{ll}
\min_{\widetilde{G}} & \mathcal{L}_{\text{attack}}\left(f_{\theta^*}\left(\widetilde{G}\right)\right) \\ [2mm]
s.t.                 & \widetilde{G}\in\mathcal{P}_{\Delta,\mathcal{A}}^{G0} \\ [2mm]
                     & \theta^*=\arg\min_{\theta}\mathcal{L}_{\text{train}}\left(f_{\theta}\left(\widetilde{G}\right)\right).
\end{array}
\end{equation}
This bilevel optimization problem in formula (\ref{bilevel optimization}) is then tackled using meta-gradients,
whose core idea is to treat the adjacency matrix of the input graph as a hyperparameter.

\subsubsection{Defense against the Adversarial Attacks} A robust GCNN requires that it is invulnerable to perturbations of the input graph.
The robust GCN (RGCN) \cite{180} can fortify the GCNs against adversarial attacks. More specifically, it adopts Gaussian
distributions as the hidden representations of nodes, i.e.
\begin{displaymath}
X^{(l+1)}[j,:]\sim N\left(\mu_j^{(l+1)},\text{diag}(\sigma_j^{(l+1)})\right),
\end{displaymath}
in each graph convolution layer so that the effects of adversarial attacks can be absorbed into the variances of the Gaussian distributions.
The Gaussian based graph convolution is defined as
\begin{displaymath}
\mu_j^{(l+1)}=\rho\left(\displaystyle\sum_{v_k\in N_G(v_j)}\frac{\mu_k^{(l)}\circledast\alpha_k^{(l)}}{\sqrt{\widetilde{D}_{j,j}\widetilde{D}_{k,k}}}W_{\mu}^{(l)}\right),
\sigma_j^{(l+1)}=\rho\left(\displaystyle\sum_{v_k\in N_G(v_j)}\frac{\sigma_k^{(l)}\circledast\alpha_k^{(l)}\circledast\alpha_k^{(l)}}{\widetilde{D}_{j,j}\widetilde{D}_{k,k}}W_{\sigma}^{(l)}\right),
\end{displaymath}
where $\alpha_j^{(k)}$ are attention weights. Finally, the overall loss function is defined as regularized cross-entropy.
The literature \cite{176} presents a batch virtual adversarial training method
which appends a novel regularization term to the conventional objective function of the GCNNs, i.e.
\begin{displaymath}
\mathcal{L}=\mathcal{L}_0+\alpha\cdot\frac{1}{N}\sum_{u\in V}E(p(y|X_u,W))+\beta\cdot\mathcal{R}_{vadv}(V,W),
\end{displaymath}
where $\mathcal{L}_0$ is an average cross-entropy loss of all labelled nodes, $E(\cdot)$ is the conditional entropy of a distribution,
and $\mathcal{R}_{\text{vadv}}(V,W)$ is the average Local Distributional Smoothness (LDS) loss for all nodes. Specifically,
$\mathcal{R}_{\text{vadv}}(V,W)=\frac{1}{N}\sum_{u\in V}\text{LDS}(X[u,:],W,r_{\text{vadv},u})$ where $\text{LDS}(x,w,r_{\text{vadv}})=D_{KL}\left(p(y|x,\widehat{W})||p(y|x+r_{\text{vadv}},W)\right)$
and $r_{\text{vadv}}$ is the virtual adversarial perturbation. Additionally, there are some other studies aiming at verifying certifiable (non-)robustness to structure and feature perturbations for the GCNNs
and developing robust training algorithm \cite{177,178}. The literature \cite{180} proposes to improve GCN generalization by minimizing the expected loss
under small perturbations of the input graph.
Its basic assumption is that the adjacency matrix $A_G$ is perturbed by some random noises. Under this assumption, the objective function is defined as
$\min_{W}\int q(\epsilon|\alpha)\mathcal{L}(X,Y,A_G(\epsilon),W)d\epsilon$, where $A_G(\epsilon)$ denotes the perturbed adjacency matrix of $G$ and $q(\epsilon|\alpha)$ is a zero-centered density of the noise $\epsilon$
so that the learned GCN is robust to these noises and generalizes well.

%%%%%%%%%%%%%%%%%%%%%%%%%%%%%%%%%%%%%%%%%%%%%%%%%%%%%%%%%%%%%%%%%%%%%%%%%%%%%%%%
\subsection{Graph Neural Architecture Search}
%%%%%%%%%%%%%%%%%%%%%%%%%%%%%%%%%%%%%%%%%%%%%%%%%%%%%%%%%%%%%%%%%%%%%%%%%%%%%%%%

Neural Architecture Search (NAS) \cite{171} has achieved tremendous success in discovering
the optimal neural network architecture for image and language learning tasks. However, existing NAS algorithms cannot be
directly generalized to find the optimal GNN architecture. Fortunately, there have been some studies to bridge this gap.
The graph neural architecture search \cite{172,173} aims to search for an optimal GNN architecture within a designed search space.
It usually exploits a reinforcement learning based controller, which is a RNN, to greedily validate the generated architecture, and then the validation results are fed back to the controller.
The literature \cite{175} proposes a Graph HyperNetwork (GHN) to amortize the search cost of training thousands of different networks,
which is trained to minimize the training loss of the sampled network with the weights generated by a GCNN.

%%%%%%%%%%%%%%%%%%%%%%%%%%%%%%%%%%%%%%%%%%%%%%%%%%%%%%%%%%%%%%%%%%%%%%%%%%%%%%%%
\subsection{Graph Reinforcement Learning}
%%%%%%%%%%%%%%%%%%%%%%%%%%%%%%%%%%%%%%%%%%%%%%%%%%%%%%%%%%%%%%%%%%%%%%%%%%%%%%%%

The GNNs can also be combined with the reinforcement learning so as to solve sequential decision-making problems on graphs.
The literature \cite{188} learns to walk over a graph from a source node towards a target node for a given query via reinforcement learning.
The proposed agent M-Walk is composed of a deep RNN and Monte Carlo Tree Search (MCTS). The former maps a hidden vector representation $h_t$,
yielded by a special RNN encoding the state $s_t$ at time $t$, to a policy and Q-values, and the latter is employed to generate trajectories yielding more positive rewards.
The NerveNet \cite{187} propagates information over the underlying graph of an agent via a GCNN, and then predicts actions for different parts of the agent.
The literature \cite{186} combines the GNNs and Deep Reinforcement Learning (DRL), named DRL+GNN, to learn, operate and generalize over arbitrary network topologies.
The DRL+GNN agent employs a GCNN to model the Q-value function.

\subsection{Applications}

In this subsection, we introduce the applications of the GNNs.
Due to the space limitation, we only list the application fields, including complex network analysis \cite{233,234,238}, combinatorial optimization \cite{215,216,223,224},
knowledge graph \cite{272,273,289}, bioinformatics \cite{201,210,211}, chemistry \cite{197,202,203}, brain network analysis \cite{205,206},
physical system \cite{102,338,339}, source code analysis \cite{340,341,342}, intelligent traffic \cite{353,355,403}, recommender systems \cite{343,344,349},
computer vision \cite{112,343,344,345,346} and natural language processing \cite{303,308,312,318,326}.

%\begin{figure}[htb]
%  \centering
%  \includegraphics[width=1.0\textwidth]{ApplicationSummarization.pdf}
%  \caption{The GNN Application Summarizations}
%  \label{application summarization}
%\end{figure}

%%%%%%%%%%%%%%%%%%%%%%%%%%%%%%%%%%%%%%%%%%%%%%%%%%%%%%%%%%%%%%%%%%%%%%%%%%%%%%%%
\section{Benchmarks and Evaluation Pitfalls}
%%%%%%%%%%%%%%%%%%%%%%%%%%%%%%%%%%%%%%%%%%%%%%%%%%%%%%%%%%%%%%%%%%%%%%%%%%%%%%%%

In this section, we briefly introduce benchmarks and evaluation Pitfalls. The benchmarks provide ground truth for various GNN architectures so that different GNNs can be compared fairly,
the evaluation pitfalls empirically show that the existing evaluation criterion have potential pitfalls.

\subsubsection{Benchmarks}
Graph neural networks have become a powerful toolkit for mining complex graphs.
It becomes more and more critical to evaluate the effectiveness of new GNN architectures and
compare different GNN models under a standardized benchmark with consistent experimental settings and large datasets.
A feasible benchmark for the GNNs should include appropriate graph datasets, robust coding interfaces and experimental settings
so that different GNN architectures can be compared in the same settings. The literature \cite{169} makes a pioneering effort to
construct a reproducible GNN benchmarking framework in order to facilitate researchers to gauge the effectiveness of different GNN architectures.
Specifically, it releases an open-source benchmark infrastructures for GNNs, hosted on GitHub based on PyTorch and DGL libraries \cite{402},
introduces medium-scale graph datasets with 12k-70k graphs of variable sizes 9-500 nodes, and identifies important building blocks of GNNs
(graph convolutions, anistropic diffusion, residual connections and normalization layers) with the proposed benchmark infrastructures.
The literature \cite{372} presents an Open Graph Benchmark (OGB) including challenging, real-world and large-scale benchmark graph datasets, encompassing
multiple important graph deep learning tasks ranging from social and information networks to biological networks, molecular graphs and knowledge graphs,
to facilitate scalable, robust and reproducible deep learning research on graphs. The OGB datasets, which provide a unified evaluation protocol
using application-specific train/validation/test dataset splits and evaluation metrics,
releases an end-to-end graph processing pipeline including graph data loading, experimental settings and model evaluations.

\subsubsection{Evaluation Pitfalls}
The literature \cite{170} compares four typical GCNN architectures: GCN \cite{6}, MoNet \cite{18},
GAT \cite{66} and GraphSAGE using three aggregation strategies \cite{104}
against 4 baseline models: logistic regression, multi-layer perceptron, label propagation and normalized laplacian label propagation,
and uses a standardized training and hyper-parameter tuning procedure for all these models so as to perform a more fair comparison.
The experimental results show that different train/validation/test splits of datasets lead to dramatically different rankings of models.
In addition, their findings also demonstrate that simpler GCNN architectures can outperform more sophisticated ones
only if the hyper-parameters and training procedures are tuned fairly for all models.

%%%%%%%%%%%%%%%%%%%%%%%%%%%%%%%%%%%%%%%%%%%%%%%%%%%%%%%%%%%%%%%%%%%%%%%%%%%%%%%%
\section{Future Research Directions}
%%%%%%%%%%%%%%%%%%%%%%%%%%%%%%%%%%%%%%%%%%%%%%%%%%%%%%%%%%%%%%%%%%%%%%%%%%%%%%%%

Although the GNNs have achieved tremendous success in many fields, there still exists some open problems.
This section summarizes the future research directions of the GNNs.

\subsubsection{Highly Scalable GNNs}
The real-world graphs usually contain hundreds of millions of nodes and edges, and have dynamically evolving characteristics.
It turns out that it is difficult for the existing GNN architectures to scale up to the huge real-world graphs.
This motivates us to design highly scalable GNN architectures which can efficiently and effectively learn node/edge/graph representations
for the huge dynamically-evolving graphs.

\subsubsection{Robust GNNs}
The existing GNN architectures are vulnerable to adversarial attacks.
That is, the performance of the GNN models will sharply drop once the structure and/or initial features of the input graph are attacked by adversaries.
Therefore, we should incorporate the attack-and-defense mechanism into the GNN architectures, i.e. constructing robust GNN architecture,
so as to reinforce them against adversarial attacks.

\subsubsection{GNNs Going Beyond WL Test}
The capabilities of the spatial GCNNs are limited by the 1-WL test,
and the higher-order WL test is computationally expensive. Consequently, two non-isomorphic graphs
will produce the same node/edge/graph representations under appropriate conditions.
This motivates us to develop a novel GNN framework going beyond WL test, or design an elegant higher-order GNN architectures
corresponding to the higher-order WL test.

\subsubsection{Interpretable GNNs}
The existing GNNs work in a black box. We do not understand why they achieve state-of-the-art performance in terms of
the node classification task, graph classification task and graph embedding task etc. Interpretability has become a major obstacle to
apply the GNNs to real-world issues. Although there have been some studies to interpret some specific GNN models,
they cannot interpret general GNN models. This motivates us to construct a unified interpretable framework for the GNNs.

%%%%%%%%%%%%%%%%%%%%%%%%%%%%%%%%%%%%%%%%%%%%%%%%%%%%%%%%%%%%%%%%%%%%%%%%%%%%%%%%
\section{Conclusions}
%%%%%%%%%%%%%%%%%%%%%%%%%%%%%%%%%%%%%%%%%%%%%%%%%%%%%%%%%%%%%%%%%%%%%%%%%%%%%%%%

This paper aims to provide a taxonomy, advances and trends for the GNNs. We expand the content of this paper from 4 dimensions: architectures, extensions and applications, benchmarks and evaluation pitfalls,
and future research directions.
The GNN architectures are expanded from 4 perspectives: graph convolutional neural networks, graph pooling operators, graph attention mechanisms and graph recurrent neural networks.
The extensions are expanded from 8 perspectives: GCNNS on special graphs, capability and interpretability, deep graph representation learning, deep graph generative models, combinations of the PI and GNNs
adversarial attachks for the GNNs, graph neural architecture search and graph reinforcement. In the future directions, we propose 4 prospective topics on the GNNs: highly scalable GNNs, robust GNNs,
GNNs going beyond WL test and interpretable GNNs. We expect that the relevant scholars can understand the computational principles of the GNNs,
consolidate the foundations of the GNNs and apply them to more and more real-world issues, through reading this review.

\begin{acks}
This work was supported by National Natural Science Foundation of China (Grant No. 62002255, 62076177, 61972273).
\end{acks}

\bibliographystyle{ACM-Reference-Format}
\bibliography{sample-base}


\begin{thebibliography}{251}

%%% the Web address.

\ifx \showCODEN    \undefined \def \showCODEN     #1{\unskip}     \fi
\ifx \showDOI      \undefined \def \showDOI       #1{#1}\fi
\ifx \showISBNx    \undefined \def \showISBNx     #1{\unskip}     \fi
\ifx \showISBNxiii \undefined \def \showISBNxiii  #1{\unskip}     \fi
\ifx \showISSN     \undefined \def \showISSN      #1{\unskip}     \fi
\ifx \showLCCN     \undefined \def \showLCCN      #1{\unskip}     \fi
\ifx \shownote     \undefined \def \shownote      #1{#1}          \fi
\ifx \showarticletitle \undefined \def \showarticletitle #1{#1}   \fi
\ifx \showURL      \undefined \def \showURL       {\relax}        \fi
% invisible to TeX
\providecommand\bibfield[2]{#2}
\providecommand\bibinfo[2]{#2}
\providecommand\natexlab[1]{#1}
\providecommand\showeprint[2][]{arXiv:#2}

\bibitem[\protect\citeauthoryear{{Aditya Grover} and {Jure Leskovec}}{{Aditya
  Grover} and {Jure Leskovec}}{2016}]%
        {155}
\bibfield{author}{\bibinfo{person}{{Aditya Grover}} {and}
  \bibinfo{person}{{Jure Leskovec}}.} \bibinfo{year}{2016}\natexlab{}.
\newblock \showarticletitle{node2vec: scalable feature learning for networks}.
  In \bibinfo{booktitle}{\emph{The International Conference on Knowledge
  Discovery and Data Mining (SIGKDD)}}. \bibinfo{pages}{855--864}.
\newblock


\bibitem[\protect\citeauthoryear{{Afshin Rahimi}, {Trevor Cohn}, and {Timothy
  Baldwin}}{{Afshin Rahimi} et~al\mbox{.}}{2018}]%
        {398}
\bibfield{author}{\bibinfo{person}{{Afshin Rahimi}}, \bibinfo{person}{{Trevor
  Cohn}}, {and} \bibinfo{person}{{Timothy Baldwin}}.}
  \bibinfo{year}{2018}\natexlab{}.
\newblock \showarticletitle{Semi-supervised user geolocation via graph
  convolutional networks}. In \bibinfo{booktitle}{\emph{The Annual Meeting of
  the Association for Computational Linguistics ({ACL})}}.
  \bibinfo{pages}{2009--2019}.
\newblock


\bibitem[\protect\citeauthoryear{{Aleksandar Bojchevski} and {Stephan
  G\"{u}nnemann}}{{Aleksandar Bojchevski} and {Stephan G\"{u}nnemann}}{2019}]%
        {178}
\bibfield{author}{\bibinfo{person}{{Aleksandar Bojchevski}} {and}
  \bibinfo{person}{{Stephan G\"{u}nnemann}}.} \bibinfo{year}{2019}\natexlab{}.
\newblock \showarticletitle{Certifiable robustness to graph perturbations}.
\newblock  (\bibinfo{year}{2019}).
\newblock
\urldef\tempurl%
\url{https://arxiv.org/abs/1910.14356}
\showURL{%
\tempurl}


\bibitem[\protect\citeauthoryear{{Alvaro Sanchez-Gonzalez}, {Nicolas Heess},
  {Jost Tobias Springenberg}, {Josh Merel}, {Martin Riedmiller}, {Raia
  Hadsell}, and {Peter Battaglia}}{{Alvaro Sanchez-Gonzalez}
  et~al\mbox{.}}{2018}]%
        {102}
\bibfield{author}{\bibinfo{person}{{Alvaro Sanchez-Gonzalez}},
  \bibinfo{person}{{Nicolas Heess}}, \bibinfo{person}{{Jost Tobias
  Springenberg}}, \bibinfo{person}{{Josh Merel}}, \bibinfo{person}{{Martin
  Riedmiller}}, \bibinfo{person}{{Raia Hadsell}}, {and} \bibinfo{person}{{Peter
  Battaglia}}.} \bibinfo{year}{2018}\natexlab{}.
\newblock \showarticletitle{Graph networks as learnable physics engines for
  inference and control}. In \bibinfo{booktitle}{\emph{The International
  Conference on Machine Learning ({ICML})}}. \bibinfo{pages}{4470--4479}.
\newblock


\bibitem[\protect\citeauthoryear{{Amir H. Khasahmadi}, {Kaveh Hassani}, {Parsa
  Moradi}, {Leo Lee}, and {Quaid Morris}}{{Amir H. Khasahmadi}
  et~al\mbox{.}}{2020}]%
        {84}
\bibfield{author}{\bibinfo{person}{{Amir H. Khasahmadi}},
  \bibinfo{person}{{Kaveh Hassani}}, \bibinfo{person}{{Parsa Moradi}},
  \bibinfo{person}{{Leo Lee}}, {and} \bibinfo{person}{{Quaid Morris}}.}
  \bibinfo{year}{2020}\natexlab{}.
\newblock \showarticletitle{Memory-based graph networks}. In
  \bibinfo{booktitle}{\emph{The International Conference on Learning
  Representations ({ICLR})}}.
\newblock


\bibitem[\protect\citeauthoryear{{Ana \u\{S\}u\u\{s\}njara}, {Nathana\"{e}l
  Perraudin}, {Daniel Kressner}, and {Pierre Vandergheynst}}{{Ana
  \u\{S\}u\u\{s\}njara} et~al\mbox{.}}{2015}]%
        {30}
\bibfield{author}{\bibinfo{person}{{Ana \u\{S\}u\u\{s\}njara}},
  \bibinfo{person}{{Nathana\"{e}l Perraudin}}, \bibinfo{person}{{Daniel
  Kressner}}, {and} \bibinfo{person}{{Pierre Vandergheynst}}.}
  \bibinfo{year}{2015}\natexlab{}.
\newblock \showarticletitle{Accelerated filtering on graphs using Lanczos
  method}.
\newblock  (\bibinfo{year}{2015}).
\newblock
\urldef\tempurl%
\url{https://arxiv.org/abs/1509.04537}
\showURL{%
\tempurl}


\bibitem[\protect\citeauthoryear{{Andreas Loukas}}{{Andreas Loukas}}{2020}]%
        {163}
\bibfield{author}{\bibinfo{person}{{Andreas Loukas}}.}
  \bibinfo{year}{2020}\natexlab{}.
\newblock \showarticletitle{What graph neural networks cannot learn-depth vs
  width}. In \bibinfo{booktitle}{\emph{The International Conference on Learning
  Representations ({ICLR})}}.
\newblock


\bibitem[\protect\citeauthoryear{{Antoni Buades}, {Bartomeu Coll}, and
  {Jean-Michael Morel}}{{Antoni Buades} et~al\mbox{.}}{2005}]%
        {381}
\bibfield{author}{\bibinfo{person}{{Antoni Buades}}, \bibinfo{person}{{Bartomeu
  Coll}}, {and} \bibinfo{person}{{Jean-Michael Morel}}.}
  \bibinfo{year}{2005}\natexlab{}.
\newblock \showarticletitle{A non-local algorithm for image denoising}. In
  \bibinfo{booktitle}{\emph{The IEEE International Conference on Computer
  Vision and Pattern Recognition (CVPR)}}. \bibinfo{pages}{60--65}.
\newblock


\bibitem[\protect\citeauthoryear{{Anuththari Gamage}, {Eli Chien}, {Jianhao
  Peng}, and {Olgica Milenkovic}}{{Anuththari Gamage} et~al\mbox{.}}{2019}]%
        {147}
\bibfield{author}{\bibinfo{person}{{Anuththari Gamage}}, \bibinfo{person}{{Eli
  Chien}}, \bibinfo{person}{{Jianhao Peng}}, {and} \bibinfo{person}{{Olgica
  Milenkovic}}.} \bibinfo{year}{2019}\natexlab{}.
\newblock \showarticletitle{{Multi-MotifGAN} ({MMGAN}): motif-targeted graph
  generation and prediction}.
\newblock  (\bibinfo{year}{2019}).
\newblock
\urldef\tempurl%
\url{https://arxiv.org/abs/1911.05469v1}
\showURL{%
\tempurl}


\bibitem[\protect\citeauthoryear{{Ashesh Jain}, {Amir R. Zamir}, {Silvio
  Savarese}, and {Ashutosh Saxena}}{{Ashesh Jain} et~al\mbox{.}}{2016}]%
        {56}
\bibfield{author}{\bibinfo{person}{{Ashesh Jain}}, \bibinfo{person}{{Amir R.
  Zamir}}, \bibinfo{person}{{Silvio Savarese}}, {and}
  \bibinfo{person}{{Ashutosh Saxena}}.} \bibinfo{year}{2016}\natexlab{}.
\newblock \showarticletitle{Structural-{RNN}: deep learning on spatio-temporal
  graphs}. In \bibinfo{booktitle}{\emph{The {IEEE} Conference on Computer
  Vision and Pattern Recognition ({CVPR})}}. \bibinfo{pages}{5308--5317}.
\newblock


\bibitem[\protect\citeauthoryear{{Ashish Vaswani}, {Noam Shazeer}, {Niki
  Parmar}, {Jakob Uszkoreit}, {Llion Jones}, {Aidan N. Gomiz}, {\l\{L\}ukasz
  Kaiser}, and {Illia Polosukhin}}{{Ashish Vaswani} et~al\mbox{.}}{2017}]%
        {73}
\bibfield{author}{\bibinfo{person}{{Ashish Vaswani}}, \bibinfo{person}{{Noam
  Shazeer}}, \bibinfo{person}{{Niki Parmar}}, \bibinfo{person}{{Jakob
  Uszkoreit}}, \bibinfo{person}{{Llion Jones}}, \bibinfo{person}{{Aidan N.
  Gomiz}}, \bibinfo{person}{{\l\{L\}ukasz Kaiser}}, {and}
  \bibinfo{person}{{Illia Polosukhin}}.} \bibinfo{year}{2017}\natexlab{}.
\newblock \showarticletitle{Attention is all you need}. In
  \bibinfo{booktitle}{\emph{The International Conference on Neural Information
  Processing ({NIPS})}}. \bibinfo{pages}{5998--6008}.
\newblock


\bibitem[\protect\citeauthoryear{{Bill Yuchen Lin}, {Xinyue Chen}, {Jamin
  Chen}, and {Xiang Ren}}{{Bill Yuchen Lin} et~al\mbox{.}}{2019}]%
        {289}
\bibfield{author}{\bibinfo{person}{{Bill Yuchen Lin}}, \bibinfo{person}{{Xinyue
  Chen}}, \bibinfo{person}{{Jamin Chen}}, {and} \bibinfo{person}{{Xiang Ren}}.}
  \bibinfo{year}{2019}\natexlab{}.
\newblock \showarticletitle{{KagNet}: knowledge-aware graph networks for
  commonsense reasoning}. In \bibinfo{booktitle}{\emph{The International
  Conference on Empirical Methods in Natural Language Processing ({EMNLP})}}.
  \bibinfo{pages}{2829--2839}.
\newblock


\bibitem[\protect\citeauthoryear{{Bing Yu}, {Haoteng Yin}, and {Zhanxing
  Zhu}}{{Bing Yu} et~al\mbox{.}}{2018}]%
        {59}
\bibfield{author}{\bibinfo{person}{{Bing Yu}}, \bibinfo{person}{{Haoteng Yin}},
  {and} \bibinfo{person}{{Zhanxing Zhu}}.} \bibinfo{year}{2018}\natexlab{}.
\newblock \showarticletitle{Spatio-temporal graph convolutional networks: a
  deep learning framework for traffic forecasting}. In
  \bibinfo{booktitle}{\emph{The International Joint Conference on Artificial
  Intelligence ({IJCAI})}}. \bibinfo{pages}{3634--3640}.
\newblock


\bibitem[\protect\citeauthoryear{{Bingbing Xu}, {Huawei Shen}, {Qi Cao}, {Yunqi
  Qiu}, and {Xueqi Cheng}}{{Bingbing Xu} et~al\mbox{.}}{2019}]%
        {191}
\bibfield{author}{\bibinfo{person}{{Bingbing Xu}}, \bibinfo{person}{{Huawei
  Shen}}, \bibinfo{person}{{Qi Cao}}, \bibinfo{person}{{Yunqi Qiu}}, {and}
  \bibinfo{person}{{Xueqi Cheng}}.} \bibinfo{year}{2019}\natexlab{}.
\newblock \showarticletitle{Graph Wavelet Neural Network}. In
  \bibinfo{booktitle}{\emph{The International Conference on Learning
  Representations ({ICLR})}}.
\newblock


\bibitem[\protect\citeauthoryear{{Bryan Perozzi}, {Rami Ai-Rfou}, and {Steven
  Skiena}}{{Bryan Perozzi} et~al\mbox{.}}{2014}]%
        {154}
\bibfield{author}{\bibinfo{person}{{Bryan Perozzi}}, \bibinfo{person}{{Rami
  Ai-Rfou}}, {and} \bibinfo{person}{{Steven Skiena}}.}
  \bibinfo{year}{2014}\natexlab{}.
\newblock \showarticletitle{DeepWalk: online learning of social
  representations}. In \bibinfo{booktitle}{\emph{The International Conference
  on Knowledge Discovery and Data Mining (SIGKDD)}}. \bibinfo{pages}{701--710}.
\newblock


\bibitem[\protect\citeauthoryear{{Caglar Gulcehre}, {Misha Denil}, {Mateusz
  Malinowski}, {Ali Razavi}, {Razvan Pascanu}, {Karl Moritz Hermann}, {Peter
  Battaglia}, {Victor Bapst}, {David Raposo}, {Adam Santoro}, and {Nando de
  Freitas}}{{Caglar Gulcehre} et~al\mbox{.}}{2019}]%
        {69}
\bibfield{author}{\bibinfo{person}{{Caglar Gulcehre}}, \bibinfo{person}{{Misha
  Denil}}, \bibinfo{person}{{Mateusz Malinowski}}, \bibinfo{person}{{Ali
  Razavi}}, \bibinfo{person}{{Razvan Pascanu}}, \bibinfo{person}{{Karl Moritz
  Hermann}}, \bibinfo{person}{{Peter Battaglia}}, \bibinfo{person}{{Victor
  Bapst}}, \bibinfo{person}{{David Raposo}}, \bibinfo{person}{{Adam Santoro}},
  {and} \bibinfo{person}{{Nando de Freitas}}.} \bibinfo{year}{2019}\natexlab{}.
\newblock \showarticletitle{Hyperbolic Attention Networks}. In
  \bibinfo{booktitle}{\emph{The International Conference on Learning
  Representations}}.
\newblock


\bibitem[\protect\citeauthoryear{{Chao Shang}, {Qinqing Liu}, {Ko-Shin Chen},
  {Jiangwen Sun}, {Jin Lu}, {Jinfeng Yi}, and {Jinbo Bi}}{{Chao Shang}
  et~al\mbox{.}}{2018}]%
        {68}
\bibfield{author}{\bibinfo{person}{{Chao Shang}}, \bibinfo{person}{{Qinqing
  Liu}}, \bibinfo{person}{{Ko-Shin Chen}}, \bibinfo{person}{{Jiangwen Sun}},
  \bibinfo{person}{{Jin Lu}}, \bibinfo{person}{{Jinfeng Yi}}, {and}
  \bibinfo{person}{{Jinbo Bi}}.} \bibinfo{year}{2018}\natexlab{}.
\newblock \showarticletitle{Edge attention-based multi-relational graph
  convolutional networks}.
\newblock  (\bibinfo{year}{2018}).
\newblock
\urldef\tempurl%
\url{https://arxiv.org/abs/1802.04944}
\showURL{%
\tempurl}


\bibitem[\protect\citeauthoryear{{Chenyi Zhuang} and {Qiang Ma}}{{Chenyi
  Zhuang} and {Qiang Ma}}{2018}]%
        {19}
\bibfield{author}{\bibinfo{person}{{Chenyi Zhuang}} {and}
  \bibinfo{person}{{Qiang Ma}}.} \bibinfo{year}{2018}\natexlab{}.
\newblock \showarticletitle{Dual graph convolutional networks for graph-based
  semi-supervised classification}. In \bibinfo{booktitle}{\emph{The World Wide
  Web Conference ({WWW})}}. \bibinfo{pages}{499--508}.
\newblock


\bibitem[\protect\citeauthoryear{{Chris Zhang}, {Mengye Ren}, and {Raquel
  Urtasun}}{{Chris Zhang} et~al\mbox{.}}{2019}]%
        {175}
\bibfield{author}{\bibinfo{person}{{Chris Zhang}}, \bibinfo{person}{{Mengye
  Ren}}, {and} \bibinfo{person}{{Raquel Urtasun}}.}
  \bibinfo{year}{2019}\natexlab{}.
\newblock \showarticletitle{Graph hypernetworks for neural architecture
  search}. In \bibinfo{booktitle}{\emph{The International Conference on
  Learning Representations ({ICLR})}}.
\newblock


\bibitem[\protect\citeauthoryear{{Christian Szegedy}, {Wei Liu}, {Yangqing
  Jia}, {Pierre Sermanet}, {Scott Reed}, {Dragomir Anguelov}, {Dumitru Erhan},
  {Vincent Vanhoucke}, and {Andrew Rabinovich}}{{Christian Szegedy}
  et~al\mbox{.}}{2015}]%
        {397}
\bibfield{author}{\bibinfo{person}{{Christian Szegedy}}, \bibinfo{person}{{Wei
  Liu}}, \bibinfo{person}{{Yangqing Jia}}, \bibinfo{person}{{Pierre Sermanet}},
  \bibinfo{person}{{Scott Reed}}, \bibinfo{person}{{Dragomir Anguelov}},
  \bibinfo{person}{{Dumitru Erhan}}, \bibinfo{person}{{Vincent Vanhoucke}},
  {and} \bibinfo{person}{{Andrew Rabinovich}}.}
  \bibinfo{year}{2015}\natexlab{}.
\newblock \showarticletitle{Going deeper with convolutions}. In
  \bibinfo{booktitle}{\emph{The IEEE International Conference on Computer
  Vision and Pattern Recognition (CVPR)}}. \bibinfo{pages}{1--9}.
\newblock


\bibitem[\protect\citeauthoryear{{Christopher Morris}, {Martin Ritzert},
  {Matthias Fey}, {William L. Hamilton}, {Jan Eric Lenssen}, {Gaurav Rattan},
  and {Martin Grohe}}{{Christopher Morris} et~al\mbox{.}}{2019}]%
        {88}
\bibfield{author}{\bibinfo{person}{{Christopher Morris}},
  \bibinfo{person}{{Martin Ritzert}}, \bibinfo{person}{{Matthias Fey}},
  \bibinfo{person}{{William L. Hamilton}}, \bibinfo{person}{{Jan Eric
  Lenssen}}, \bibinfo{person}{{Gaurav Rattan}}, {and} \bibinfo{person}{{Martin
  Grohe}}.} \bibinfo{year}{2019}\natexlab{}.
\newblock \showarticletitle{Weisfeiler and Leman Go neural: higher-order graph
  neural networks}. In \bibinfo{booktitle}{\emph{The {AAAI} Conference on
  Artificial Intelligence ({AAAI})}}. \bibinfo{pages}{4602--4609}.
\newblock


\bibitem[\protect\citeauthoryear{{Chuanpan Zheng}, {Xiaoliang Fan}, {Cheng
  Wang}, and {Jianzhong Qi}}{{Chuanpan Zheng} et~al\mbox{.}}{2020}]%
        {353}
\bibfield{author}{\bibinfo{person}{{Chuanpan Zheng}},
  \bibinfo{person}{{Xiaoliang Fan}}, \bibinfo{person}{{Cheng Wang}}, {and}
  \bibinfo{person}{{Jianzhong Qi}}.} \bibinfo{year}{2020}\natexlab{}.
\newblock \showarticletitle{{GMAN}: a graph multi-attention network for traffic
  prediction}. In \bibinfo{booktitle}{\emph{The {AAAI} Conference on Artificial
  Intelligence ({AAAI})}}.
\newblock


\bibitem[\protect\citeauthoryear{{Chun Wang}, {Shirui Pan}, {Guodong Long},
  {Xingquan Zhu}, and {Jing Jiang}}{{Chun Wang} et~al\mbox{.}}{2017}]%
        {148}
\bibfield{author}{\bibinfo{person}{{Chun Wang}}, \bibinfo{person}{{Shirui
  Pan}}, \bibinfo{person}{{Guodong Long}}, \bibinfo{person}{{Xingquan Zhu}},
  {and} \bibinfo{person}{{Jing Jiang}}.} \bibinfo{year}{2017}\natexlab{}.
\newblock \showarticletitle{{MGAE}: marginalized graph autoencoder for graph
  clustering}. In \bibinfo{booktitle}{\emph{The International Conference on
  Information and Knowledge Management ({CIKM})}}. \bibinfo{pages}{889--898}.
\newblock


\bibitem[\protect\citeauthoryear{Chung}{Chung}{1992}]%
        {374}
\bibfield{author}{\bibinfo{person}{Fan~R.K. Chung}.}
  \bibinfo{year}{1992}\natexlab{}.
\newblock \bibinfo{booktitle}{\emph{Spectral graph theory}}.
\newblock \bibinfo{publisher}{American Mathematical Society}.
\newblock


\bibitem[\protect\citeauthoryear{{Chuxu Zhang}, {Dongjin Song}, {Chao Huang},
  {Ananthram Swami}, and {Nitesh V. Chawla}}{{Chuxu Zhang}
  et~al\mbox{.}}{2019}]%
        {61}
\bibfield{author}{\bibinfo{person}{{Chuxu Zhang}}, \bibinfo{person}{{Dongjin
  Song}}, \bibinfo{person}{{Chao Huang}}, \bibinfo{person}{{Ananthram Swami}},
  {and} \bibinfo{person}{{Nitesh V. Chawla}}.} \bibinfo{year}{2019}\natexlab{}.
\newblock \showarticletitle{Heterogeneous Graph Neural Network}. In
  \bibinfo{booktitle}{\emph{Th {ACM} International Conference on Knowledge
  Discovery and Data Mining ({SIGKDD})}}. \bibinfo{pages}{793--803}.
\newblock


\bibitem[\protect\citeauthoryear{{Claire Donnat}, {Marinka Zitnik}, {David
  Hallac}, and {Jure Leskovec}}{{Claire Donnat} et~al\mbox{.}}{2018}]%
        {194}
\bibfield{author}{\bibinfo{person}{{Claire Donnat}}, \bibinfo{person}{{Marinka
  Zitnik}}, \bibinfo{person}{{David Hallac}}, {and} \bibinfo{person}{{Jure
  Leskovec}}.} \bibinfo{year}{2018}\natexlab{}.
\newblock \showarticletitle{Learning structural node embeddings via diffusion
  wavelets}. In \bibinfo{booktitle}{\emph{The International Conference on
  Knowledge Discovery and Data Mining ({SIGKDD})}}.
  \bibinfo{pages}{1320--1329}.
\newblock


\bibitem[\protect\citeauthoryear{{Connor W. Coley}, {Wengong Jin}, {Luke
  Rogers}, {Timothy F. Jamison}, {Tommi S. Jaakkola}, {William H. Green},
  {Regina Barzilay}, and {Klavs F. Jensen}}{{Connor W. Coley}
  et~al\mbox{.}}{2019}]%
        {202}
\bibfield{author}{\bibinfo{person}{{Connor W. Coley}},
  \bibinfo{person}{{Wengong Jin}}, \bibinfo{person}{{Luke Rogers}},
  \bibinfo{person}{{Timothy F. Jamison}}, \bibinfo{person}{{Tommi S.
  Jaakkola}}, \bibinfo{person}{{William H. Green}}, \bibinfo{person}{{Regina
  Barzilay}}, {and} \bibinfo{person}{{Klavs F. Jensen}}.}
  \bibinfo{year}{2019}\natexlab{}.
\newblock \showarticletitle{A graph-convolutional neural network model for the
  prediction of chemical reactivity}.
\newblock \bibinfo{journal}{\emph{Chemical Science}} \bibinfo{volume}{10},
  \bibinfo{number}{2} (\bibinfo{year}{2019}), \bibinfo{pages}{370--377}.
\newblock


\bibitem[\protect\citeauthoryear{{C\u\{a\}t\u\{a\}lina Cangea}, {Petar
  Veli\u\{c\}kovi\'\{c\}}, Jovanovi\'{c}, {Thomas N. Kipf}, and {Pietro
  Li\`{o}}}{{C\u\{a\}t\u\{a\}lina Cangea} et~al\mbox{.}}{2018}]%
        {96}
\bibfield{author}{\bibinfo{person}{{C\u\{a\}t\u\{a\}lina Cangea}},
  \bibinfo{person}{{Petar Veli\u\{c\}kovi\'\{c\}}}, \bibinfo{person}{Nikola
  Jovanovi\'{c}}, \bibinfo{person}{{Thomas N. Kipf}}, {and}
  \bibinfo{person}{{Pietro Li\`{o}}}.} \bibinfo{year}{2018}\natexlab{}.
\newblock \showarticletitle{Towards sparse hierarchical graph classifiers}.
\newblock  (\bibinfo{year}{2018}).
\newblock
\urldef\tempurl%
\url{https://arxiv.org/abs/1811.01287}
\showURL{%
\tempurl}


\bibitem[\protect\citeauthoryear{{Daixin Wang}, {Peng Cui}, and {Wenwu
  Zhu}}{{Daixin Wang} et~al\mbox{.}}{2016}]%
        {132}
\bibfield{author}{\bibinfo{person}{{Daixin Wang}}, \bibinfo{person}{{Peng
  Cui}}, {and} \bibinfo{person}{{Wenwu Zhu}}.} \bibinfo{year}{2016}\natexlab{}.
\newblock \showarticletitle{Structural Deep Network Embedding}. In
  \bibinfo{booktitle}{\emph{The {ACM} International Conference on Knowledge
  Discovery and Data Mining ({SIGKDD})}}. \bibinfo{pages}{1225--1234}.
\newblock


\bibitem[\protect\citeauthoryear{{Dan Busbridge}, {Dane Sherburn}, {Pietro
  Cavallo}, and {Nils Y. Yhammerla}}{{Dan Busbridge} et~al\mbox{.}}{2019}]%
        {361}
\bibfield{author}{\bibinfo{person}{{Dan Busbridge}}, \bibinfo{person}{{Dane
  Sherburn}}, \bibinfo{person}{{Pietro Cavallo}}, {and} \bibinfo{person}{{Nils
  Y. Yhammerla}}.} \bibinfo{year}{2019}\natexlab{}.
\newblock \showarticletitle{Relational graph attention networks}.
\newblock  (\bibinfo{year}{2019}).
\newblock
\urldef\tempurl%
\url{https://arxiv.org/abs/1904.05811}
\showURL{%
\tempurl}


\bibitem[\protect\citeauthoryear{{Daniel Beck}, {Gholamreza Haffari}, and
  {Trevor Cohn}}{{Daniel Beck} et~al\mbox{.}}{2018}]%
        {303}
\bibfield{author}{\bibinfo{person}{{Daniel Beck}}, \bibinfo{person}{{Gholamreza
  Haffari}}, {and} \bibinfo{person}{{Trevor Cohn}}.}
  \bibinfo{year}{2018}\natexlab{}.
\newblock \showarticletitle{Graph-to-Sequence learning using gated graph neural
  networks}. In \bibinfo{booktitle}{\emph{The Annual Meeting of the Association
  for Computational Linguistics}}. \bibinfo{pages}{273--283}.
\newblock


\bibitem[\protect\citeauthoryear{{Daniel Z\"{u}gner} and {Stephan
  G\"{u}nnemann}}{{Daniel Z\"{u}gner} and {Stephan G\"{u}nnemann}}{2019a}]%
        {183}
\bibfield{author}{\bibinfo{person}{{Daniel Z\"{u}gner}} {and}
  \bibinfo{person}{{Stephan G\"{u}nnemann}}.} \bibinfo{year}{2019}\natexlab{a}.
\newblock \showarticletitle{Adversarial attacks on graph neural networks via
  meta learning}. In \bibinfo{booktitle}{\emph{The International Conference on
  Learning Representations ({ICLR})}}.
\newblock


\bibitem[\protect\citeauthoryear{{Daniel Z\"{u}gner} and {Stephan
  G\"{u}nnemann}}{{Daniel Z\"{u}gner} and {Stephan G\"{u}nnemann}}{2019b}]%
        {177}
\bibfield{author}{\bibinfo{person}{{Daniel Z\"{u}gner}} {and}
  \bibinfo{person}{{Stephan G\"{u}nnemann}}.} \bibinfo{year}{2019}\natexlab{b}.
\newblock \showarticletitle{Certifiable robustness and robust training for
  graph convolutional networks}. In \bibinfo{booktitle}{\emph{The International
  Conference on Knowledge Discovery and Data Mining ({SIGKDD})}}.
  \bibinfo{pages}{246--256}.
\newblock


\bibitem[\protect\citeauthoryear{{David Duvenaud}, {Dougal Maclaurin}, {Jorge
  Aguilera-Iparraguirre}, {Rafael Gomez-Bombarelli}, {Timothy Hirzel}, {Alan
  Aspuru-Guzik}, and {Ryan P. Adams}}{{David Duvenaud} et~al\mbox{.}}{2015}]%
        {103}
\bibfield{author}{\bibinfo{person}{{David Duvenaud}}, \bibinfo{person}{{Dougal
  Maclaurin}}, \bibinfo{person}{{Jorge Aguilera-Iparraguirre}},
  \bibinfo{person}{{Rafael Gomez-Bombarelli}}, \bibinfo{person}{{Timothy
  Hirzel}}, \bibinfo{person}{{Alan Aspuru-Guzik}}, {and} \bibinfo{person}{{Ryan
  P. Adams}}.} \bibinfo{year}{2015}\natexlab{}.
\newblock \showarticletitle{Convolutional networks on graphs for learning
  molecular fingerprints}. In \bibinfo{booktitle}{\emph{The International
  Conference on Neural Information Processing Systems}}.
  \bibinfo{pages}{2224--2232}.
\newblock


\bibitem[\protect\citeauthoryear{{David I. Shuman}, {Sunil K. Narang}, {Pascal
  Frossard}, {Antonio Ortega}, and {Pierre Vandergheynst}}{{David I. Shuman}
  et~al\mbox{.}}{2013}]%
        {5}
\bibfield{author}{\bibinfo{person}{{David I. Shuman}}, \bibinfo{person}{{Sunil
  K. Narang}}, \bibinfo{person}{{Pascal Frossard}}, \bibinfo{person}{{Antonio
  Ortega}}, {and} \bibinfo{person}{{Pierre Vandergheynst}}.}
  \bibinfo{year}{2013}\natexlab{}.
\newblock \showarticletitle{The emerging field of signal processing on graphs:
  extending high-dimensional data analysis to networks and other irregular
  domains}.
\newblock \bibinfo{journal}{\emph{{IEEE} Signal Processing Magazine}}
  \bibinfo{volume}{30}, \bibinfo{number}{3} (\bibinfo{year}{2013}),
  \bibinfo{pages}{83--98}.
\newblock


\bibitem[\protect\citeauthoryear{{David K. Hammond}, {Pierre Vandergheynst},
  and {R\'{e}mi Gribonval}}{{David K. Hammond} et~al\mbox{.}}{2011}]%
        {189}
\bibfield{author}{\bibinfo{person}{{David K. Hammond}},
  \bibinfo{person}{{Pierre Vandergheynst}}, {and} \bibinfo{person}{{R\'{e}mi
  Gribonval}}.} \bibinfo{year}{2011}\natexlab{}.
\newblock \showarticletitle{Wavelets on graphs via spectral graph theory}.
\newblock \bibinfo{journal}{\emph{Applied and Computational Harmonic Analysis}}
  \bibinfo{volume}{30}, \bibinfo{number}{2} (\bibinfo{year}{2011}),
  \bibinfo{pages}{129--150}.
\newblock


\bibitem[\protect\citeauthoryear{{David M. Blei}, {Alp Kucukelbir}, and {Jon D.
  McAuliffe}}{{David M. Blei} et~al\mbox{.}}{2017}]%
        {151}
\bibfield{author}{\bibinfo{person}{{David M. Blei}}, \bibinfo{person}{{Alp
  Kucukelbir}}, {and} \bibinfo{person}{{Jon D. McAuliffe}}.}
  \bibinfo{year}{2017}\natexlab{}.
\newblock \showarticletitle{Variational inference: a review for statisticians}.
\newblock \bibinfo{journal}{\emph{Journal of the Americian Statistical
  Association}} \bibinfo{volume}{112}, \bibinfo{number}{518}
  (\bibinfo{year}{2017}), \bibinfo{pages}{859--877}.
\newblock


\bibitem[\protect\citeauthoryear{{Dingyuan Zhu}, {Peng Cui}, {Daixin Wang}, and
  {Wenwu Zhu}}{{Dingyuan Zhu} et~al\mbox{.}}{2018}]%
        {133}
\bibfield{author}{\bibinfo{person}{{Dingyuan Zhu}}, \bibinfo{person}{{Peng
  Cui}}, \bibinfo{person}{{Daixin Wang}}, {and} \bibinfo{person}{{Wenwu Zhu}}.}
  \bibinfo{year}{2018}\natexlab{}.
\newblock \showarticletitle{Deep Variational Network Embedding in Wasserstein
  Space}. In \bibinfo{booktitle}{\emph{The {ACM} International Conference on
  Knowledge Discovery and Data Mining ({SIGKDD})}}.
  \bibinfo{pages}{2827--2836}.
\newblock


\bibitem[\protect\citeauthoryear{{Dongmian Zou} and {Gilad Lerman}}{{Dongmian
  Zou} and {Gilad Lerman}}{2019}]%
        {190}
\bibfield{author}{\bibinfo{person}{{Dongmian Zou}} {and}
  \bibinfo{person}{{Gilad Lerman}}.} \bibinfo{year}{2019}\natexlab{}.
\newblock \showarticletitle{Graph convolutional neural networks via
  scattering}.
\newblock \bibinfo{journal}{\emph{Applied and Computational Harmonic Analysis}}
  (\bibinfo{year}{2019}).
\newblock


\bibitem[\protect\citeauthoryear{{Dzmitry Bahdanau}, {Kyunghyun Cho}, and
  {Yoshua Bengio}}{{Dzmitry Bahdanau} et~al\mbox{.}}{2015}]%
        {306}
\bibfield{author}{\bibinfo{person}{{Dzmitry Bahdanau}},
  \bibinfo{person}{{Kyunghyun Cho}}, {and} \bibinfo{person}{{Yoshua Bengio}}.}
  \bibinfo{year}{2015}\natexlab{}.
\newblock \showarticletitle{Neural machine translation by jointly learning to
  align and translate}. In \bibinfo{booktitle}{\emph{The International
  Conference on Learning Representations ({ICLR})}}.
\newblock


\bibitem[\protect\citeauthoryear{{Edouard Pineau} and {Nathan de
  Lara}}{{Edouard Pineau} and {Nathan de Lara}}{2019}]%
        {111}
\bibfield{author}{\bibinfo{person}{{Edouard Pineau}} {and}
  \bibinfo{person}{{Nathan de Lara}}.} \bibinfo{year}{2019}\natexlab{}.
\newblock \showarticletitle{Variational recurrent neural networks for graph
  classification}.
\newblock  (\bibinfo{year}{2019}).
\newblock
\urldef\tempurl%
\url{https://arxiv.org/abs/1902.02721}
\showURL{%
\tempurl}


\bibitem[\protect\citeauthoryear{{Ehsan Hajiramezanali}, {Arman Hasanzadeh},
  {Krishna Narayanan}, {Nick Duffield}, {Mingyuan Zhou}, and {Xiaoning
  Qian}}{{Ehsan Hajiramezanali} et~al\mbox{.}}{2019}]%
        {113}
\bibfield{author}{\bibinfo{person}{{Ehsan Hajiramezanali}},
  \bibinfo{person}{{Arman Hasanzadeh}}, \bibinfo{person}{{Krishna Narayanan}},
  \bibinfo{person}{{Nick Duffield}}, \bibinfo{person}{{Mingyuan Zhou}}, {and}
  \bibinfo{person}{{Xiaoning Qian}}.} \bibinfo{year}{2019}\natexlab{}.
\newblock \showarticletitle{Variational Graph Recurrent Neural Networks}. In
  \bibinfo{booktitle}{\emph{The International Conference on Neural Information
  Processing Systems}}. \bibinfo{pages}{10701--10711}.
\newblock


\bibitem[\protect\citeauthoryear{{Emanuele Rossi}, {Federico Monti}, {Michael
  Bronstein}, and {Pietro Li\`{o}}}{{Emanuele Rossi} et~al\mbox{.}}{2019}]%
        {211}
\bibfield{author}{\bibinfo{person}{{Emanuele Rossi}},
  \bibinfo{person}{{Federico Monti}}, \bibinfo{person}{{Michael Bronstein}},
  {and} \bibinfo{person}{{Pietro Li\`{o}}}.} \bibinfo{year}{2019}\natexlab{}.
\newblock \showarticletitle{{ncRNA} classification with graph convolutional
  networks}.
\newblock  (\bibinfo{year}{2019}).
\newblock
\urldef\tempurl%
\url{https://arxiv.org/abs/1905.06515}
\showURL{%
\tempurl}


\bibitem[\protect\citeauthoryear{{Federico Baldassarre} and {Hossein
  Azizpour}}{{Federico Baldassarre} and {Hossein Azizpour}}{2019}]%
        {165}
\bibfield{author}{\bibinfo{person}{{Federico Baldassarre}} {and}
  \bibinfo{person}{{Hossein Azizpour}}.} \bibinfo{year}{2019}\natexlab{}.
\newblock \showarticletitle{Explainability techniques for graph convolutional
  networks}.
\newblock  (\bibinfo{year}{2019}).
\newblock
\urldef\tempurl%
\url{https://arxiv.org/abs/1905.13686}
\showURL{%
\tempurl}


\bibitem[\protect\citeauthoryear{{Federico Monti}, {Davide Boscaini}, {Jonathan
  Masci}, {Emanuele Rodol\`{a}}, {Jan Svoboda}, and {Michael M.
  Bronstein}}{{Federico Monti} et~al\mbox{.}}{2017a}]%
        {18}
\bibfield{author}{\bibinfo{person}{{Federico Monti}}, \bibinfo{person}{{Davide
  Boscaini}}, \bibinfo{person}{{Jonathan Masci}}, \bibinfo{person}{{Emanuele
  Rodol\`{a}}}, \bibinfo{person}{{Jan Svoboda}}, {and}
  \bibinfo{person}{{Michael M. Bronstein}}.} \bibinfo{year}{2017}\natexlab{a}.
\newblock \showarticletitle{Geometric deep learning on graphs and manifolds
  using mixture model {CNNs}}. In \bibinfo{booktitle}{\emph{The {IEEE}
  Conference on Computer Vision and Pattern Recognition ({CVPR})}}.
  \bibinfo{pages}{5425--5434}.
\newblock


\bibitem[\protect\citeauthoryear{{Federico Monti}, {Michael M. Bronstein}, and
  {Xavier Bresson}}{{Federico Monti} et~al\mbox{.}}{2017b}]%
        {112}
\bibfield{author}{\bibinfo{person}{{Federico Monti}}, \bibinfo{person}{{Michael
  M. Bronstein}}, {and} \bibinfo{person}{{Xavier Bresson}}.}
  \bibinfo{year}{2017}\natexlab{b}.
\newblock \showarticletitle{Geometric matrix completion with recurrent
  multi-graph neural network}. In \bibinfo{booktitle}{\emph{The International
  Conference on Neural Information Processing Systems}}.
  \bibinfo{pages}{3697--3707}.
\newblock


\bibitem[\protect\citeauthoryear{{Felipe Petroski Such}, {Shagan Sah}, {Miguel
  Dominguez}, {Suhas Pillai}, {Chao Zhang}, {Andrew Michael}, {Nathan D.
  Cahill}, and {Raymond Ptucha}}{{Felipe Petroski Such} et~al\mbox{.}}{2017}]%
        {32}
\bibfield{author}{\bibinfo{person}{{Felipe Petroski Such}},
  \bibinfo{person}{{Shagan Sah}}, \bibinfo{person}{{Miguel Dominguez}},
  \bibinfo{person}{{Suhas Pillai}}, \bibinfo{person}{{Chao Zhang}},
  \bibinfo{person}{{Andrew Michael}}, \bibinfo{person}{{Nathan D. Cahill}},
  {and} \bibinfo{person}{{Raymond Ptucha}}.} \bibinfo{year}{2017}\natexlab{}.
\newblock \showarticletitle{Robust spatial filtering with graph convolutional
  neural networks}.
\newblock \bibinfo{journal}{\emph{{IEEE} Journal of Selected Topics in Signal
  Processing}} \bibinfo{volume}{11}, \bibinfo{number}{6}
  (\bibinfo{year}{2017}), \bibinfo{pages}{884--896}.
\newblock


\bibitem[\protect\citeauthoryear{{Felix Wu}, {Tianyi Zhang}, {Amauri Holanda de
  Souza}, {Christopher Fifty}, {Tao Yu}, and {Kilian Q. Weinberger}}{{Felix Wu}
  et~al\mbox{.}}{2019}]%
        {22}
\bibfield{author}{\bibinfo{person}{{Felix Wu}}, \bibinfo{person}{{Tianyi
  Zhang}}, \bibinfo{person}{{Amauri Holanda de Souza}},
  \bibinfo{person}{{Christopher Fifty}}, \bibinfo{person}{{Tao Yu}}, {and}
  \bibinfo{person}{{Kilian Q. Weinberger}}.} \bibinfo{year}{2019}\natexlab{}.
\newblock \showarticletitle{Simplifying Graph Convolutional Networks}. In
  \bibinfo{booktitle}{\emph{The International Conference on Machine Learning
  ({ICML})}}. \bibinfo{pages}{6861--6871}.
\newblock


\bibitem[\protect\citeauthoryear{{Fenyu Hu}, {Yanqiao Zhu}, {Shu Wu}, {Liang
  Wang}, and {Tieniu Tan}}{{Fenyu Hu} et~al\mbox{.}}{2019}]%
        {42}
\bibfield{author}{\bibinfo{person}{{Fenyu Hu}}, \bibinfo{person}{{Yanqiao
  Zhu}}, \bibinfo{person}{{Shu Wu}}, \bibinfo{person}{{Liang Wang}}, {and}
  \bibinfo{person}{{Tieniu Tan}}.} \bibinfo{year}{2019}\natexlab{}.
\newblock \showarticletitle{Semi-supervised node classification via
  hierarchical graph convolutional networks}.
\newblock  (\bibinfo{year}{2019}).
\newblock
\urldef\tempurl%
\url{https://arxiv.org/abs/1902.06667v2}
\showURL{%
\tempurl}


\bibitem[\protect\citeauthoryear{{Fisher Yu} and {Vladlen Koltun}}{{Fisher Yu}
  and {Vladlen Koltun}}{2016}]%
        {384}
\bibfield{author}{\bibinfo{person}{{Fisher Yu}} {and} \bibinfo{person}{{Vladlen
  Koltun}}.} \bibinfo{year}{2016}\natexlab{}.
\newblock \showarticletitle{Multi-scale context aggregation by dilated
  convolutions}.
\newblock  (\bibinfo{year}{2016}).
\newblock
\urldef\tempurl%
\url{https://arxiv.org/abs/1511.07122v3}
\showURL{%
\tempurl}


\bibitem[\protect\citeauthoryear{{Franco Manessi}, {Alessandro Rozza}, and
  {Mario Manzo}}{{Franco Manessi} et~al\mbox{.}}{2020}]%
        {105}
\bibfield{author}{\bibinfo{person}{{Franco Manessi}},
  \bibinfo{person}{{Alessandro Rozza}}, {and} \bibinfo{person}{{Mario Manzo}}.}
  \bibinfo{year}{2020}\natexlab{}.
\newblock \showarticletitle{Dynamic graph convolutional networks}.
\newblock \bibinfo{journal}{\emph{Pattern Recognition}} \bibinfo{volume}{97},
  \bibinfo{number}{2020} (\bibinfo{year}{2020}), \bibinfo{pages}{No. 107000}.
\newblock


\bibitem[\protect\citeauthoryear{{Franco Scarselli}, {Marco Gori}, {Ah Chung
  Tsoi}, {Markus Hagenbuchner}, and {Gabriele Monfardini}}{{Franco Scarselli}
  et~al\mbox{.}}{2008}]%
        {380}
\bibfield{author}{\bibinfo{person}{{Franco Scarselli}}, \bibinfo{person}{{Marco
  Gori}}, \bibinfo{person}{{Ah Chung Tsoi}}, \bibinfo{person}{{Markus
  Hagenbuchner}}, {and} \bibinfo{person}{{Gabriele Monfardini}}.}
  \bibinfo{year}{2008}\natexlab{}.
\newblock \showarticletitle{The graph neural network model}.
\newblock \bibinfo{journal}{\emph{{IEEE} Transactions on Neural Networks}}
  \bibinfo{volume}{20}, \bibinfo{number}{1} (\bibinfo{year}{2008}),
  \bibinfo{pages}{61--80}.
\newblock


\bibitem[\protect\citeauthoryear{{Franco Scarselli}, {Marco Gori}, {Ah Chung
  Tsoi}, {Markus Hagenbuchner}, and {Gabriele Monfardini}}{{Franco Scarselli}
  et~al\mbox{.}}{2009}]%
        {77}
\bibfield{author}{\bibinfo{person}{{Franco Scarselli}}, \bibinfo{person}{{Marco
  Gori}}, \bibinfo{person}{{Ah Chung Tsoi}}, \bibinfo{person}{{Markus
  Hagenbuchner}}, {and} \bibinfo{person}{{Gabriele Monfardini}}.}
  \bibinfo{year}{2009}\natexlab{}.
\newblock \showarticletitle{Computational Capabilities of Graph Neural
  Networks}.
\newblock \bibinfo{journal}{\emph{{IEEE} Transactions on Neural Networks}}
  \bibinfo{volume}{20}, \bibinfo{number}{1} (\bibinfo{year}{2009}),
  \bibinfo{pages}{81--102}.
\newblock


\bibitem[\protect\citeauthoryear{{Gao Huang}, {Zhuang Liu}, {Laurens Van Der
  Maaten}, and {Kilian Q. Weinberger}}{{Gao Huang} et~al\mbox{.}}{2017}]%
        {383}
\bibfield{author}{\bibinfo{person}{{Gao Huang}}, \bibinfo{person}{{Zhuang
  Liu}}, \bibinfo{person}{{Laurens Van Der Maaten}}, {and}
  \bibinfo{person}{{Kilian Q. Weinberger}}.} \bibinfo{year}{2017}\natexlab{}.
\newblock \showarticletitle{Densely connected convolutional networks}. In
  \bibinfo{booktitle}{\emph{The IEEE International Conference on Computer
  Vision and Pattern Recognition (CVPR)}}. \bibinfo{pages}{2261--2269}.
\newblock


\bibitem[\protect\citeauthoryear{Gori, Monfardini, and Scarselli}{Gori
  et~al\mbox{.}}{2005}]%
        {76}
\bibfield{author}{\bibinfo{person}{Marco Gori}, \bibinfo{person}{Gabriele
  Monfardini}, {and} \bibinfo{person}{Franco Scarselli}.}
  \bibinfo{year}{2005}\natexlab{}.
\newblock \showarticletitle{A new model for learning in graph domains}. In
  \bibinfo{booktitle}{\emph{The International Joint Conference on Neural
  Networks (IJCNN)}}. \bibinfo{pages}{729--734}.
\newblock


\bibitem[\protect\citeauthoryear{{Guohao Li}, {Matthias M\"{u}ller}, {Ali
  Thabet}, and {Bernard Ghanem}}{{Guohao Li} et~al\mbox{.}}{2019}]%
        {13}
\bibfield{author}{\bibinfo{person}{{Guohao Li}}, \bibinfo{person}{{Matthias
  M\"{u}ller}}, \bibinfo{person}{{Ali Thabet}}, {and} \bibinfo{person}{{Bernard
  Ghanem}}.} \bibinfo{year}{2019}\natexlab{}.
\newblock \showarticletitle{{DeepGCNs}: can {GCNs} go as deep as {CNNs}}. In
  \bibinfo{booktitle}{\emph{The {IEEE} International Conference on Computer
  Vision ({ICCV})}}. \bibinfo{pages}{9267--9276}.
\newblock


\bibitem[\protect\citeauthoryear{{Haggai Maron}, {Heli Ben-Hamu}, {Nadav
  Shamir}, and {Yaron Lipman}}{{Haggai Maron} et~al\mbox{.}}{2019}]%
        {389}
\bibfield{author}{\bibinfo{person}{{Haggai Maron}}, \bibinfo{person}{{Heli
  Ben-Hamu}}, \bibinfo{person}{{Nadav Shamir}}, {and} \bibinfo{person}{{Yaron
  Lipman}}.} \bibinfo{year}{2019}\natexlab{}.
\newblock \showarticletitle{Invariant and equivariant graph networks}. In
  \bibinfo{booktitle}{\emph{The International Conference on Learning
  Representations ({ICLR})}}.
\newblock


\bibitem[\protect\citeauthoryear{{Hanjun Dai}, {Hui Li}, {Tian Tian}, {Xin
  Huang}, {Lin Wang}, {Jun Zhu}, and {Le Song}}{{Hanjun Dai}
  et~al\mbox{.}}{2018}]%
        {182}
\bibfield{author}{\bibinfo{person}{{Hanjun Dai}}, \bibinfo{person}{{Hui Li}},
  \bibinfo{person}{{Tian Tian}}, \bibinfo{person}{{Xin Huang}},
  \bibinfo{person}{{Lin Wang}}, \bibinfo{person}{{Jun Zhu}}, {and}
  \bibinfo{person}{{Le Song}}.} \bibinfo{year}{2018}\natexlab{}.
\newblock \showarticletitle{Adversarial Attack on Graph Structured Data}. In
  \bibinfo{booktitle}{\emph{The International Conference on Machine Learning
  ({ICML})}}. \bibinfo{pages}{1115--1124}.
\newblock


\bibitem[\protect\citeauthoryear{{Hao Yuan} and {Shuiwang Ji}}{{Hao Yuan} and
  {Shuiwang Ji}}{2020}]%
        {390}
\bibfield{author}{\bibinfo{person}{{Hao Yuan}} {and} \bibinfo{person}{{Shuiwang
  Ji}}.} \bibinfo{year}{2020}\natexlab{}.
\newblock \showarticletitle{{StructPool}: structured graph pooling via
  conditional random fields}. In \bibinfo{booktitle}{\emph{The International
  Conference on Learning Representations ({ICLR})}}.
\newblock


\bibitem[\protect\citeauthoryear{{Heng Chang}, {Yu Rong}, {Tingyang Xu},
  {Wenbing Huang}, {Somayeh Sojoudi}, {Junzhou Huang}, and {Wenwu Zhu}}{{Heng
  Chang} et~al\mbox{.}}{2020}]%
        {362}
\bibfield{author}{\bibinfo{person}{{Heng Chang}}, \bibinfo{person}{{Yu Rong}},
  \bibinfo{person}{{Tingyang Xu}}, \bibinfo{person}{{Wenbing Huang}},
  \bibinfo{person}{{Somayeh Sojoudi}}, \bibinfo{person}{{Junzhou Huang}}, {and}
  \bibinfo{person}{{Wenwu Zhu}}.} \bibinfo{year}{2020}\natexlab{}.
\newblock \showarticletitle{Spectral graph attention network}.
\newblock  (\bibinfo{year}{2020}).
\newblock
\urldef\tempurl%
\url{https://arxiv.org/abs/2003.07450}
\showURL{%
\tempurl}


\bibitem[\protect\citeauthoryear{{Hongbin Pei}, {Bingzhe Wei}, {Kevin
  Chen-Chuan Chang}, {Yu Lei}, and {Bo Yang}}{{Hongbin Pei}
  et~al\mbox{.}}{2020}]%
        {17}
\bibfield{author}{\bibinfo{person}{{Hongbin Pei}}, \bibinfo{person}{{Bingzhe
  Wei}}, \bibinfo{person}{{Kevin Chen-Chuan Chang}}, \bibinfo{person}{{Yu
  Lei}}, {and} \bibinfo{person}{{Bo Yang}}.} \bibinfo{year}{2020}\natexlab{}.
\newblock \showarticletitle{Geom-{GCN}: geometric graph convolutional
  networks}. In \bibinfo{booktitle}{\emph{The International Conference on
  Learning Representations ({ICLR})}}.
\newblock


\bibitem[\protect\citeauthoryear{{Hongchang Gao}, {Jian Pei}, and {Heng
  Huang}}{{Hongchang Gao} et~al\mbox{.}}{2019}]%
        {162}
\bibfield{author}{\bibinfo{person}{{Hongchang Gao}}, \bibinfo{person}{{Jian
  Pei}}, {and} \bibinfo{person}{{Heng Huang}}.}
  \bibinfo{year}{2019}\natexlab{}.
\newblock \showarticletitle{Conditional Random Field Enhanced Graph
  Convolutional Neural Networks}. In \bibinfo{booktitle}{\emph{The {ACM}
  International Conference on Knowledge Discovery and Data Mining ({SIGKDD})}}.
  \bibinfo{pages}{276--284}.
\newblock


\bibitem[\protect\citeauthoryear{{Hongyang Gao} and {Shuiwang Ji}}{{Hongyang
  Gao} and {Shuiwang Ji}}{2019}]%
        {95}
\bibfield{author}{\bibinfo{person}{{Hongyang Gao}} {and}
  \bibinfo{person}{{Shuiwang Ji}}.} \bibinfo{year}{2019}\natexlab{}.
\newblock \showarticletitle{Graph U-Nets}.
\newblock  (\bibinfo{year}{2019}).
\newblock
\urldef\tempurl%
\url{https://arxiv.org/abs/1905.05178}
\showURL{%
\tempurl}


\bibitem[\protect\citeauthoryear{{Hongyang Gao}, {Zhengyang Wang}, and
  {Shuiwang Ji}}{{Hongyang Gao} et~al\mbox{.}}{2018}]%
        {25}
\bibfield{author}{\bibinfo{person}{{Hongyang Gao}}, \bibinfo{person}{{Zhengyang
  Wang}}, {and} \bibinfo{person}{{Shuiwang Ji}}.}
  \bibinfo{year}{2018}\natexlab{}.
\newblock \showarticletitle{Large-scale learnable graph convolutional
  networks}. In \bibinfo{booktitle}{\emph{The {ACM} International Conference on
  Knowledge Discovery and Data Mining ({SIGKDD})}}.
  \bibinfo{pages}{1415--1424}.
\newblock


\bibitem[\protect\citeauthoryear{{Hyeoncheol Cho} and {Insung S.
  Choi}}{{Hyeoncheol Cho} and {Insung S. Choi}}{2018}]%
        {197}
\bibfield{author}{\bibinfo{person}{{Hyeoncheol Cho}} {and}
  \bibinfo{person}{{Insung S. Choi}}.} \bibinfo{year}{2018}\natexlab{}.
\newblock \showarticletitle{Three-dimensionally embedded graph convolutional
  network ({3DGCN}) for molecule interpretation}.
\newblock  (\bibinfo{year}{2018}).
\newblock
\urldef\tempurl%
\url{https://arxiv.org/abs/1811.09794}
\showURL{%
\tempurl}


\bibitem[\protect\citeauthoryear{{Ian J. Goodfellow}, {Jean Pouget-Abadie},
  {Mehdi Mirza}, {Bing Xu}, {David Warde-Farley}, {Sherjil Ozair}, {Azron
  Courville}, and {Yoshua Bengio}}{{Ian J. Goodfellow} et~al\mbox{.}}{2014}]%
        {149}
\bibfield{author}{\bibinfo{person}{{Ian J. Goodfellow}}, \bibinfo{person}{{Jean
  Pouget-Abadie}}, \bibinfo{person}{{Mehdi Mirza}}, \bibinfo{person}{{Bing
  Xu}}, \bibinfo{person}{{David Warde-Farley}}, \bibinfo{person}{{Sherjil
  Ozair}}, \bibinfo{person}{{Azron Courville}}, {and} \bibinfo{person}{{Yoshua
  Bengio}}.} \bibinfo{year}{2014}\natexlab{}.
\newblock \showarticletitle{Generative Adversarial Nets}. In
  \bibinfo{booktitle}{\emph{The International Conference on Neural Inforamtion
  Processing Systems (NIPS)}}. \bibinfo{pages}{2672--2680}.
\newblock


\bibitem[\protect\citeauthoryear{{Inderjit S. Dhillon}, {Yuqiang Guan}, and
  {Brian Kulis}}{{Inderjit S. Dhillon} et~al\mbox{.}}{2007}]%
        {99}
\bibfield{author}{\bibinfo{person}{{Inderjit S. Dhillon}},
  \bibinfo{person}{{Yuqiang Guan}}, {and} \bibinfo{person}{{Brian Kulis}}.}
  \bibinfo{year}{2007}\natexlab{}.
\newblock \showarticletitle{Weighted graph cuts without eigenvectors: a
  multilevel approach}.
\newblock \bibinfo{journal}{\emph{{IEEE} Transactions on Pattern Analysis and
  Mchine Intelligence}} \bibinfo{volume}{29}, \bibinfo{number}{11}
  (\bibinfo{year}{2007}), \bibinfo{pages}{1944--1957}.
\newblock


\bibitem[\protect\citeauthoryear{{Ines Chami}, {Rex Ying}, {Christopher Re},
  and {Jure Leskovec}}{{Ines Chami} et~al\mbox{.}}{2019}]%
        {15}
\bibfield{author}{\bibinfo{person}{{Ines Chami}}, \bibinfo{person}{{Rex Ying}},
  \bibinfo{person}{{Christopher Re}}, {and} \bibinfo{person}{{Jure Leskovec}}.}
  \bibinfo{year}{2019}\natexlab{}.
\newblock \showarticletitle{Hyperbolic Graph Convolutional Neural Networks}. In
  \bibinfo{booktitle}{\emph{The International Conference on Neural Information
  Processing Systems ({NeurPS})}}. \bibinfo{pages}{4868--4879}.
\newblock


\bibitem[\protect\citeauthoryear{{James Atwood} and {Don Towsley}}{{James
  Atwood} and {Don Towsley}}{2016}]%
        {20}
\bibfield{author}{\bibinfo{person}{{James Atwood}} {and} \bibinfo{person}{{Don
  Towsley}}.} \bibinfo{year}{2016}\natexlab{}.
\newblock \showarticletitle{Diffusion-Convolutional Neural Network}. In
  \bibinfo{booktitle}{\emph{The International Conference on Neural Information
  Processing Systems ({NIPS})}}. \bibinfo{pages}{1993--2001}.
\newblock


\bibitem[\protect\citeauthoryear{{Jeremy Kawahara}, {Colin J. Brown}, {Steven
  P. Miller}, {Brian G. Booth}, {Vann Chau}, {Ruth E. Grunau}, {Jill G.
  Zwicker}, and {Ghassan Hamarneh}}{{Jeremy Kawahara} et~al\mbox{.}}{2017}]%
        {206}
\bibfield{author}{\bibinfo{person}{{Jeremy Kawahara}}, \bibinfo{person}{{Colin
  J. Brown}}, \bibinfo{person}{{Steven P. Miller}}, \bibinfo{person}{{Brian G.
  Booth}}, \bibinfo{person}{{Vann Chau}}, \bibinfo{person}{{Ruth E. Grunau}},
  \bibinfo{person}{{Jill G. Zwicker}}, {and} \bibinfo{person}{{Ghassan
  Hamarneh}}.} \bibinfo{year}{2017}\natexlab{}.
\newblock \showarticletitle{{BrainNetCNN}: convolutional neural networks for
  brain networks; towards predicting neurodevelopment}.
\newblock \bibinfo{journal}{\emph{{NeuroImage}}}  \bibinfo{volume}{146}
  (\bibinfo{year}{2017}), \bibinfo{pages}{1038--1049}.
\newblock


\bibitem[\protect\citeauthoryear{{Jessica Schrouff}, {Kai Wohlfahrt}, {Bruno
  Marnette}, and {Liam Atkinson}}{{Jessica Schrouff} et~al\mbox{.}}{2019}]%
        {342}
\bibfield{author}{\bibinfo{person}{{Jessica Schrouff}}, \bibinfo{person}{{Kai
  Wohlfahrt}}, \bibinfo{person}{{Bruno Marnette}}, {and} \bibinfo{person}{{Liam
  Atkinson}}.} \bibinfo{year}{2019}\natexlab{}.
\newblock \showarticletitle{Inferring Javascript types using graph neural
  networks}.
\newblock  (\bibinfo{year}{2019}).
\newblock
\urldef\tempurl%
\url{https://arxiv.org/abs/1905.06707}
\showURL{%
\tempurl}


\bibitem[\protect\citeauthoryear{{Jian Du}, {Shanghang Zhang}, {Guanhang Wu},
  {Jos\'{e} M.F. Moura}, and {Soummya Kar}}{{Jian Du} et~al\mbox{.}}{2018}]%
        {38}
\bibfield{author}{\bibinfo{person}{{Jian Du}}, \bibinfo{person}{{Shanghang
  Zhang}}, \bibinfo{person}{{Guanhang Wu}}, \bibinfo{person}{{Jos\'{e} M.F.
  Moura}}, {and} \bibinfo{person}{{Soummya Kar}}.}
  \bibinfo{year}{2018}\natexlab{}.
\newblock \showarticletitle{Topology adaptive graph convolutional networks}.
\newblock  (\bibinfo{year}{2018}).
\newblock
\urldef\tempurl%
\url{https://arxiv.org/abs/1710.10370}
\showURL{%
\tempurl}


\bibitem[\protect\citeauthoryear{{Jianfei Chen}, {Jun Zhu}, and {Le
  Song}}{{Jianfei Chen} et~al\mbox{.}}{2018}]%
        {11}
\bibfield{author}{\bibinfo{person}{{Jianfei Chen}}, \bibinfo{person}{{Jun
  Zhu}}, {and} \bibinfo{person}{{Le Song}}.} \bibinfo{year}{2018}\natexlab{}.
\newblock \showarticletitle{Stochastic training of graph convolutional networks
  with variance reduction}. In \bibinfo{booktitle}{\emph{The International
  Conference on Machine Learning ({ICML})}}. \bibinfo{pages}{942--950}.
\newblock


\bibitem[\protect\citeauthoryear{{Jiani Zhang}, {Xingjian Shi}, {Junyuan Xie},
  {Hao Ma}, {Irwin King}, and {Dit-Yan Yeung}}{{Jiani Zhang}
  et~al\mbox{.}}{2018}]%
        {74}
\bibfield{author}{\bibinfo{person}{{Jiani Zhang}}, \bibinfo{person}{{Xingjian
  Shi}}, \bibinfo{person}{{Junyuan Xie}}, \bibinfo{person}{{Hao Ma}},
  \bibinfo{person}{{Irwin King}}, {and} \bibinfo{person}{{Dit-Yan Yeung}}.}
  \bibinfo{year}{2018}\natexlab{}.
\newblock \showarticletitle{{GaAN}: gated attention networks for learning on
  large and spatiotemporal graphs}. In \bibinfo{booktitle}{\emph{The
  International Conference on Uncertainty in Artificial Intelligence ({UAI})}}.
  \bibinfo{pages}{No. 139}.
\newblock


\bibitem[\protect\citeauthoryear{{Jiawei Zhang}, {Haopeng Zhang}, {Congying
  Xia}, and {Li Sun}}{{Jiawei Zhang} et~al\mbox{.}}{2020}]%
        {367}
\bibfield{author}{\bibinfo{person}{{Jiawei Zhang}}, \bibinfo{person}{{Haopeng
  Zhang}}, \bibinfo{person}{{Congying Xia}}, {and} \bibinfo{person}{{Li Sun}}.}
  \bibinfo{year}{2020}\natexlab{}.
\newblock \showarticletitle{{Graph-Bert}: only attention is needed for learning
  graph representations}.
\newblock  (\bibinfo{year}{2020}).
\newblock
\urldef\tempurl%
\url{https://arxiv.org/abs/2001.05140}
\showURL{%
\tempurl}


\bibitem[\protect\citeauthoryear{{Jiaxuan You}, {Haoze Wu}, {Clark Barrett},
  {Raghuram Ramanujan}, and {Jure Leskovec}}{{Jiaxuan You}
  et~al\mbox{.}}{2019}]%
        {224}
\bibfield{author}{\bibinfo{person}{{Jiaxuan You}}, \bibinfo{person}{{Haoze
  Wu}}, \bibinfo{person}{{Clark Barrett}}, \bibinfo{person}{{Raghuram
  Ramanujan}}, {and} \bibinfo{person}{{Jure Leskovec}}.}
  \bibinfo{year}{2019}\natexlab{}.
\newblock \showarticletitle{G2SAT: learning to generate {SAT} formulas}. In
  \bibinfo{booktitle}{\emph{The International Conference on Neural Information
  Processing Systems ({NeruPS})}}. \bibinfo{pages}{10552--10563}.
\newblock


\bibitem[\protect\citeauthoryear{{Jiaxuan You}, {Rex Ying}, {Xiang Ren},
  {William L. Hamilton}, and {Jure Leskovec}}{{Jiaxuan You}
  et~al\mbox{.}}{2018}]%
        {152}
\bibfield{author}{\bibinfo{person}{{Jiaxuan You}}, \bibinfo{person}{{Rex
  Ying}}, \bibinfo{person}{{Xiang Ren}}, \bibinfo{person}{{William L.
  Hamilton}}, {and} \bibinfo{person}{{Jure Leskovec}}.}
  \bibinfo{year}{2018}\natexlab{}.
\newblock \showarticletitle{{GraphRNN}: generating realistic graphs with deep
  auto-regressive models}. In \bibinfo{booktitle}{\emph{The Internatinonal
  Conference on Machine Learing ({ICML})}}. \bibinfo{pages}{5708--5717}.
\newblock


\bibitem[\protect\citeauthoryear{{Jiaxue You}, {Rex Ying}, and {Jure
  Leskovec}}{{Jiaxue You} et~al\mbox{.}}{2019}]%
        {46}
\bibfield{author}{\bibinfo{person}{{Jiaxue You}}, \bibinfo{person}{{Rex Ying}},
  {and} \bibinfo{person}{{Jure Leskovec}}.} \bibinfo{year}{2019}\natexlab{}.
\newblock \showarticletitle{Position-aware graph neural networks}. In
  \bibinfo{booktitle}{\emph{The International Conference on Machine Learning
  ({ICML})}}. \bibinfo{pages}{7134--7143}.
\newblock


\bibitem[\protect\citeauthoryear{{Jiayi Wei}, {Maruth Goyal}, {Greg Durrett},
  and {Isil Dilling}}{{Jiayi Wei} et~al\mbox{.}}{2020}]%
        {341}
\bibfield{author}{\bibinfo{person}{{Jiayi Wei}}, \bibinfo{person}{{Maruth
  Goyal}}, \bibinfo{person}{{Greg Durrett}}, {and} \bibinfo{person}{{Isil
  Dilling}}.} \bibinfo{year}{2020}\natexlab{}.
\newblock \showarticletitle{{LambdaNet}: probabilistic type inference using
  graph neural networks}. In \bibinfo{booktitle}{\emph{The International
  Conference on Learning Representations ({ICLR})}}.
\newblock


\bibitem[\protect\citeauthoryear{{Jie Chen}, {Tengfei Ma}, and {Cao Xiao}}{{Jie
  Chen} et~al\mbox{.}}{2018}]%
        {10}
\bibfield{author}{\bibinfo{person}{{Jie Chen}}, \bibinfo{person}{{Tengfei Ma}},
  {and} \bibinfo{person}{{Cao Xiao}}.} \bibinfo{year}{2018}\natexlab{}.
\newblock \showarticletitle{{FastGCN}: fast learning with graph convolutional
  networks via importance sampling}. In \bibinfo{booktitle}{\emph{The
  International Conference on Learning Representations ({ICLR})}}.
\newblock


\bibitem[\protect\citeauthoryear{{Jie Zhou}, {Gangqu Cui}, {Zhengyan Zhang},
  {Cheng Yang}, {Zhiyuan Liu}, and {Maosong Sun}}{{Jie Zhou}
  et~al\mbox{.}}{2018}]%
        {3}
\bibfield{author}{\bibinfo{person}{{Jie Zhou}}, \bibinfo{person}{{Gangqu Cui}},
  \bibinfo{person}{{Zhengyan Zhang}}, \bibinfo{person}{{Cheng Yang}},
  \bibinfo{person}{{Zhiyuan Liu}}, {and} \bibinfo{person}{{Maosong Sun}}.}
  \bibinfo{year}{2018}\natexlab{}.
\newblock \showarticletitle{Graph neural networks: a review of methods and
  applications}.
\newblock  (\bibinfo{year}{2018}).
\newblock
\urldef\tempurl%
\url{https://arxiv.org/abs/1812.08434}
\showURL{%
\tempurl}


\bibitem[\protect\citeauthoryear{{Jie Zhou}, {Xu Han}, {Cheng Yang}, {Zhiyuan
  Liu}, {Lifeng Wang}, {Changcheng Li}, and {Maosong Sun}}{{Jie Zhou}
  et~al\mbox{.}}{2019}]%
        {273}
\bibfield{author}{\bibinfo{person}{{Jie Zhou}}, \bibinfo{person}{{Xu Han}},
  \bibinfo{person}{{Cheng Yang}}, \bibinfo{person}{{Zhiyuan Liu}},
  \bibinfo{person}{{Lifeng Wang}}, \bibinfo{person}{{Changcheng Li}}, {and}
  \bibinfo{person}{{Maosong Sun}}.} \bibinfo{year}{2019}\natexlab{}.
\newblock \showarticletitle{{GEAR}: graph-based evidence aggregating and
  reasoning for fact verification}. In \bibinfo{booktitle}{\emph{The Annual
  Meeting of the Association for Computational Linguistics ({ACL})}}.
  \bibinfo{pages}{892--901}.
\newblock


\bibitem[\protect\citeauthoryear{{Jiezhong Qiu}, {Jian Tang}, {Hao Ma}, {Yuxiao
  Dong}, {Kuansan Wang}, and {Jie Tang}}{{Jiezhong Qiu} et~al\mbox{.}}{2018}]%
        {234}
\bibfield{author}{\bibinfo{person}{{Jiezhong Qiu}}, \bibinfo{person}{{Jian
  Tang}}, \bibinfo{person}{{Hao Ma}}, \bibinfo{person}{{Yuxiao Dong}},
  \bibinfo{person}{{Kuansan Wang}}, {and} \bibinfo{person}{{Jie Tang}}.}
  \bibinfo{year}{2018}\natexlab{}.
\newblock \showarticletitle{{DeepInf}: social influence prediction with deep
  learning}. In \bibinfo{booktitle}{\emph{The International Conference on
  Knowledge Discovery and Data Mining ({SIGKDD})}}.
  \bibinfo{pages}{2110--2119}.
\newblock


\bibitem[\protect\citeauthoryear{{Jiwoong Park}, {Minsik Lee}, {Hyung Jin
  Chang}, {Kyuewang Lee}, and Choi}{{Jiwoong Park} et~al\mbox{.}}{2019}]%
        {128}
\bibfield{author}{\bibinfo{person}{{Jiwoong Park}}, \bibinfo{person}{{Minsik
  Lee}}, \bibinfo{person}{{Hyung Jin Chang}}, \bibinfo{person}{{Kyuewang Lee}},
  {and} \bibinfo{person}{Jin~Young Choi}.} \bibinfo{year}{2019}\natexlab{}.
\newblock \showarticletitle{Symmetric graph convolutional autoencoder for
  unsupervised graph representation learning}. In \bibinfo{booktitle}{\emph{The
  International Conference on Computer Vision}}. \bibinfo{pages}{6519--6528}.
\newblock


\bibitem[\protect\citeauthoryear{{Joan Bruna}, {Wojciech Zaremba}, {Arthur
  Szlam}, and {Yann LeCun}}{{Joan Bruna} et~al\mbox{.}}{2014}]%
        {7}
\bibfield{author}{\bibinfo{person}{{Joan Bruna}}, \bibinfo{person}{{Wojciech
  Zaremba}}, \bibinfo{person}{{Arthur Szlam}}, {and} \bibinfo{person}{{Yann
  LeCun}}.} \bibinfo{year}{2014}\natexlab{}.
\newblock \showarticletitle{Spectral networks and locally connected networks on
  graphs}. In \bibinfo{booktitle}{\emph{The International Conference on
  Learning Representations ({ICLR})}}.
\newblock


\bibitem[\protect\citeauthoryear{{Johannes Klicpera}, {Aleksandar Bojchevski},
  and {Stephan G\"{u}nnemann}}{{Johannes Klicpera} et~al\mbox{.}}{2019}]%
        {233}
\bibfield{author}{\bibinfo{person}{{Johannes Klicpera}},
  \bibinfo{person}{{Aleksandar Bojchevski}}, {and} \bibinfo{person}{{Stephan
  G\"{u}nnemann}}.} \bibinfo{year}{2019}\natexlab{}.
\newblock \showarticletitle{Predict the propagate: graph neural networks meet
  personalized pagerank}. In \bibinfo{booktitle}{\emph{The International
  Conference on Learning Representations ({ICLR})}}.
\newblock


\bibitem[\protect\citeauthoryear{{John B. Lee}, {Ryan A. Rossi}, {Sungchul
  Kim}, {Nesreen K. Ahmed}, and {Eunyee Koh}}{{John B. Lee}
  et~al\mbox{.}}{2019}]%
        {394}
\bibfield{author}{\bibinfo{person}{{John B. Lee}}, \bibinfo{person}{{Ryan A.
  Rossi}}, \bibinfo{person}{{Sungchul Kim}}, \bibinfo{person}{{Nesreen K.
  Ahmed}}, {and} \bibinfo{person}{{Eunyee Koh}}.}
  \bibinfo{year}{2019}\natexlab{}.
\newblock \showarticletitle{Attention models in graphs: a survey}.
\newblock \bibinfo{journal}{\emph{{ACM} Transactions on Knowledge Discovery
  from Data}} \bibinfo{volume}{13}, \bibinfo{number}{6} (\bibinfo{year}{2019}),
  \bibinfo{pages}{No. 62}.
\newblock


\bibitem[\protect\citeauthoryear{{John Boaz Lee}, {Ryan Rossi}, and {Xiangnan
  Kong}}{{John Boaz Lee} et~al\mbox{.}}{2018}]%
        {81}
\bibfield{author}{\bibinfo{person}{{John Boaz Lee}}, \bibinfo{person}{{Ryan
  Rossi}}, {and} \bibinfo{person}{{Xiangnan Kong}}.}
  \bibinfo{year}{2018}\natexlab{}.
\newblock \showarticletitle{Graph classification using structural attention}.
  In \bibinfo{booktitle}{\emph{The International Conference on Knowledge
  Discovery and Data Mining ({SIGKDD})}}. \bibinfo{pages}{1666--1674}.
\newblock


\bibitem[\protect\citeauthoryear{{John Lafferty}, {Andrew McCallum}, and
  {Fernando CN Pereira}}{{John Lafferty} et~al\mbox{.}}{2001}]%
        {391}
\bibfield{author}{\bibinfo{person}{{John Lafferty}}, \bibinfo{person}{{Andrew
  McCallum}}, {and} \bibinfo{person}{{Fernando CN Pereira}}.}
  \bibinfo{year}{2001}\natexlab{}.
\newblock \showarticletitle{Conditional random field: probabilistic models for
  segmenting and labeling sequence data}. In \bibinfo{booktitle}{\emph{The
  International Conference on Machine Learning (ICML)}}.
  \bibinfo{pages}{282--289}.
\newblock


\bibitem[\protect\citeauthoryear{{Jun Wu}, {Jingrui He}, and {Jiejun Xu}}{{Jun
  Wu} et~al\mbox{.}}{2019}]%
        {21}
\bibfield{author}{\bibinfo{person}{{Jun Wu}}, \bibinfo{person}{{Jingrui He}},
  {and} \bibinfo{person}{{Jiejun Xu}}.} \bibinfo{year}{2019}\natexlab{}.
\newblock \showarticletitle{{DEMO}-Net: degree-specific graph neural networks
  for node and graph classification}. In \bibinfo{booktitle}{\emph{The {ACM}
  International Conference on Knowledge Discovery and Data Mining ({SIGKDD})}}.
  \bibinfo{pages}{406--415}.
\newblock


\bibitem[\protect\citeauthoryear{{Junhyun Lee}, {Inyeop Lee}, and {Jaewoo
  Kang}}{{Junhyun Lee} et~al\mbox{.}}{2019}]%
        {86}
\bibfield{author}{\bibinfo{person}{{Junhyun Lee}}, \bibinfo{person}{{Inyeop
  Lee}}, {and} \bibinfo{person}{{Jaewoo Kang}}.}
  \bibinfo{year}{2019}\natexlab{}.
\newblock \showarticletitle{Self-Attention Graph Pooling}. In
  \bibinfo{booktitle}{\emph{The International Conference on Machine Learning
  ({ICML})}}. \bibinfo{pages}{3734--3743}.
\newblock


\bibitem[\protect\citeauthoryear{{Justin Gilmer}, {Samuel S. Schoenholz},
  {Patrick F. Riley}, {Oriol Vinyals}, and {George E. Dahl}}{{Justin Gilmer}
  et~al\mbox{.}}{2017}]%
        {93}
\bibfield{author}{\bibinfo{person}{{Justin Gilmer}}, \bibinfo{person}{{Samuel
  S. Schoenholz}}, \bibinfo{person}{{Patrick F. Riley}},
  \bibinfo{person}{{Oriol Vinyals}}, {and} \bibinfo{person}{{George E. Dahl}}.}
  \bibinfo{year}{2017}\natexlab{}.
\newblock \showarticletitle{Neural Message Passing for Quantum Chemistry}. In
  \bibinfo{booktitle}{\emph{The International Conference on Machine Learning
  ({ICML})}}. \bibinfo{pages}{1263--1272}.
\newblock


\bibitem[\protect\citeauthoryear{{Kai Sheng Tai}, {Richard Socher}, and
  {Christopher D. Manning}}{{Kai Sheng Tai} et~al\mbox{.}}{2015}]%
        {123}
\bibfield{author}{\bibinfo{person}{{Kai Sheng Tai}}, \bibinfo{person}{{Richard
  Socher}}, {and} \bibinfo{person}{{Christopher D. Manning}}.}
  \bibinfo{year}{2015}\natexlab{}.
\newblock \showarticletitle{Improved semantic representations from
  tree-structured long short-term memory networks}. In
  \bibinfo{booktitle}{\emph{The Annual Meeting of the Association for
  Computational Linguistics}}. \bibinfo{pages}{1556--1566}.
\newblock


\bibitem[\protect\citeauthoryear{{Kaiming He}, {Xiangyu Zhang}, {Shaoqing Ren},
  and {Jian Sun}}{{Kaiming He} et~al\mbox{.}}{2016}]%
        {382}
\bibfield{author}{\bibinfo{person}{{Kaiming He}}, \bibinfo{person}{{Xiangyu
  Zhang}}, \bibinfo{person}{{Shaoqing Ren}}, {and} \bibinfo{person}{{Jian
  Sun}}.} \bibinfo{year}{2016}\natexlab{}.
\newblock \showarticletitle{Deep residual learning for image recognition}. In
  \bibinfo{booktitle}{\emph{The IEEE International Conference on Computer
  Vision and Pattern Recognition (CVPR)}}. \bibinfo{pages}{770--778}.
\newblock


\bibitem[\protect\citeauthoryear{{Kaixiong Zhou}, {Qingquan Song}, {Xiao
  Huang}, and {Xia Hu}}{{Kaixiong Zhou} et~al\mbox{.}}{2019}]%
        {172}
\bibfield{author}{\bibinfo{person}{{Kaixiong Zhou}}, \bibinfo{person}{{Qingquan
  Song}}, \bibinfo{person}{{Xiao Huang}}, {and} \bibinfo{person}{{Xia Hu}}.}
  \bibinfo{year}{2019}\natexlab{}.
\newblock \showarticletitle{Auto-GNN: neural architecture search of graph
  neural networks}.
\newblock  (\bibinfo{year}{2019}).
\newblock
\urldef\tempurl%
\url{https://arxiv.org/abs/1909.03184}
\showURL{%
\tempurl}


\bibitem[\protect\citeauthoryear{{Ke Sun}, {Piotr Koniusz}, and {Zhen
  Wang}}{{Ke Sun} et~al\mbox{.}}{2019}]%
        {180}
\bibfield{author}{\bibinfo{person}{{Ke Sun}}, \bibinfo{person}{{Piotr
  Koniusz}}, {and} \bibinfo{person}{{Zhen Wang}}.}
  \bibinfo{year}{2019}\natexlab{}.
\newblock \showarticletitle{Fisher-Bures adversary graph convolutional
  networks}. In \bibinfo{booktitle}{\emph{The International Conference on
  Uncertainty in Artificial Intelligence}}. \bibinfo{pages}{No. 161}.
\newblock


\bibitem[\protect\citeauthoryear{{Ke Tu}, {Peng Cui}, {Xiao Wang}, {Fei Wang},
  and {Wenwu Zhu}}{{Ke Tu} et~al\mbox{.}}{2018b}]%
        {136}
\bibfield{author}{\bibinfo{person}{{Ke Tu}}, \bibinfo{person}{{Peng Cui}},
  \bibinfo{person}{{Xiao Wang}}, \bibinfo{person}{{Fei Wang}}, {and}
  \bibinfo{person}{{Wenwu Zhu}}.} \bibinfo{year}{2018}\natexlab{b}.
\newblock \showarticletitle{Structural Deep Embedding for Hyper-Networks}. In
  \bibinfo{booktitle}{\emph{The {AAAI} Conference on Artificial Intelligence
  ({AAAI})}}. \bibinfo{pages}{426--433}.
\newblock


\bibitem[\protect\citeauthoryear{{Ke Tu}, {Peng Cui}, {Xiao Wang}, and {Philip
  S. Yu}}{{Ke Tu} et~al\mbox{.}}{2018a}]%
        {134}
\bibfield{author}{\bibinfo{person}{{Ke Tu}}, \bibinfo{person}{{Peng Cui}},
  \bibinfo{person}{{Xiao Wang}}, {and} \bibinfo{person}{{Philip S. Yu}}.}
  \bibinfo{year}{2018}\natexlab{a}.
\newblock \showarticletitle{Deep Recursive Network Embedding with Regular
  Equivalence}. In \bibinfo{booktitle}{\emph{The {ACM} International Conference
  on Knowledge Discovery and Data Mining ({SIGKDD})}}.
  \bibinfo{pages}{2357--2366}.
\newblock


\bibitem[\protect\citeauthoryear{{Keyulu Xu}, {Chengtao Li}, {Yonglong Tian},
  {Tomohiro Sonobe}, {Ken-ichi Kawarabayashi}, and {Stefanie Jegelka}}{{Keyulu
  Xu} et~al\mbox{.}}{2018}]%
        {89}
\bibfield{author}{\bibinfo{person}{{Keyulu Xu}}, \bibinfo{person}{{Chengtao
  Li}}, \bibinfo{person}{{Yonglong Tian}}, \bibinfo{person}{{Tomohiro Sonobe}},
  \bibinfo{person}{{Ken-ichi Kawarabayashi}}, {and} \bibinfo{person}{{Stefanie
  Jegelka}}.} \bibinfo{year}{2018}\natexlab{}.
\newblock \showarticletitle{Representation Learning on Graphs with Jumping
  Knowledge Networks}. In \bibinfo{booktitle}{\emph{The International
  Conference on Machine Learning ({ICML})}}. \bibinfo{pages}{5453--5462}.
\newblock


\bibitem[\protect\citeauthoryear{{Keyulu Xu}, {Jingling Li}, {Mozhi Zhang},
  {Simon S. Du}, {Ken-ichi Kawarabayashi}, and {Stefanie Jegelka}}{{Keyulu Xu}
  et~al\mbox{.}}{2020}]%
        {401}
\bibfield{author}{\bibinfo{person}{{Keyulu Xu}}, \bibinfo{person}{{Jingling
  Li}}, \bibinfo{person}{{Mozhi Zhang}}, \bibinfo{person}{{Simon S. Du}},
  \bibinfo{person}{{Ken-ichi Kawarabayashi}}, {and} \bibinfo{person}{{Stefanie
  Jegelka}}.} \bibinfo{year}{2020}\natexlab{}.
\newblock \showarticletitle{What can neural networks reason about}. In
  \bibinfo{booktitle}{\emph{The International Conference on Learning
  Representations ({ICLR})}}.
\newblock


\bibitem[\protect\citeauthoryear{{Keyulu Xu}, {Weihua Hu}, {Jure Leskovec}, and
  {Stefanie Jegelka}}{{Keyulu Xu} et~al\mbox{.}}{2019}]%
        {83}
\bibfield{author}{\bibinfo{person}{{Keyulu Xu}}, \bibinfo{person}{{Weihua Hu}},
  \bibinfo{person}{{Jure Leskovec}}, {and} \bibinfo{person}{{Stefanie
  Jegelka}}.} \bibinfo{year}{2019}\natexlab{}.
\newblock \showarticletitle{How powerful are graph neural networks}. In
  \bibinfo{booktitle}{\emph{The International Conference on Learning
  Representations ({ICLR})}}.
\newblock


\bibitem[\protect\citeauthoryear{{KiJung Yoon}, {Renjie Liao}, {Yuwen Xiong},
  {Lisa Zhang}, {Ethan Fetaya}, {Raquel Urtasun}, {Richard Zemel}, and {Xaq
  Pitkow}}{{KiJung Yoon} et~al\mbox{.}}{2018}]%
        {158}
\bibfield{author}{\bibinfo{person}{{KiJung Yoon}}, \bibinfo{person}{{Renjie
  Liao}}, \bibinfo{person}{{Yuwen Xiong}}, \bibinfo{person}{{Lisa Zhang}},
  \bibinfo{person}{{Ethan Fetaya}}, \bibinfo{person}{{Raquel Urtasun}},
  \bibinfo{person}{{Richard Zemel}}, {and} \bibinfo{person}{{Xaq Pitkow}}.}
  \bibinfo{year}{2018}\natexlab{}.
\newblock \showarticletitle{Inference in probabilistic graphical models by
  graph neural networks}. In \bibinfo{booktitle}{\emph{The International
  Conference on Learning Representations ({ICLR})}}.
\newblock


\bibitem[\protect\citeauthoryear{{Kilian Weinberger}, {Anirban Dasgupta}, and
  {John Langford}}{{Kilian Weinberger} et~al\mbox{.}}{2009}]%
        {386}
\bibfield{author}{\bibinfo{person}{{Kilian Weinberger}},
  \bibinfo{person}{{Anirban Dasgupta}}, {and} \bibinfo{person}{{John
  Langford}}.} \bibinfo{year}{2009}\natexlab{}.
\newblock \showarticletitle{Feature hashing for large scale multitask
  learning}. In \bibinfo{booktitle}{\emph{The International Conference on
  Machine Learning (ICML)}}. \bibinfo{pages}{1113--1120}.
\newblock


\bibitem[\protect\citeauthoryear{{Kiran K. Thekumparampil}, {Chong Wang},
  {Sewoong Oh}, and {Lijia Li}}{{Kiran K. Thekumparampil}
  et~al\mbox{.}}{2018}]%
        {67}
\bibfield{author}{\bibinfo{person}{{Kiran K. Thekumparampil}},
  \bibinfo{person}{{Chong Wang}}, \bibinfo{person}{{Sewoong Oh}}, {and}
  \bibinfo{person}{{Lijia Li}}.} \bibinfo{year}{2018}\natexlab{}.
\newblock \showarticletitle{Attention-based graph neural network for
  semi-supervised learning}.
\newblock  (\bibinfo{year}{2018}).
\newblock
\urldef\tempurl%
\url{https://arxiv.org/abs/1803.03735}
\showURL{%
\tempurl}


\bibitem[\protect\citeauthoryear{{Kun Xu}, {Lingfei Wu}, {Zhiguo Wang},
  {Yansong Feng}, {Michael Witbrock}, and {Vadim Sheinin}}{{Kun Xu}
  et~al\mbox{.}}{2018}]%
        {106}
\bibfield{author}{\bibinfo{person}{{Kun Xu}}, \bibinfo{person}{{Lingfei Wu}},
  \bibinfo{person}{{Zhiguo Wang}}, \bibinfo{person}{{Yansong Feng}},
  \bibinfo{person}{{Michael Witbrock}}, {and} \bibinfo{person}{{Vadim
  Sheinin}}.} \bibinfo{year}{2018}\natexlab{}.
\newblock \showarticletitle{{Graph2Seq}: graph to sequence learning with
  attention-based neural networks}.
\newblock  (\bibinfo{year}{2018}).
\newblock
\urldef\tempurl%
\url{https://arxiv.org/abs/1804.00823}
\showURL{%
\tempurl}


\bibitem[\protect\citeauthoryear{{Kyunghyun Cho}, {Bart van Merrienboer},
  {Caglar Gulcehre}, {Dzmitry Bahdanau}, {Fethi Bougares}, {Holger Schwenk},
  and {Youshua Bengio}}{{Kyunghyun Cho} et~al\mbox{.}}{2014}]%
        {115}
\bibfield{author}{\bibinfo{person}{{Kyunghyun Cho}}, \bibinfo{person}{{Bart van
  Merrienboer}}, \bibinfo{person}{{Caglar Gulcehre}}, \bibinfo{person}{{Dzmitry
  Bahdanau}}, \bibinfo{person}{{Fethi Bougares}}, \bibinfo{person}{{Holger
  Schwenk}}, {and} \bibinfo{person}{{Youshua Bengio}}.}
  \bibinfo{year}{2014}\natexlab{}.
\newblock \showarticletitle{Learning phrase representations using RNN
  encoder-decoder for statistical machine translation}. In
  \bibinfo{booktitle}{\emph{The Conference on Empirical Methods in Natural
  Language Processing (EMNLP)}}. \bibinfo{pages}{1724--1734}.
\newblock


\bibitem[\protect\citeauthoryear{{Liang Yao}, {Chengsheng Mao}, and {Yuan
  Luo}}{{Liang Yao} et~al\mbox{.}}{2019}]%
        {318}
\bibfield{author}{\bibinfo{person}{{Liang Yao}}, \bibinfo{person}{{Chengsheng
  Mao}}, {and} \bibinfo{person}{{Yuan Luo}}.} \bibinfo{year}{2019}\natexlab{}.
\newblock \showarticletitle{Graph convolutional networks for text
  classification}. In \bibinfo{booktitle}{\emph{The {AAAI} Conference on
  Artificial Intelligence ({AAAI})}}. \bibinfo{pages}{7370--7377}.
\newblock


\bibitem[\protect\citeauthoryear{{Lichao Sun}, {Yingtong Dou}, {Carl Yang}, {Ji
  Wang}, {Philip S. Yu}, and {Bo Li}}{{Lichao Sun} et~al\mbox{.}}{2020}]%
        {365}
\bibfield{author}{\bibinfo{person}{{Lichao Sun}}, \bibinfo{person}{{Yingtong
  Dou}}, \bibinfo{person}{{Carl Yang}}, \bibinfo{person}{{Ji Wang}},
  \bibinfo{person}{{Philip S. Yu}}, {and} \bibinfo{person}{{Bo Li}}.}
  \bibinfo{year}{2020}\natexlab{}.
\newblock \showarticletitle{Adversarial attack and defense on graph data: a
  survey}.
\newblock  (\bibinfo{year}{2020}).
\newblock
\urldef\tempurl%
\url{https://arxiv.org/abs/1812.10528}
\showURL{%
\tempurl}


\bibitem[\protect\citeauthoryear{{Lin Qiu}, {Yunxuan Xiao}, {Yanru Qu}, {Hao
  Zhou}, {Lei Li}, {Weinan Zhang}, and {Yong Yu}}{{Lin Qiu}
  et~al\mbox{.}}{2019}]%
        {272}
\bibfield{author}{\bibinfo{person}{{Lin Qiu}}, \bibinfo{person}{{Yunxuan
  Xiao}}, \bibinfo{person}{{Yanru Qu}}, \bibinfo{person}{{Hao Zhou}},
  \bibinfo{person}{{Lei Li}}, \bibinfo{person}{{Weinan Zhang}}, {and}
  \bibinfo{person}{{Yong Yu}}.} \bibinfo{year}{2019}\natexlab{}.
\newblock \showarticletitle{Dynamically fused graph network for multi-hop
  reasoning}. In \bibinfo{booktitle}{\emph{The Annual Meeting of the
  Association for Computational Linguistics ({ACL})}}.
  \bibinfo{pages}{6140--6150}.
\newblock


\bibitem[\protect\citeauthoryear{{Linfeng Liu} and {Liping Liu}}{{Linfeng Liu}
  and {Liping Liu}}{2019}]%
        {156}
\bibfield{author}{\bibinfo{person}{{Linfeng Liu}} {and}
  \bibinfo{person}{{Liping Liu}}.} \bibinfo{year}{2019}\natexlab{}.
\newblock \showarticletitle{Amortized Variational Inference with Graph
  Convolutional Networks for Gaussian Processes}. In
  \bibinfo{booktitle}{\emph{The International Conference on Artificial
  Intelligence and Statistics ({AISTATS})}}. \bibinfo{pages}{2291--2300}.
\newblock


\bibitem[\protect\citeauthoryear{{Lingxiao Ma}, {Zhi Yang}, {Youshan Miao},
  {Jilong Xue}, {Ming Wu}, {Lidong Zhou}, and {Yafei Dai}}{{Lingxiao Ma}
  et~al\mbox{.}}{2018}]%
        {40}
\bibfield{author}{\bibinfo{person}{{Lingxiao Ma}}, \bibinfo{person}{{Zhi
  Yang}}, \bibinfo{person}{{Youshan Miao}}, \bibinfo{person}{{Jilong Xue}},
  \bibinfo{person}{{Ming Wu}}, \bibinfo{person}{{Lidong Zhou}}, {and}
  \bibinfo{person}{{Yafei Dai}}.} \bibinfo{year}{2018}\natexlab{}.
\newblock \showarticletitle{Towards efficient large-scale graph neural network
  computing}.
\newblock  (\bibinfo{year}{2018}).
\newblock
\urldef\tempurl%
\url{https://arxiv.org/abs/1810.08403}
\showURL{%
\tempurl}


\bibitem[\protect\citeauthoryear{{Lingxiao Zhao} and {Leman Akoglu}}{{Lingxiao
  Zhao} and {Leman Akoglu}}{2020}]%
        {373}
\bibfield{author}{\bibinfo{person}{{Lingxiao Zhao}} {and}
  \bibinfo{person}{{Leman Akoglu}}.} \bibinfo{year}{2020}\natexlab{}.
\newblock \showarticletitle{{PairNorm}: tackling oversmoothing in {GNNs}}. In
  \bibinfo{booktitle}{\emph{The International Conference on Learning
  Representations ({ICLR})}}.
\newblock


\bibitem[\protect\citeauthoryear{{Louis Tiao}, {Pantelis Elinas}, {Harrison
  Nguyen}, and {Edwin V. Bonilla}}{{Louis Tiao} et~al\mbox{.}}{2019}]%
        {157}
\bibfield{author}{\bibinfo{person}{{Louis Tiao}}, \bibinfo{person}{{Pantelis
  Elinas}}, \bibinfo{person}{{Harrison Nguyen}}, {and} \bibinfo{person}{{Edwin
  V. Bonilla}}.} \bibinfo{year}{2019}\natexlab{}.
\newblock \showarticletitle{Variational spectral graph convolutional networks}.
\newblock  (\bibinfo{year}{2019}).
\newblock
\urldef\tempurl%
\url{https://arxiv.org/abs/1906.01852v1}
\showURL{%
\tempurl}


\bibitem[\protect\citeauthoryear{{Luana Ruiz}, {Fernando Gama}, and {Alejandro
  Ribeiro}}{{Luana Ruiz} et~al\mbox{.}}{2019}]%
        {108}
\bibfield{author}{\bibinfo{person}{{Luana Ruiz}}, \bibinfo{person}{{Fernando
  Gama}}, {and} \bibinfo{person}{{Alejandro Ribeiro}}.}
  \bibinfo{year}{2019}\natexlab{}.
\newblock \showarticletitle{Gated Graph Convolutional Recurrent Neural
  Networks}.
\newblock  (\bibinfo{year}{2019}).
\newblock
\urldef\tempurl%
\url{https://arxiv.org/abs/1903.01888}
\showURL{%
\tempurl}


\bibitem[\protect\citeauthoryear{{Marcelo Daniel Gutierrez Mallea}, {Peter
  Meltzer}, and {Peter J. Bentley}}{{Marcelo Daniel Gutierrez Mallea}
  et~al\mbox{.}}{2019}]%
        {54}
\bibfield{author}{\bibinfo{person}{{Marcelo Daniel Gutierrez Mallea}},
  \bibinfo{person}{{Peter Meltzer}}, {and} \bibinfo{person}{{Peter J.
  Bentley}}.} \bibinfo{year}{2019}\natexlab{}.
\newblock \showarticletitle{Capsule neural networks for graph classification
  using explicit tensorial graph representations}.
\newblock  (\bibinfo{year}{2019}).
\newblock
\urldef\tempurl%
\url{https://arxiv.org/abs/1902.08399}
\showURL{%
\tempurl}


\bibitem[\protect\citeauthoryear{{Marcelo Prates}, {Pedro H.C. Avelar},
  {Henrique Lemos}, {Luis C. Lamb}, and {Moshe Y. Vardi}}{{Marcelo Prates}
  et~al\mbox{.}}{2019}]%
        {216}
\bibfield{author}{\bibinfo{person}{{Marcelo Prates}}, \bibinfo{person}{{Pedro
  H.C. Avelar}}, \bibinfo{person}{{Henrique Lemos}}, \bibinfo{person}{{Luis C.
  Lamb}}, {and} \bibinfo{person}{{Moshe Y. Vardi}}.}
  \bibinfo{year}{2019}\natexlab{}.
\newblock \showarticletitle{Learing to solve {NP}-Complete problems: a graph
  neural network for decision {TSP}}. In \bibinfo{booktitle}{\emph{The {AAAI}
  Conference on Artificial Intelligence ({AAAI})}}.
  \bibinfo{pages}{4731--4738}.
\newblock


\bibitem[\protect\citeauthoryear{{Marnka Zitnik}, {Monica Agrawal}, and {Jure
  Leskovec}}{{Marnka Zitnik} et~al\mbox{.}}{2018}]%
        {201}
\bibfield{author}{\bibinfo{person}{{Marnka Zitnik}}, \bibinfo{person}{{Monica
  Agrawal}}, {and} \bibinfo{person}{{Jure Leskovec}}.}
  \bibinfo{year}{2018}\natexlab{}.
\newblock \showarticletitle{Modeling polypharmacy side effects with graph
  convolutional networks}.
\newblock \bibinfo{journal}{\emph{Bioinformatics}} \bibinfo{volume}{34},
  \bibinfo{number}{13} (\bibinfo{year}{2018}), \bibinfo{pages}{i457--i466}.
\newblock


\bibitem[\protect\citeauthoryear{{Martin Simonovsky} and {Nikos
  Komodakis}}{{Martin Simonovsky} and {Nikos Komodakis}}{2017}]%
        {43}
\bibfield{author}{\bibinfo{person}{{Martin Simonovsky}} {and}
  \bibinfo{person}{{Nikos Komodakis}}.} \bibinfo{year}{2017}\natexlab{}.
\newblock \showarticletitle{Dynamic edge-conditioned filters in convolutional
  neural networks on graphs}. In \bibinfo{booktitle}{\emph{The {IEEE}
  Conference on Computer Vision and Pattern Recognition ({CVPR})}}.
  \bibinfo{pages}{29--38}.
\newblock


\bibitem[\protect\citeauthoryear{{Martin Simonovsky} and {Nikos
  Komodakis}}{{Martin Simonovsky} and {Nikos Komodakis}}{2018}]%
        {146}
\bibfield{author}{\bibinfo{person}{{Martin Simonovsky}} {and}
  \bibinfo{person}{{Nikos Komodakis}}.} \bibinfo{year}{2018}\natexlab{}.
\newblock \showarticletitle{{GraphVAE}: towards generation of small graphs
  using variational autoencoders}. In \bibinfo{booktitle}{\emph{The
  International Conference on Artificial Neural Networks ({ICANN})}}.
  \bibinfo{pages}{412--422}.
\newblock


\bibitem[\protect\citeauthoryear{{Mathias Niepert}, {Mohamed Ahmed}, and
  {Konstantin Kutzkov}}{{Mathias Niepert} et~al\mbox{.}}{2016}]%
        {48}
\bibfield{author}{\bibinfo{person}{{Mathias Niepert}},
  \bibinfo{person}{{Mohamed Ahmed}}, {and} \bibinfo{person}{{Konstantin
  Kutzkov}}.} \bibinfo{year}{2016}\natexlab{}.
\newblock \showarticletitle{Learning convolutional neural networks for graphs}.
  In \bibinfo{booktitle}{\emph{The International Conference on Machine Learning
  ({ICLR})}}. \bibinfo{pages}{2014--2023}.
\newblock


\bibitem[\protect\citeauthoryear{{Matthew Baron}}{{Matthew Baron}}{2018}]%
        {39}
\bibfield{author}{\bibinfo{person}{{Matthew Baron}}.}
  \bibinfo{year}{2018}\natexlab{}.
\newblock \showarticletitle{Topology and prediction focused research on graph
  convolutional neural networks}.
\newblock  (\bibinfo{year}{2018}).
\newblock
\urldef\tempurl%
\url{https://arxiv.org/abs/1808.07769v1}
\showURL{%
\tempurl}


\bibitem[\protect\citeauthoryear{{Meng Qu}, {Yoshua Bengio}, and {Jian
  Tang}}{{Meng Qu} et~al\mbox{.}}{2019}]%
        {80}
\bibfield{author}{\bibinfo{person}{{Meng Qu}}, \bibinfo{person}{{Yoshua
  Bengio}}, {and} \bibinfo{person}{{Jian Tang}}.}
  \bibinfo{year}{2019}\natexlab{}.
\newblock \showarticletitle{{GMNN}: Graph Markov Neural Networks}. In
  \bibinfo{booktitle}{\emph{The International Conference on Machine Learning
  ({ICML})}}. \bibinfo{pages}{5241--5250}.
\newblock


\bibitem[\protect\citeauthoryear{{Micha\"{e}l Defferrard}, {Xavier Bresson},
  and {Pierre Vandergheynst}}{{Micha\"{e}l Defferrard} et~al\mbox{.}}{2016}]%
        {78}
\bibfield{author}{\bibinfo{person}{{Micha\"{e}l Defferrard}},
  \bibinfo{person}{{Xavier Bresson}}, {and} \bibinfo{person}{{Pierre
  Vandergheynst}}.} \bibinfo{year}{2016}\natexlab{}.
\newblock \showarticletitle{Convolutional neural networks on graphs with fast
  localized spectral filtering}. In \bibinfo{booktitle}{\emph{Advances in
  Neural Information Processing Systems ({NIPS})}}.
  \bibinfo{pages}{3844--3852}.
\newblock


\bibitem[\protect\citeauthoryear{{Michael M. Bronstein}, {Joan Bruna}, {Yan
  LeCun}, {Arthur Szlam}, and {Pierre Vandergheynst}}{{Michael M. Bronstein}
  et~al\mbox{.}}{2017}]%
        {1}
\bibfield{author}{\bibinfo{person}{{Michael M. Bronstein}},
  \bibinfo{person}{{Joan Bruna}}, \bibinfo{person}{{Yan LeCun}},
  \bibinfo{person}{{Arthur Szlam}}, {and} \bibinfo{person}{{Pierre
  Vandergheynst}}.} \bibinfo{year}{2017}\natexlab{}.
\newblock \showarticletitle{Geometric deep learning: going beyond Euclidean
  data}.
\newblock \bibinfo{journal}{\emph{{IEEE} Signal Processing Magazine}}
  \bibinfo{volume}{34}, \bibinfo{number}{4} (\bibinfo{year}{2017}),
  \bibinfo{pages}{18--42}.
\newblock


\bibitem[\protect\citeauthoryear{{Mikael Henaff}, {Joan Bruna}, and {Yan
  LeCun}}{{Mikael Henaff} et~al\mbox{.}}{2015}]%
        {27}
\bibfield{author}{\bibinfo{person}{{Mikael Henaff}}, \bibinfo{person}{{Joan
  Bruna}}, {and} \bibinfo{person}{{Yan LeCun}}.}
  \bibinfo{year}{2015}\natexlab{}.
\newblock \showarticletitle{Deep convolutional networks on graph-structured
  data}.
\newblock  (\bibinfo{year}{2015}).
\newblock
\urldef\tempurl%
\url{https://arxiv.org/abs/1506.05163}
\showURL{%
\tempurl}


\bibitem[\protect\citeauthoryear{{Miltiadis Allamanis}, {Marc Brockschmidt},
  and {Mahmoud Khademi}}{{Miltiadis Allamanis} et~al\mbox{.}}{2018}]%
        {340}
\bibfield{author}{\bibinfo{person}{{Miltiadis Allamanis}},
  \bibinfo{person}{{Marc Brockschmidt}}, {and} \bibinfo{person}{{Mahmoud
  Khademi}}.} \bibinfo{year}{2018}\natexlab{}.
\newblock \showarticletitle{Learning to represent programs with graphs}. In
  \bibinfo{booktitle}{\emph{The International Conference on Learning
  Representations ({ICLR})}}.
\newblock


\bibitem[\protect\citeauthoryear{{Ming Ding}, {Chang Zhou}, {Qibin Chen},
  {Hongxia Yang}, and {Jie Tang}}{{Ming Ding} et~al\mbox{.}}{2019}]%
        {308}
\bibfield{author}{\bibinfo{person}{{Ming Ding}}, \bibinfo{person}{{Chang
  Zhou}}, \bibinfo{person}{{Qibin Chen}}, \bibinfo{person}{{Hongxia Yang}},
  {and} \bibinfo{person}{{Jie Tang}}.} \bibinfo{year}{2019}\natexlab{}.
\newblock \showarticletitle{Cognitive graph for multi-hop reading comprehension
  at scale}. In \bibinfo{booktitle}{\emph{The Annual Meeting of the Association
  for Computational Linguistics ({ACL})}}. \bibinfo{pages}{2694--2703}.
\newblock


\bibitem[\protect\citeauthoryear{{Minjie Wang}, {Lingfan Yu}, {Da Zheng}, {Quan
  Gan}, {Yu Gai}, {Zihao Ye}, {Mufei Li}, {Jinjing Zhou}, {Qi Huang}, {Chao
  Ma}, {Ziyue Huang}, {Qipeng Guo}, {Hao Zhang}, {Haibin Lin}, {Junbo Zhao},
  {Jinyang Li}, {Alexander Smola}, and {Zheng Zhang}}{{Minjie Wang}
  et~al\mbox{.}}{2019}]%
        {402}
\bibfield{author}{\bibinfo{person}{{Minjie Wang}}, \bibinfo{person}{{Lingfan
  Yu}}, \bibinfo{person}{{Da Zheng}}, \bibinfo{person}{{Quan Gan}},
  \bibinfo{person}{{Yu Gai}}, \bibinfo{person}{{Zihao Ye}},
  \bibinfo{person}{{Mufei Li}}, \bibinfo{person}{{Jinjing Zhou}},
  \bibinfo{person}{{Qi Huang}}, \bibinfo{person}{{Chao Ma}},
  \bibinfo{person}{{Ziyue Huang}}, \bibinfo{person}{{Qipeng Guo}},
  \bibinfo{person}{{Hao Zhang}}, \bibinfo{person}{{Haibin Lin}},
  \bibinfo{person}{{Junbo Zhao}}, \bibinfo{person}{{Jinyang Li}},
  \bibinfo{person}{{Alexander Smola}}, {and} \bibinfo{person}{{Zheng Zhang}}.}
  \bibinfo{year}{2019}\natexlab{}.
\newblock \showarticletitle{Deep Graph Library: towards efficient and scalable
  deep learning on graphs}. In \bibinfo{booktitle}{\emph{The Workshop at the
  International Conference on Learning Representations ({ICLR} Workshop)}}.
\newblock


\bibitem[\protect\citeauthoryear{{Muhan Zhang} and {Yixin Chen}}{{Muhan Zhang}
  and {Yixin Chen}}{2018}]%
        {238}
\bibfield{author}{\bibinfo{person}{{Muhan Zhang}} {and} \bibinfo{person}{{Yixin
  Chen}}.} \bibinfo{year}{2018}\natexlab{}.
\newblock \showarticletitle{Link prediction based on graph neural networks}. In
  \bibinfo{booktitle}{\emph{The International Conference on Neural Information
  Processing Systems ({NIPS})}}. \bibinfo{pages}{5171--5181}.
\newblock


\bibitem[\protect\citeauthoryear{{Muhan Zhang}, {Zhicheng Cui}, {Marion
  Neumann}, and {Yixin Chen}}{{Muhan Zhang} et~al\mbox{.}}{2018}]%
        {92}
\bibfield{author}{\bibinfo{person}{{Muhan Zhang}}, \bibinfo{person}{{Zhicheng
  Cui}}, \bibinfo{person}{{Marion Neumann}}, {and} \bibinfo{person}{{Yixin
  Chen}}.} \bibinfo{year}{2018}\natexlab{}.
\newblock \showarticletitle{An end-to-end deep learning architecture for graph
  classification}. In \bibinfo{booktitle}{\emph{The {AAAI} Conference on
  Artificial Intelligence ({AAAI})}}. \bibinfo{pages}{4438--4445}.
\newblock


\bibitem[\protect\citeauthoryear{{Naganand Yadati}, {Madhav Nimishakavi},
  {Prateek Yadav}, {Vikram Nitin}, {Anand Louis}, and {Partha
  Talukdar}}{{Naganand Yadati} et~al\mbox{.}}{2019}]%
        {64}
\bibfield{author}{\bibinfo{person}{{Naganand Yadati}}, \bibinfo{person}{{Madhav
  Nimishakavi}}, \bibinfo{person}{{Prateek Yadav}}, \bibinfo{person}{{Vikram
  Nitin}}, \bibinfo{person}{{Anand Louis}}, {and} \bibinfo{person}{{Partha
  Talukdar}}.} \bibinfo{year}{2019}\natexlab{}.
\newblock \showarticletitle{{HyperGCN}: a new method for training graph
  convolutional networks on hypergraphs}. In \bibinfo{booktitle}{\emph{The
  International Conference on Neural Information Processing Systems
  ({NeurPS})}}. \bibinfo{pages}{1511--1522}.
\newblock


\bibitem[\protect\citeauthoryear{{Naganand Yadati}, {Tingran Gao}, {Shahab
  Asoodeh}, {Partha Talukdar}, and {Anand Louis}}{{Naganand Yadati}
  et~al\mbox{.}}{2020}]%
        {63}
\bibfield{author}{\bibinfo{person}{{Naganand Yadati}},
  \bibinfo{person}{{Tingran Gao}}, \bibinfo{person}{{Shahab Asoodeh}},
  \bibinfo{person}{{Partha Talukdar}}, {and} \bibinfo{person}{{Anand Louis}}.}
  \bibinfo{year}{2020}\natexlab{}.
\newblock \showarticletitle{Graph neural networks for soft semi-supervised
  learning on hypergraphs}.
\newblock  (\bibinfo{year}{2020}).
\newblock


\bibitem[\protect\citeauthoryear{{Nanyun Peng}, {Hoifung Poon}, {Chris Quirk},
  {Kristina Toutanova}, and {Wen-tau Yih}}{{Nanyun Peng} et~al\mbox{.}}{2017}]%
        {118}
\bibfield{author}{\bibinfo{person}{{Nanyun Peng}}, \bibinfo{person}{{Hoifung
  Poon}}, \bibinfo{person}{{Chris Quirk}}, \bibinfo{person}{{Kristina
  Toutanova}}, {and} \bibinfo{person}{{Wen-tau Yih}}.}
  \bibinfo{year}{2017}\natexlab{}.
\newblock \showarticletitle{Cross-sentence N-ary relation extraction with graph
  {LSTM}s}.
\newblock \bibinfo{journal}{\emph{Transactions of the Association for
  Computational Linguistics}} \bibinfo{volume}{5}, \bibinfo{number}{2017}
  (\bibinfo{year}{2017}), \bibinfo{pages}{101--115}.
\newblock


\bibitem[\protect\citeauthoryear{{Nicholas Watters}, {Andrea Tacchetti},
  {Th\'{e}ophane Weber}, {Razvan Pascanu}, {Peter Battaglia}, and {Daniel
  Zoran}}{{Nicholas Watters} et~al\mbox{.}}{2017}]%
        {339}
\bibfield{author}{\bibinfo{person}{{Nicholas Watters}},
  \bibinfo{person}{{Andrea Tacchetti}}, \bibinfo{person}{{Th\'{e}ophane
  Weber}}, \bibinfo{person}{{Razvan Pascanu}}, \bibinfo{person}{{Peter
  Battaglia}}, {and} \bibinfo{person}{{Daniel Zoran}}.}
  \bibinfo{year}{2017}\natexlab{}.
\newblock \showarticletitle{Visual interaction networks: learning a physics
  simulator from video}. In \bibinfo{booktitle}{\emph{The International
  Conference on Neural Inforamtion Processing Systems ({NIPS})}}.
  \bibinfo{pages}{4539--4547}.
\newblock


\bibitem[\protect\citeauthoryear{{Nicola De Cao} and {Thomas N. Kipf}}{{Nicola
  De Cao} and {Thomas N. Kipf}}{2018}]%
        {203}
\bibfield{author}{\bibinfo{person}{{Nicola De Cao}} {and}
  \bibinfo{person}{{Thomas N. Kipf}}.} \bibinfo{year}{2018}\natexlab{}.
\newblock \showarticletitle{{MolGAN}: an implicit generative model for small
  molecular graphs}.
\newblock  (\bibinfo{year}{2018}).
\newblock
\urldef\tempurl%
\url{https://arxiv.org/abs/1805.11973}
\showURL{%
\tempurl}


\bibitem[\protect\citeauthoryear{{Nicolas Keriven} and {Gabriel
  Peyr\'{e}}}{{Nicolas Keriven} and {Gabriel Peyr\'{e}}}{2019}]%
        {50}
\bibfield{author}{\bibinfo{person}{{Nicolas Keriven}} {and}
  \bibinfo{person}{{Gabriel Peyr\'{e}}}.} \bibinfo{year}{2019}\natexlab{}.
\newblock \showarticletitle{Universal invariant and equivariant graph neural
  networks}. In \bibinfo{booktitle}{\emph{The International Conference on
  Neural Information Processing Systems ({NeurPS})}}.
  \bibinfo{pages}{7092--7101}.
\newblock


\bibitem[\protect\citeauthoryear{{Nima Dehmamy}, {Albert-Laszlo Barabasi}, and
  {Rose Yu}}{{Nima Dehmamy} et~al\mbox{.}}{2019}]%
        {167}
\bibfield{author}{\bibinfo{person}{{Nima Dehmamy}},
  \bibinfo{person}{{Albert-Laszlo Barabasi}}, {and} \bibinfo{person}{{Rose
  Yu}}.} \bibinfo{year}{2019}\natexlab{}.
\newblock \showarticletitle{Understanding the Representation Power of Graph
  Neural Networks in Learning Graph Topology}. In \bibinfo{booktitle}{\emph{The
  International Conference on Neural Information Processing Systems
  ({NeurPS})}}. \bibinfo{pages}{15413--15423}.
\newblock


\bibitem[\protect\citeauthoryear{{Olaf Ronneberger}, {Philipp Fischer}, and
  {Thomas Brox}}{{Olaf Ronneberger} et~al\mbox{.}}{2015}]%
        {72}
\bibfield{author}{\bibinfo{person}{{Olaf Ronneberger}},
  \bibinfo{person}{{Philipp Fischer}}, {and} \bibinfo{person}{{Thomas Brox}}.}
  \bibinfo{year}{2015}\natexlab{}.
\newblock \showarticletitle{U-Net: convolutional networks for biomedical image
  segmentation}. In \bibinfo{booktitle}{\emph{The International Conference on
  Medical Image Computing and Computer-assisted Intervention (MCCAI)}}.
  \bibinfo{pages}{234--241}.
\newblock


\bibitem[\protect\citeauthoryear{{Oleksandr Shchur}, {Maximilian Mumme},
  {Aleksandar Bojchevski}, and {Stephan G\"{u}nnemann}}{{Oleksandr Shchur}
  et~al\mbox{.}}{2019}]%
        {170}
\bibfield{author}{\bibinfo{person}{{Oleksandr Shchur}},
  \bibinfo{person}{{Maximilian Mumme}}, \bibinfo{person}{{Aleksandar
  Bojchevski}}, {and} \bibinfo{person}{{Stephan G\"{u}nnemann}}.}
  \bibinfo{year}{2019}\natexlab{}.
\newblock \showarticletitle{Pitfalls of graph neural network evaluation}.
\newblock  (\bibinfo{year}{2019}).
\newblock
\urldef\tempurl%
\url{https://arxiv.org/abs/1811.05868}
\showURL{%
\tempurl}


\bibitem[\protect\citeauthoryear{{Pablo Barcel\'{o}}, {Egor V. Kostylev},
  {Mikael Monet}, {Jorge P\'{e}rez}, {Juan Reutter}, and {Juan Pablo
  Silva}}{{Pablo Barcel\'{o}} et~al\mbox{.}}{2020}]%
        {164}
\bibfield{author}{\bibinfo{person}{{Pablo Barcel\'{o}}}, \bibinfo{person}{{Egor
  V. Kostylev}}, \bibinfo{person}{{Mikael Monet}}, \bibinfo{person}{{Jorge
  P\'{e}rez}}, \bibinfo{person}{{Juan Reutter}}, {and} \bibinfo{person}{{Juan
  Pablo Silva}}.} \bibinfo{year}{2020}\natexlab{}.
\newblock \showarticletitle{The logical expressiveness of graph neural
  networks}. In \bibinfo{booktitle}{\emph{The International Conference on
  Learning Representations ({ICLR})}}.
\newblock


\bibitem[\protect\citeauthoryear{{Padraig Corcoran}}{{Padraig
  Corcoran}}{2019}]%
        {97}
\bibfield{author}{\bibinfo{person}{{Padraig Corcoran}}.}
  \bibinfo{year}{2019}\natexlab{}.
\newblock \showarticletitle{Function space pooling for graph convolutional
  networks}.
\newblock  (\bibinfo{year}{2019}).
\newblock
\urldef\tempurl%
\url{https://arxiv.org/abs/1905.06259}
\showURL{%
\tempurl}


\bibitem[\protect\citeauthoryear{{Palash Goyal}, {Nitin Kamra}, {Xinran He},
  and {Yan Liu}}{{Palash Goyal} et~al\mbox{.}}{2018}]%
        {142}
\bibfield{author}{\bibinfo{person}{{Palash Goyal}}, \bibinfo{person}{{Nitin
  Kamra}}, \bibinfo{person}{{Xinran He}}, {and} \bibinfo{person}{{Yan Liu}}.}
  \bibinfo{year}{2018}\natexlab{}.
\newblock \showarticletitle{{DynGEM}: deep embedding method for dynamic
  graphs}.
\newblock  (\bibinfo{year}{2018}).
\newblock
\urldef\tempurl%
\url{https://arxiv.org/abs/1805.11273v1}
\showURL{%
\tempurl}


\bibitem[\protect\citeauthoryear{{Paul Almasan}, {Jos\'{e} Su\'{a}rez-Varela},
  {Arnau Badia-Sampera}, {Krzysztof Rusek}, {Pere Barlet-Ros}, and {Albert
  Cabellos-Aparicio}}{{Paul Almasan} et~al\mbox{.}}{2019}]%
        {186}
\bibfield{author}{\bibinfo{person}{{Paul Almasan}}, \bibinfo{person}{{Jos\'{e}
  Su\'{a}rez-Varela}}, \bibinfo{person}{{Arnau Badia-Sampera}},
  \bibinfo{person}{{Krzysztof Rusek}}, \bibinfo{person}{{Pere Barlet-Ros}},
  {and} \bibinfo{person}{{Albert Cabellos-Aparicio}}.}
  \bibinfo{year}{2019}\natexlab{}.
\newblock \showarticletitle{Deep reinforcement learning meets graph neural
  networks: exploring a routing optimization use case}.
\newblock  (\bibinfo{year}{2019}).
\newblock
\urldef\tempurl%
\url{https://arxiv.org/abs/1910.07421}
\showURL{%
\tempurl}


\bibitem[\protect\citeauthoryear{{Peng Han}, {Peng Yang}, {Peilin Zhao}, {Shuo
  Shang}, {Yong Liu}, {Jiayu Zhou}, {Xin Gao}, and {Panos Kalnis}}{{Peng Han}
  et~al\mbox{.}}{2019}]%
        {210}
\bibfield{author}{\bibinfo{person}{{Peng Han}}, \bibinfo{person}{{Peng Yang}},
  \bibinfo{person}{{Peilin Zhao}}, \bibinfo{person}{{Shuo Shang}},
  \bibinfo{person}{{Yong Liu}}, \bibinfo{person}{{Jiayu Zhou}},
  \bibinfo{person}{{Xin Gao}}, {and} \bibinfo{person}{{Panos Kalnis}}.}
  \bibinfo{year}{2019}\natexlab{}.
\newblock \showarticletitle{{GCN}-{MF}: disease-gene association identification
  by graph convolutional networks and matrix factorization}. In
  \bibinfo{booktitle}{\emph{The International Conference on Knowledge Discovery
  and Data Mining ({SIGKDD})}}. \bibinfo{pages}{705--713}.
\newblock


\bibitem[\protect\citeauthoryear{{Petar Veli\u\{c\}kovi\'{c}}, {Guillem
  Cucurull}, {Arantxa Casanova}, {Adriana Romero}, {Pietro Li\`{o}}, and
  {Yoshua Bengio}}{{Petar Veli\u\{c\}kovi\'{c}} et~al\mbox{.}}{2018}]%
        {66}
\bibfield{author}{\bibinfo{person}{{Petar Veli\u\{c\}kovi\'{c}}},
  \bibinfo{person}{{Guillem Cucurull}}, \bibinfo{person}{{Arantxa Casanova}},
  \bibinfo{person}{{Adriana Romero}}, \bibinfo{person}{{Pietro Li\`{o}}}, {and}
  \bibinfo{person}{{Yoshua Bengio}}.} \bibinfo{year}{2018}\natexlab{}.
\newblock \showarticletitle{Graph Attention Networks}. In
  \bibinfo{booktitle}{\emph{The International Conference on Learning
  Representations ({ICLR})}}.
\newblock


\bibitem[\protect\citeauthoryear{{Petar Veli\u\{c\}kovi\'{c}}, {William Fedus},
  {William L. Hamilton}, {Pietro Li\`{o}}, {Yoshua Bengio}, and {R Devon
  Hjelm}}{{Petar Veli\u\{c\}kovi\'{c}} et~al\mbox{.}}{2019}]%
        {140}
\bibfield{author}{\bibinfo{person}{{Petar Veli\u\{c\}kovi\'{c}}},
  \bibinfo{person}{{William Fedus}}, \bibinfo{person}{{William L. Hamilton}},
  \bibinfo{person}{{Pietro Li\`{o}}}, \bibinfo{person}{{Yoshua Bengio}}, {and}
  \bibinfo{person}{{R Devon Hjelm}}.} \bibinfo{year}{2019}\natexlab{}.
\newblock \showarticletitle{Deep graph infomax}. In
  \bibinfo{booktitle}{\emph{The International Conference on Learning
  Representations ({ICLR})}}.
\newblock


\bibitem[\protect\citeauthoryear{{Peter Battaglia}, {Razvan Pascanu}, {Matthew
  Lai}, {Danilo Jimenez Rezende}, and {Koray Kavukcuoglu}}{{Peter Battaglia}
  et~al\mbox{.}}{2016}]%
        {336}
\bibfield{author}{\bibinfo{person}{{Peter Battaglia}}, \bibinfo{person}{{Razvan
  Pascanu}}, \bibinfo{person}{{Matthew Lai}}, \bibinfo{person}{{Danilo Jimenez
  Rezende}}, {and} \bibinfo{person}{{Koray Kavukcuoglu}}.}
  \bibinfo{year}{2016}\natexlab{}.
\newblock \showarticletitle{Interaction networks for learning about objects,
  relations and physics}. In \bibinfo{booktitle}{\emph{The International
  Conference on Neural Information Processing Systems ({NIPS})}}.
  \bibinfo{pages}{4502--4510}.
\newblock


\bibitem[\protect\citeauthoryear{{Peter Meltzer}, {Marcelo Daniel Gutierrez
  Mallea}, and {Peter J. Bentley}}{{Peter Meltzer} et~al\mbox{.}}{2019}]%
        {41}
\bibfield{author}{\bibinfo{person}{{Peter Meltzer}}, \bibinfo{person}{{Marcelo
  Daniel Gutierrez Mallea}}, {and} \bibinfo{person}{{Peter J. Bentley}}.}
  \bibinfo{year}{2019}\natexlab{}.
\newblock \showarticletitle{PiNet: a permutation invariant graph neural network
  for graph classification}.
\newblock  (\bibinfo{year}{2019}).
\newblock
\urldef\tempurl%
\url{https://arxiv.org/abs/1905.03046}
\showURL{%
\tempurl}


\bibitem[\protect\citeauthoryear{{Peter W. Battaglia}, {Jessica B. Hamrick},
  {Victor Bapst}, {Alvaro Sanchez-Gonzalez}, {Vinicius Zambaldi}, {Mateusz
  Malinowski}, {Andrea Tacchetti}, {David Raposo}, {Adam Santoro}, {Ryan
  Faulkner}, {Caglar Gulcehre}, {Francis Song}, {Andrew Ballard}, {Justin
  Gilmer}, {George E. Dahl}, {Ashish Vaswani}, {Kelsey Allen}, {Charles Nash},
  {Victoria Langston}, {Chris Dyer}, {Nicolas Heess}, {Daan Wierstra},
  {Pushmeet Kohli}, {Matt Botvinick}, {Oriol Vinyals}, {Yujia Li}, and {Razvan
  Pascanu}}{{Peter W. Battaglia} et~al\mbox{.}}{2018}]%
        {85}
\bibfield{author}{\bibinfo{person}{{Peter W. Battaglia}},
  \bibinfo{person}{{Jessica B. Hamrick}}, \bibinfo{person}{{Victor Bapst}},
  \bibinfo{person}{{Alvaro Sanchez-Gonzalez}}, \bibinfo{person}{{Vinicius
  Zambaldi}}, \bibinfo{person}{{Mateusz Malinowski}}, \bibinfo{person}{{Andrea
  Tacchetti}}, \bibinfo{person}{{David Raposo}}, \bibinfo{person}{{Adam
  Santoro}}, \bibinfo{person}{{Ryan Faulkner}}, \bibinfo{person}{{Caglar
  Gulcehre}}, \bibinfo{person}{{Francis Song}}, \bibinfo{person}{{Andrew
  Ballard}}, \bibinfo{person}{{Justin Gilmer}}, \bibinfo{person}{{George E.
  Dahl}}, \bibinfo{person}{{Ashish Vaswani}}, \bibinfo{person}{{Kelsey Allen}},
  \bibinfo{person}{{Charles Nash}}, \bibinfo{person}{{Victoria Langston}},
  \bibinfo{person}{{Chris Dyer}}, \bibinfo{person}{{Nicolas Heess}},
  \bibinfo{person}{{Daan Wierstra}}, \bibinfo{person}{{Pushmeet Kohli}},
  \bibinfo{person}{{Matt Botvinick}}, \bibinfo{person}{{Oriol Vinyals}},
  \bibinfo{person}{{Yujia Li}}, {and} \bibinfo{person}{{Razvan Pascanu}}.}
  \bibinfo{year}{2018}\natexlab{}.
\newblock \showarticletitle{Relational inductive biases, deep learning, and
  graph networks}.
\newblock  (\bibinfo{year}{2018}).
\newblock
\urldef\tempurl%
\url{https://arxiv.org/abs/1806.01261}
\showURL{%
\tempurl}


\bibitem[\protect\citeauthoryear{{Qimai Li}, {Zhichao Han}, and {Xiaoming
  Wu}}{{Qimai Li} et~al\mbox{.}}{2018}]%
        {16}
\bibfield{author}{\bibinfo{person}{{Qimai Li}}, \bibinfo{person}{{Zhichao
  Han}}, {and} \bibinfo{person}{{Xiaoming Wu}}.}
  \bibinfo{year}{2018}\natexlab{}.
\newblock \showarticletitle{Deeper insights into graph convolutional networks
  for semi-supervised learning}. In \bibinfo{booktitle}{\emph{The {AAAI}
  Conference on Artificial Intelligence ({AAAI})}}.
  \bibinfo{pages}{3538--3545}.
\newblock


\bibitem[\protect\citeauthoryear{{Radford M. Neal} and {Geoffrey E.
  Hinton}}{{Radford M. Neal} and {Geoffrey E. Hinton}}{[n.d.]}]%
        {161}
\bibfield{author}{\bibinfo{person}{{Radford M. Neal}} {and}
  \bibinfo{person}{{Geoffrey E. Hinton}}.} \bibinfo{year}{[n.d.]}\natexlab{}.
\newblock \showarticletitle{A view of the EM algorithm that justifies
  incremental, sparse and other variants}.
\newblock  (\bibinfo{year}{[n.\,d.]}).
\newblock


\bibitem[\protect\citeauthoryear{{Renjie Liao}, {Marc Brockschmidt}, {Daniel
  Tarlow}, {Alexander L. Gaunt}, {Raquel Urtasun}, and {Richard Zemel}}{{Renjie
  Liao} et~al\mbox{.}}{2018}]%
        {35}
\bibfield{author}{\bibinfo{person}{{Renjie Liao}}, \bibinfo{person}{{Marc
  Brockschmidt}}, \bibinfo{person}{{Daniel Tarlow}},
  \bibinfo{person}{{Alexander L. Gaunt}}, \bibinfo{person}{{Raquel Urtasun}},
  {and} \bibinfo{person}{{Richard Zemel}}.} \bibinfo{year}{2018}\natexlab{}.
\newblock \showarticletitle{Graph partition neural networks for semi-supervised
  classification}. In \bibinfo{booktitle}{\emph{The International Conference on
  Learning Representations ({ICLR} Workshop)}}.
\newblock


\bibitem[\protect\citeauthoryear{{Renjie Liao}, {Zhizhen Zhao}, {Raquel
  Urtasun}, and {Richard Zemel}}{{Renjie Liao} et~al\mbox{.}}{2019}]%
        {14}
\bibfield{author}{\bibinfo{person}{{Renjie Liao}}, \bibinfo{person}{{Zhizhen
  Zhao}}, \bibinfo{person}{{Raquel Urtasun}}, {and} \bibinfo{person}{{Richard
  Zemel}}.} \bibinfo{year}{2019}\natexlab{}.
\newblock \showarticletitle{{LanczosNet}: Multi-scale deep graph convolutional
  networks}. In \bibinfo{booktitle}{\emph{The International Conference on
  Learning Representations ({ICLR})}}.
\newblock


\bibitem[\protect\citeauthoryear{{Rex Ying}, {Jiaxuan You}, {Christopher
  Morris}, {Xiang Ren}, {William L. Hamilton}, and {Jure Leskovec}}{{Rex Ying}
  et~al\mbox{.}}{2018}]%
        {94}
\bibfield{author}{\bibinfo{person}{{Rex Ying}}, \bibinfo{person}{{Jiaxuan
  You}}, \bibinfo{person}{{Christopher Morris}}, \bibinfo{person}{{Xiang Ren}},
  \bibinfo{person}{{William L. Hamilton}}, {and} \bibinfo{person}{{Jure
  Leskovec}}.} \bibinfo{year}{2018}\natexlab{}.
\newblock \showarticletitle{Hierarchical graph representation learning with
  differentiable pooling}. In \bibinfo{booktitle}{\emph{The International
  Conference on Nueral Information Processing Systems ({NeurPS})}}.
  \bibinfo{pages}{4805--4815}.
\newblock


\bibitem[\protect\citeauthoryear{{Richard C. Wilson}, {Edwin R. Hancock},
  {El\.{z}bieta Pekalska}, and {Robert P.W. Duin}}{{Richard C. Wilson}
  et~al\mbox{.}}{2014}]%
        {193}
\bibfield{author}{\bibinfo{person}{{Richard C. Wilson}},
  \bibinfo{person}{{Edwin R. Hancock}}, \bibinfo{person}{{El\.{z}bieta
  Pekalska}}, {and} \bibinfo{person}{{Robert P.W. Duin}}.}
  \bibinfo{year}{2014}\natexlab{}.
\newblock \showarticletitle{Spherical and Hyperbolic Embeddings of Data}.
\newblock \bibinfo{journal}{\emph{IEEE Transactions on Pattern Recognition and
  Machine Intelligence}} \bibinfo{volume}{36}, \bibinfo{number}{11}
  (\bibinfo{year}{2014}), \bibinfo{pages}{2255--2269}.
\newblock


\bibitem[\protect\citeauthoryear{{Rik Koncel-Kedziorski}, {Dhanush Bekal}, {Yi
  Luan}, {Mirella Lapata}, and {Hannaneh Hajishirzi}}{{Rik Koncel-Kedziorski}
  et~al\mbox{.}}{2019}]%
        {326}
\bibfield{author}{\bibinfo{person}{{Rik Koncel-Kedziorski}},
  \bibinfo{person}{{Dhanush Bekal}}, \bibinfo{person}{{Yi Luan}},
  \bibinfo{person}{{Mirella Lapata}}, {and} \bibinfo{person}{{Hannaneh
  Hajishirzi}}.} \bibinfo{year}{2019}\natexlab{}.
\newblock \showarticletitle{Text generation from knowledge graphs with graph
  transformers}.
\newblock  (\bibinfo{year}{2019}).
\newblock
\urldef\tempurl%
\url{https://arxiv.org/abs/1904.02342}
\showURL{%
\tempurl}


\bibitem[\protect\citeauthoryear{{Risi Kondor}, {Truong Son Hy}, {Horace Pan},
  {Brandon M. Anderson}, and {Shubhendu Trivedi}}{{Risi Kondor}
  et~al\mbox{.}}{2018}]%
        {213}
\bibfield{author}{\bibinfo{person}{{Risi Kondor}}, \bibinfo{person}{{Truong Son
  Hy}}, \bibinfo{person}{{Horace Pan}}, \bibinfo{person}{{Brandon M.
  Anderson}}, {and} \bibinfo{person}{{Shubhendu Trivedi}}.}
  \bibinfo{year}{2018}\natexlab{}.
\newblock \showarticletitle{Covariant compositional networks for learning
  graphs}. In \bibinfo{booktitle}{\emph{The International Conference on
  Learning Representations ({ICLR} Workshop)}}.
\newblock


\bibitem[\protect\citeauthoryear{{Ron Levie}, {Federico Monti}, {Xavier
  Bresson}, and {Michael M. Bronstein}}{{Ron Levie} et~al\mbox{.}}{2019}]%
        {12}
\bibfield{author}{\bibinfo{person}{{Ron Levie}}, \bibinfo{person}{{Federico
  Monti}}, \bibinfo{person}{{Xavier Bresson}}, {and} \bibinfo{person}{{Michael
  M. Bronstein}}.} \bibinfo{year}{2019}\natexlab{}.
\newblock \showarticletitle{{CayleyNets}: Graph Convolutional Neural Networks
  with Complex Relational Spectral Filters}.
\newblock \bibinfo{journal}{\emph{{IEEE} Transactions on Signal Processing}}
  \bibinfo{volume}{67}, \bibinfo{number}{1} (\bibinfo{year}{2019}),
  \bibinfo{pages}{97--109}.
\newblock


\bibitem[\protect\citeauthoryear{{Rui Dai}, {Shenkun Xu}, {Qian Gu}, {Chenguang
  Ji}, and {Kaikui Liu}}{{Rui Dai} et~al\mbox{.}}{2020}]%
        {403}
\bibfield{author}{\bibinfo{person}{{Rui Dai}}, \bibinfo{person}{{Shenkun Xu}},
  \bibinfo{person}{{Qian Gu}}, \bibinfo{person}{{Chenguang Ji}}, {and}
  \bibinfo{person}{{Kaikui Liu}}.} \bibinfo{year}{2020}\natexlab{}.
\newblock \showarticletitle{Hybrid spatio-temporal graph convolutional network:
  improving traffic prediction with navigation data}. In
  \bibinfo{booktitle}{\emph{The International Conference on Knowledge Discovery
  and Data Mining (SIGKDD)}}. \bibinfo{pages}{3074--3082}.
\newblock


\bibitem[\protect\citeauthoryear{{Ruoyu Li}, {Sheng Wang}, {Feiyun Zhu}, and
  {Junzhou Huang}}{{Ruoyu Li} et~al\mbox{.}}{2018}]%
        {8}
\bibfield{author}{\bibinfo{person}{{Ruoyu Li}}, \bibinfo{person}{{Sheng Wang}},
  \bibinfo{person}{{Feiyun Zhu}}, {and} \bibinfo{person}{{Junzhou Huang}}.}
  \bibinfo{year}{2018}\natexlab{}.
\newblock \showarticletitle{Adaptive Graph Convolutional Neural Networks}. In
  \bibinfo{booktitle}{\emph{The {AAAI} Conference on Artificial Intelligence
  ({AAAI})}}. \bibinfo{pages}{3546--3553}.
\newblock


\bibitem[\protect\citeauthoryear{{Rupesh Kumar Srivastava}, {Klaus Greff}, and
  {J\"{u}rgen Schmidhuber}}{{Rupesh Kumar Srivastava} et~al\mbox{.}}{2015}]%
        {399}
\bibfield{author}{\bibinfo{person}{{Rupesh Kumar Srivastava}},
  \bibinfo{person}{{Klaus Greff}}, {and} \bibinfo{person}{{J\"{u}rgen
  Schmidhuber}}.} \bibinfo{year}{2015}\natexlab{}.
\newblock \showarticletitle{Highways networks}.
\newblock  (\bibinfo{year}{2015}).
\newblock
\urldef\tempurl%
\url{https://arxiv.org/abs/1505.00387}
\showURL{%
\tempurl}


\bibitem[\protect\citeauthoryear{{Ryan Murphy}, {Balasubramaniam Srinivasan},
  {Vinayak Rao}, and {Bruno Ribeiro}}{{Ryan Murphy} et~al\mbox{.}}{2019}]%
        {98}
\bibfield{author}{\bibinfo{person}{{Ryan Murphy}},
  \bibinfo{person}{{Balasubramaniam Srinivasan}}, \bibinfo{person}{{Vinayak
  Rao}}, {and} \bibinfo{person}{{Bruno Ribeiro}}.}
  \bibinfo{year}{2019}\natexlab{}.
\newblock \showarticletitle{Relational pooling for graph representations}. In
  \bibinfo{booktitle}{\emph{The International Conference on Machine Learning
  ({ICML})}}. \bibinfo{pages}{4663--4673}.
\newblock


\bibitem[\protect\citeauthoryear{{Ryoma Sato}}{{Ryoma Sato}}{2020}]%
        {360}
\bibfield{author}{\bibinfo{person}{{Ryoma Sato}}.}
  \bibinfo{year}{2020}\natexlab{}.
\newblock \showarticletitle{A survey on the expressive power of graph neural
  networks}.
\newblock  (\bibinfo{year}{2020}).
\newblock
\urldef\tempurl%
\url{https://arxiv.org/abs/2003.04078}
\showURL{%
\tempurl}


\bibitem[\protect\citeauthoryear{{Ryoma Sato}, {Makoto Yamada}, and {Hisashi
  Kashima}}{{Ryoma Sato} et~al\mbox{.}}{2019}]%
        {223}
\bibfield{author}{\bibinfo{person}{{Ryoma Sato}}, \bibinfo{person}{{Makoto
  Yamada}}, {and} \bibinfo{person}{{Hisashi Kashima}}.}
  \bibinfo{year}{2019}\natexlab{}.
\newblock \showarticletitle{Approximation Ratios of Graph Neural Networks for
  Combinatorial Problems}. In \bibinfo{booktitle}{\emph{The International
  Conference on Neural Information Processing Systems ({NeurPS})}}.
  \bibinfo{pages}{4081--4090}.
\newblock


\bibitem[\protect\citeauthoryear{{Sami Abu-EL-Haija}, {Amol Kapoor}, {Bryan
  Perozzi}, and {Joonseok Lee}}{{Sami Abu-EL-Haija} et~al\mbox{.}}{2019a}]%
        {375}
\bibfield{author}{\bibinfo{person}{{Sami Abu-EL-Haija}}, \bibinfo{person}{{Amol
  Kapoor}}, \bibinfo{person}{{Bryan Perozzi}}, {and} \bibinfo{person}{{Joonseok
  Lee}}.} \bibinfo{year}{2019}\natexlab{a}.
\newblock \showarticletitle{{N-GCN}: multi-scale graph convolution for
  semi-supervised node classification}. In \bibinfo{booktitle}{\emph{The
  Conference on Uncertainty in Artificial Intelligence (UAI)}}.
  \bibinfo{pages}{No. 310}.
\newblock


\bibitem[\protect\citeauthoryear{{Sami Abu-EL-Haija}, {Bryan Perozzi}, {Amol
  Kapoor}, {Nazanin Alipourfard}, {Kristina Lerman}, {Hrayr Harutyunyan}, {Greg
  Ver Steeg}, and {Aram Galstyan}}{{Sami Abu-EL-Haija} et~al\mbox{.}}{2019b}]%
        {90}
\bibfield{author}{\bibinfo{person}{{Sami Abu-EL-Haija}},
  \bibinfo{person}{{Bryan Perozzi}}, \bibinfo{person}{{Amol Kapoor}},
  \bibinfo{person}{{Nazanin Alipourfard}}, \bibinfo{person}{{Kristina Lerman}},
  \bibinfo{person}{{Hrayr Harutyunyan}}, \bibinfo{person}{{Greg Ver Steeg}},
  {and} \bibinfo{person}{{Aram Galstyan}}.} \bibinfo{year}{2019}\natexlab{b}.
\newblock \showarticletitle{{MixHop}: higher-order graph convolutional
  architectures via sparsified neighborhood mixing}. In
  \bibinfo{booktitle}{\emph{The International Conference on Machine Learning
  ({ICML})}}. \bibinfo{pages}{21--29}.
\newblock


\bibitem[\protect\citeauthoryear{{Sara Sabour}, {Nicholas Frosst}, and
  {Geoffrey E. Hinton}}{{Sara Sabour} et~al\mbox{.}}{2017}]%
        {400}
\bibfield{author}{\bibinfo{person}{{Sara Sabour}}, \bibinfo{person}{{Nicholas
  Frosst}}, {and} \bibinfo{person}{{Geoffrey E. Hinton}}.}
  \bibinfo{year}{2017}\natexlab{}.
\newblock \showarticletitle{Dynamic routing between capsules}. In
  \bibinfo{booktitle}{\emph{The International Conference on Neural Information
  Processing Systems (NIPS)}}. \bibinfo{pages}{3856--3866}.
\newblock


\bibitem[\protect\citeauthoryear{{Saurabh Verma} and {Zhili Zhang}}{{Saurabh
  Verma} and {Zhili Zhang}}{2018}]%
        {53}
\bibfield{author}{\bibinfo{person}{{Saurabh Verma}} {and}
  \bibinfo{person}{{Zhili Zhang}}.} \bibinfo{year}{2018}\natexlab{}.
\newblock \showarticletitle{Graph Capsule Convolutional Neural Networks}.
\newblock  (\bibinfo{year}{2018}).
\newblock
\urldef\tempurl%
\url{https://arxiv.org/abs/1805.08090}
\showURL{%
\tempurl}


\bibitem[\protect\citeauthoryear{{S\'\{e\}bastien Lerique}, {Jacob Levy
  Abitol}, and {M\'{a}rton Karsai}}{{S\'\{e\}bastien Lerique}
  et~al\mbox{.}}{2020}]%
        {127}
\bibfield{author}{\bibinfo{person}{{S\'\{e\}bastien Lerique}},
  \bibinfo{person}{{Jacob Levy Abitol}}, {and} \bibinfo{person}{{M\'{a}rton
  Karsai}}.} \bibinfo{year}{2020}\natexlab{}.
\newblock \showarticletitle{Joint embedding of structure and features via graph
  convolutional networks}.
\newblock \bibinfo{journal}{\emph{Applied Network Science}}
  \bibinfo{volume}{5}, \bibinfo{number}{2020} (\bibinfo{year}{2020}),
  \bibinfo{pages}{No. 5}.
\newblock


\bibitem[\protect\citeauthoryear{{Seongjun Yun}, {Minbyul Jeong}, {Raehyun
  Kim}, {Jaewoo Kang}, and {Hyunwoo J. Kim}}{{Seongjun Yun}
  et~al\mbox{.}}{2019}]%
        {60}
\bibfield{author}{\bibinfo{person}{{Seongjun Yun}}, \bibinfo{person}{{Minbyul
  Jeong}}, \bibinfo{person}{{Raehyun Kim}}, \bibinfo{person}{{Jaewoo Kang}},
  {and} \bibinfo{person}{{Hyunwoo J. Kim}}.} \bibinfo{year}{2019}\natexlab{}.
\newblock \showarticletitle{Graph Transformer Networks}.
\newblock  (\bibinfo{year}{2019}).
\newblock
\urldef\tempurl%
\url{https://arxiv.org/abs/1911.06455}
\showURL{%
\tempurl}


\bibitem[\protect\citeauthoryear{{Sepp Hochreiter} and {J\"{u}rgen
  Schmidhuber}}{{Sepp Hochreiter} and {J\"{u}rgen Schmidhuber}}{1997}]%
        {160}
\bibfield{author}{\bibinfo{person}{{Sepp Hochreiter}} {and}
  \bibinfo{person}{{J\"{u}rgen Schmidhuber}}.} \bibinfo{year}{1997}\natexlab{}.
\newblock \showarticletitle{Long short-term memory}.
\newblock \bibinfo{journal}{\emph{Neural Computation}} \bibinfo{volume}{9},
  \bibinfo{number}{8} (\bibinfo{year}{1997}), \bibinfo{pages}{1735--1780}.
\newblock


\bibitem[\protect\citeauthoryear{{Shaohua Fan}, {Junxiong Zhu}, {Xiaotian Han},
  {Chuan Shi}, {Linmei Hu}, {Biyu Ma}, and {Yongliang Li}}{{Shaohua Fan}
  et~al\mbox{.}}{2019}]%
        {345}
\bibfield{author}{\bibinfo{person}{{Shaohua Fan}}, \bibinfo{person}{{Junxiong
  Zhu}}, \bibinfo{person}{{Xiaotian Han}}, \bibinfo{person}{{Chuan Shi}},
  \bibinfo{person}{{Linmei Hu}}, \bibinfo{person}{{Biyu Ma}}, {and}
  \bibinfo{person}{{Yongliang Li}}.} \bibinfo{year}{2019}\natexlab{}.
\newblock \showarticletitle{Metapath-guided heterogeneous graph neural network
  for intent recommendation}. In \bibinfo{booktitle}{\emph{The International
  Conference on Knowledge Discovery and Data Mining ({SIGKDD})}}.
  \bibinfo{pages}{2478--2486}.
\newblock


\bibitem[\protect\citeauthoryear{{Shaosheng Cao}, {Wei Lu}, and {Qiongkai
  Xu}}{{Shaosheng Cao} et~al\mbox{.}}{2016}]%
        {124}
\bibfield{author}{\bibinfo{person}{{Shaosheng Cao}}, \bibinfo{person}{{Wei
  Lu}}, {and} \bibinfo{person}{{Qiongkai Xu}}.}
  \bibinfo{year}{2016}\natexlab{}.
\newblock \showarticletitle{Deep neural networks for learning graph
  representations}. In \bibinfo{booktitle}{\emph{The {AAAI} Conference on
  Artificial Intelligence ({AAAI})}}. \bibinfo{pages}{1145--1152}.
\newblock


\bibitem[\protect\citeauthoryear{{Shengding Hu}, {Meng Qu}, {Zhiyuan Liu}, and
  {Jian Tang}}{{Shengding Hu} et~al\mbox{.}}{2020}]%
        {45}
\bibfield{author}{\bibinfo{person}{{Shengding Hu}}, \bibinfo{person}{{Meng
  Qu}}, \bibinfo{person}{{Zhiyuan Liu}}, {and} \bibinfo{person}{{Jian Tang}}.}
  \bibinfo{year}{2020}\natexlab{}.
\newblock \showarticletitle{Transfer active learning for graph neural
  networks}.
\newblock  (\bibinfo{year}{2020}).
\newblock
\urldef\tempurl%
\url{https://openreview.net/forum?i}
\showURL{%
\tempurl}


\bibitem[\protect\citeauthoryear{{Shengnan Guo}, {Youfang Lin}, {Ning Feng},
  {Chao Song}, and {Huaiyu Wan}}{{Shengnan Guo} et~al\mbox{.}}{2019}]%
        {71}
\bibfield{author}{\bibinfo{person}{{Shengnan Guo}}, \bibinfo{person}{{Youfang
  Lin}}, \bibinfo{person}{{Ning Feng}}, \bibinfo{person}{{Chao Song}}, {and}
  \bibinfo{person}{{Huaiyu Wan}}.} \bibinfo{year}{2019}\natexlab{}.
\newblock \showarticletitle{Attention based spatial-temporal graph
  convolutional networks for traffic flow forecasting}. In
  \bibinfo{booktitle}{\emph{The {AAAI} Conference on Artificial Intelligence
  ({AAAI})}}. \bibinfo{pages}{pages = {922--929},}.
\newblock


\bibitem[\protect\citeauthoryear{{Shirui Pan}, {Ruiqi Hu}, {Guodong Long},
  {Jing Jiang}, {Lina Yao}, and {Chengqi Zhang}}{{Shirui Pan}
  et~al\mbox{.}}{2018}]%
        {129}
\bibfield{author}{\bibinfo{person}{{Shirui Pan}}, \bibinfo{person}{{Ruiqi Hu}},
  \bibinfo{person}{{Guodong Long}}, \bibinfo{person}{{Jing Jiang}},
  \bibinfo{person}{{Lina Yao}}, {and} \bibinfo{person}{{Chengqi Zhang}}.}
  \bibinfo{year}{2018}\natexlab{}.
\newblock \showarticletitle{Adversarially regularized graph autoencoder for
  graph embedding}. In \bibinfo{booktitle}{\emph{The International Joint
  Conference on Artificial Intelligence ({IJCAI})}}.
  \bibinfo{pages}{2609--2615}.
\newblock


\bibitem[\protect\citeauthoryear{{Shu Wu}, {Yuyuan Tang}, {Yanqiao Zhu}, {Liang
  Wang}, {Xing Xie}, and {Tieniu Tan}}{{Shu Wu} et~al\mbox{.}}{2019}]%
        {343}
\bibfield{author}{\bibinfo{person}{{Shu Wu}}, \bibinfo{person}{{Yuyuan Tang}},
  \bibinfo{person}{{Yanqiao Zhu}}, \bibinfo{person}{{Liang Wang}},
  \bibinfo{person}{{Xing Xie}}, {and} \bibinfo{person}{{Tieniu Tan}}.}
  \bibinfo{year}{2019}\natexlab{}.
\newblock \showarticletitle{Session-based recommendation with graph neural
  networks}. In \bibinfo{booktitle}{\emph{The {AAAI} Conference on Artificial
  Intelligence ({AAAI})}}. \bibinfo{pages}{346--353}.
\newblock


\bibitem[\protect\citeauthoryear{{Sijie Yan}, {Yuanjun Xiong}, and {Dahua
  Lin}}{{Sijie Yan} et~al\mbox{.}}{2018}]%
        {55}
\bibfield{author}{\bibinfo{person}{{Sijie Yan}}, \bibinfo{person}{{Yuanjun
  Xiong}}, {and} \bibinfo{person}{{Dahua Lin}}.}
  \bibinfo{year}{2018}\natexlab{}.
\newblock \showarticletitle{Spatial temporal graph convolutional networks for
  skeleton-based action recognition}. In \bibinfo{booktitle}{\emph{The {AAAI}
  Conference on Artificial Intelligence ({AAAI})}}.
  \bibinfo{pages}{7444--7452}.
\newblock


\bibitem[\protect\citeauthoryear{{Sofia Ira Ktena}, {Sarah Parisot}, {Enzo
  Ferrante}, {Martin Rajchl}, {Matthew Lee}, {Ben Glocker}, and {Daniel
  Ruekert}}{{Sofia Ira Ktena} et~al\mbox{.}}{2017}]%
        {205}
\bibfield{author}{\bibinfo{person}{{Sofia Ira Ktena}}, \bibinfo{person}{{Sarah
  Parisot}}, \bibinfo{person}{{Enzo Ferrante}}, \bibinfo{person}{{Martin
  Rajchl}}, \bibinfo{person}{{Matthew Lee}}, \bibinfo{person}{{Ben Glocker}},
  {and} \bibinfo{person}{{Daniel Ruekert}}.} \bibinfo{year}{2017}\natexlab{}.
\newblock \showarticletitle{Distance metric learning using graph convolutional
  networks: application to functional brain networks}. In
  \bibinfo{booktitle}{\emph{The International Conference on Medical Image
  Computing and Computer-Assisted Intervention ({MICCAI})}}.
  \bibinfo{pages}{469--477}.
\newblock


\bibitem[\protect\citeauthoryear{{Songtao He}, {Favyen Bastani}, {Satvat
  Jagwani}, {Edward Park}, {Sofiane Abbar}, {Mohammad Alizadeh}, {Hari
  Balakrishnan}, {Sanjay Chawla}, {Samuel Madden}, and {Mohammad Amin
  Sadeghi}}{{Songtao He} et~al\mbox{.}}{2020}]%
        {355}
\bibfield{author}{\bibinfo{person}{{Songtao He}}, \bibinfo{person}{{Favyen
  Bastani}}, \bibinfo{person}{{Satvat Jagwani}}, \bibinfo{person}{{Edward
  Park}}, \bibinfo{person}{{Sofiane Abbar}}, \bibinfo{person}{{Mohammad
  Alizadeh}}, \bibinfo{person}{{Hari Balakrishnan}}, \bibinfo{person}{{Sanjay
  Chawla}}, \bibinfo{person}{{Samuel Madden}}, {and} \bibinfo{person}{{Mohammad
  Amin Sadeghi}}.} \bibinfo{year}{2020}\natexlab{}.
\newblock \showarticletitle{{RoadTagger}: robust road attribute inference with
  graph neural networks}. In \bibinfo{booktitle}{\emph{The {AAAI} Conference on
  Artificial Intelligence ({AAAI})}}.
\newblock


\bibitem[\protect\citeauthoryear{{Steven Kearnes}, {Kevin McCloskey}, {Marc
  Berndl}, {Vijay S. Pande}, and {Patrick F. Riley}}{{Steven Kearnes}
  et~al\mbox{.}}{2016}]%
        {195}
\bibfield{author}{\bibinfo{person}{{Steven Kearnes}}, \bibinfo{person}{{Kevin
  McCloskey}}, \bibinfo{person}{{Marc Berndl}}, \bibinfo{person}{{Vijay S.
  Pande}}, {and} \bibinfo{person}{{Patrick F. Riley}}.}
  \bibinfo{year}{2016}\natexlab{}.
\newblock \showarticletitle{Molecular graph convolutions: moving beyond
  fingerprints}.
\newblock \bibinfo{journal}{\emph{Journal of Computer-Aided Molecular Design}}
  \bibinfo{volume}{30}, \bibinfo{number}{8} (\bibinfo{year}{2016}),
  \bibinfo{pages}{595--608}.
\newblock


\bibitem[\protect\citeauthoryear{{Sungmin Rhee}, {Seokjun Seo}, and {Sun
  Kim}}{{Sungmin Rhee} et~al\mbox{.}}{2018}]%
        {91}
\bibfield{author}{\bibinfo{person}{{Sungmin Rhee}}, \bibinfo{person}{{Seokjun
  Seo}}, {and} \bibinfo{person}{{Sun Kim}}.} \bibinfo{year}{2018}\natexlab{}.
\newblock \showarticletitle{Hybrid approach of relation network and localized
  graph convolutional filtering for breast cancer subtype classification}. In
  \bibinfo{booktitle}{\emph{Th International Joint Conference on Artificial
  Intelligence ({IJCAI})}}. \bibinfo{pages}{3527--3534}.
\newblock


\bibitem[\protect\citeauthoryear{{Tengfei Ma}, Shang, and {Jimeng
  Sun}}{{Tengfei Ma} et~al\mbox{.}}{2019}]%
        {159}
\bibfield{author}{\bibinfo{person}{{Tengfei Ma}}, \bibinfo{person}{Junyuan
  Shang}, {and} \bibinfo{person}{{Jimeng Sun}}.}
  \bibinfo{year}{2019}\natexlab{}.
\newblock \showarticletitle{CGNF: conditional graph neural fields}.
\newblock  (\bibinfo{year}{2019}).
\newblock
\urldef\tempurl%
\url{https://openreview.net/forum?i}
\showURL{%
\tempurl}


\bibitem[\protect\citeauthoryear{{Thomas N. Kipf} and {Max Welling}}{{Thomas N.
  Kipf} and {Max Welling}}{2016}]%
        {87}
\bibfield{author}{\bibinfo{person}{{Thomas N. Kipf}} {and}
  \bibinfo{person}{{Max Welling}}.} \bibinfo{year}{2016}\natexlab{}.
\newblock \showarticletitle{Variational Graph Auto-Encoders}.
\newblock  (\bibinfo{year}{2016}).
\newblock
\urldef\tempurl%
\url{https://arxiv.org/abs/1611.07308}
\showURL{%
\tempurl}


\bibitem[\protect\citeauthoryear{{Thomas N. Kipf} and {Max Welling}}{{Thomas N.
  Kipf} and {Max Welling}}{2017}]%
        {6}
\bibfield{author}{\bibinfo{person}{{Thomas N. Kipf}} {and}
  \bibinfo{person}{{Max Welling}}.} \bibinfo{year}{2017}\natexlab{}.
\newblock \showarticletitle{Semi-supervised classification with graph
  convolutional networks}. In \bibinfo{booktitle}{\emph{The International
  Conference on Learning Representations ({ICLR})}}.
\newblock


\bibitem[\protect\citeauthoryear{{Tingwu Wang}, {Renjie Liao}, {Jimmy Ba}, and
  {Sanja Fidler}}{{Tingwu Wang} et~al\mbox{.}}{2018}]%
        {187}
\bibfield{author}{\bibinfo{person}{{Tingwu Wang}}, \bibinfo{person}{{Renjie
  Liao}}, \bibinfo{person}{{Jimmy Ba}}, {and} \bibinfo{person}{{Sanja
  Fidler}}.} \bibinfo{year}{2018}\natexlab{}.
\newblock \showarticletitle{{NerveNet}: learning structured policy with graph
  neural networks}. In \bibinfo{booktitle}{\emph{The International Conference
  on Learning Representations ({ICLR})}}.
\newblock


\bibitem[\protect\citeauthoryear{{Tong Zhang}, {Wenming Zheng}, {Zhen Cui}, and
  {Yang Li}}{{Tong Zhang} et~al\mbox{.}}{2018}]%
        {37}
\bibfield{author}{\bibinfo{person}{{Tong Zhang}}, \bibinfo{person}{{Wenming
  Zheng}}, \bibinfo{person}{{Zhen Cui}}, {and} \bibinfo{person}{{Yang Li}}.}
  \bibinfo{year}{2018}\natexlab{}.
\newblock \showarticletitle{Tensor graph convolutional neural network}.
\newblock  (\bibinfo{year}{2018}).
\newblock
\urldef\tempurl%
\url{https://arxiv.org/abs/1803.10071v1}
\showURL{%
\tempurl}


\bibitem[\protect\citeauthoryear{{Trang Pham}, {Truyen Tran}, {Dinh Phung}, and
  {Svetha Venkatesh}}{{Trang Pham} et~al\mbox{.}}{2017}]%
        {29}
\bibfield{author}{\bibinfo{person}{{Trang Pham}}, \bibinfo{person}{{Truyen
  Tran}}, \bibinfo{person}{{Dinh Phung}}, {and} \bibinfo{person}{{Svetha
  Venkatesh}}.} \bibinfo{year}{2017}\natexlab{}.
\newblock \showarticletitle{Column Networks for Collective Classification}. In
  \bibinfo{booktitle}{\emph{The {AAAI} Conference on Artificial Intelligence
  ({AAAI})}}. \bibinfo{pages}{2485--2491}.
\newblock


\bibitem[\protect\citeauthoryear{{Tyler Derr}, {Yao Ma}, and {Jiliang
  Tang}}{{Tyler Derr} et~al\mbox{.}}{2018}]%
        {28}
\bibfield{author}{\bibinfo{person}{{Tyler Derr}}, \bibinfo{person}{{Yao Ma}},
  {and} \bibinfo{person}{{Jiliang Tang}}.} \bibinfo{year}{2018}\natexlab{}.
\newblock \showarticletitle{Signed Graph Convolutional Network}. In
  \bibinfo{booktitle}{\emph{The {IEEE} International Conference on Data Mining
  ({ICDM})}}. \bibinfo{pages}{929--934}.
\newblock


\bibitem[\protect\citeauthoryear{{Victoria Zayats} and {Mari
  Ostendorf}}{{Victoria Zayats} and {Mari Ostendorf}}{2018}]%
        {117}
\bibfield{author}{\bibinfo{person}{{Victoria Zayats}} {and}
  \bibinfo{person}{{Mari Ostendorf}}.} \bibinfo{year}{2018}\natexlab{}.
\newblock \showarticletitle{Conversation modeling on Reddit using a
  graph-structured LSTM}.
\newblock \bibinfo{journal}{\emph{Transactions of the Association for
  Computational Linguistics}} \bibinfo{volume}{6}, \bibinfo{number}{2018}
  (\bibinfo{year}{2018}), \bibinfo{pages}{121--132}.
\newblock


\bibitem[\protect\citeauthoryear{{Vijay Prakash Dwivedi}, {Chaitanya K. Joshi},
  {Thomas Laurent}, {Yoshua Bengio}, and {Xavier Bresson}}{{Vijay Prakash
  Dwivedi} et~al\mbox{.}}{2020}]%
        {169}
\bibfield{author}{\bibinfo{person}{{Vijay Prakash Dwivedi}},
  \bibinfo{person}{{Chaitanya K. Joshi}}, \bibinfo{person}{{Thomas Laurent}},
  \bibinfo{person}{{Yoshua Bengio}}, {and} \bibinfo{person}{{Xavier Bresson}}.}
  \bibinfo{year}{2020}\natexlab{}.
\newblock \showarticletitle{Benchmarking graph neural networks}.
\newblock  (\bibinfo{year}{2020}).
\newblock
\urldef\tempurl%
\url{https://arxiv.org/abs/2003.00982}
\showURL{%
\tempurl}


\bibitem[\protect\citeauthoryear{{Vikas K. Garg}, {Stefanie Jegelka}, and
  {Tommi Jaakkola}}{{Vikas K. Garg} et~al\mbox{.}}{2020}]%
        {407}
\bibfield{author}{\bibinfo{person}{{Vikas K. Garg}}, \bibinfo{person}{{Stefanie
  Jegelka}}, {and} \bibinfo{person}{{Tommi Jaakkola}}.}
  \bibinfo{year}{2020}\natexlab{}.
\newblock \showarticletitle{Generalization and representational limits of graph
  neural networks}.
\newblock  (\bibinfo{year}{2020}).
\newblock
\urldef\tempurl%
\url{https://arxiv.org/abs/2002.06157}
\showURL{%
\tempurl}


\bibitem[\protect\citeauthoryear{{Vikas Verma}, {Meng Qu}, {Alex Lamb}, {Yoshua
  Bengio}, {Juho Kannala}, and {Jian Tang}}{{Vikas Verma}
  et~al\mbox{.}}{2019}]%
        {179}
\bibfield{author}{\bibinfo{person}{{Vikas Verma}}, \bibinfo{person}{{Meng Qu}},
  \bibinfo{person}{{Alex Lamb}}, \bibinfo{person}{{Yoshua Bengio}},
  \bibinfo{person}{{Juho Kannala}}, {and} \bibinfo{person}{{Jian Tang}}.}
  \bibinfo{year}{2019}\natexlab{}.
\newblock \showarticletitle{GraphMix: regularized training of graph neural
  networks for semi-supervised learning}.
\newblock  (\bibinfo{year}{2019}).
\newblock
\urldef\tempurl%
\url{https://arxiv.org/abs/1909.11715}
\showURL{%
\tempurl}


\bibitem[\protect\citeauthoryear{{Vincenzo Di Massa}, {Cabriele Monfardini},
  {Lorenzo Sarti}, {Franco Scarselli}, {Marco Maggini}, and {Marco
  Gori}}{{Vincenzo Di Massa} et~al\mbox{.}}{2006}]%
        {47}
\bibfield{author}{\bibinfo{person}{{Vincenzo Di Massa}},
  \bibinfo{person}{{Cabriele Monfardini}}, \bibinfo{person}{{Lorenzo Sarti}},
  \bibinfo{person}{{Franco Scarselli}}, \bibinfo{person}{{Marco Maggini}},
  {and} \bibinfo{person}{{Marco Gori}}.} \bibinfo{year}{2006}\natexlab{}.
\newblock \showarticletitle{A comparison between recursive neural networks and
  graph neural networks}. In \bibinfo{booktitle}{\emph{The {IEEE} Joint
  Conference on Neural Networks ({IJCNN})}}. \bibinfo{pages}{778--785}.
\newblock


\bibitem[\protect\citeauthoryear{{Volodymyr Minh}, {Nicolas Heess}, {Alex
  Graves}, and {Koray Kavukcuoglu}}{{Volodymyr Minh} et~al\mbox{.}}{2014}]%
        {393}
\bibfield{author}{\bibinfo{person}{{Volodymyr Minh}}, \bibinfo{person}{{Nicolas
  Heess}}, \bibinfo{person}{{Alex Graves}}, {and} \bibinfo{person}{{Koray
  Kavukcuoglu}}.} \bibinfo{year}{2014}\natexlab{}.
\newblock \showarticletitle{Recurrent models of visual attention}. In
  \bibinfo{booktitle}{\emph{The International Conference on Neural Information
  Processing Systems (NIPS)}}. \bibinfo{pages}{2204--2212}.
\newblock


\bibitem[\protect\citeauthoryear{{Wei Li}, {Shaogang Gong}, and {Shaogang
  Gong}}{{Wei Li} et~al\mbox{.}}{2020}]%
        {171}
\bibfield{author}{\bibinfo{person}{{Wei Li}}, \bibinfo{person}{{Shaogang
  Gong}}, {and} \bibinfo{person}{{Shaogang Gong}}.}
  \bibinfo{year}{2020}\natexlab{}.
\newblock \showarticletitle{Neural graph embedding for neural architecture
  search}. In \bibinfo{booktitle}{\emph{The {AAAI} Conference on Artificial
  Intelligence ({AAAI})}}.
\newblock


\bibitem[\protect\citeauthoryear{{Wei Liu} and {Sanjay Chawla}}{{Wei Liu} and
  {Sanjay Chawla}}{2009}]%
        {364}
\bibfield{author}{\bibinfo{person}{{Wei Liu}} {and} \bibinfo{person}{{Sanjay
  Chawla}}.} \bibinfo{year}{2009}\natexlab{}.
\newblock \showarticletitle{A game theoretical model for adversarial learning}.
  In \bibinfo{booktitle}{\emph{The {IEEE} International Conference on Data
  Mining Workshops ({ICDM} Workshop)}}. \bibinfo{pages}{25--30}.
\newblock


\bibitem[\protect\citeauthoryear{{Weihua Hu}, {Bowen Liu}, {Joseph Gomes},
  {Marinka Zitnik}, {Percy Liang}, {Vijay S. Pande}, and {Jure
  Leskovec}}{{Weihua Hu} et~al\mbox{.}}{2020a}]%
        {130}
\bibfield{author}{\bibinfo{person}{{Weihua Hu}}, \bibinfo{person}{{Bowen Liu}},
  \bibinfo{person}{{Joseph Gomes}}, \bibinfo{person}{{Marinka Zitnik}},
  \bibinfo{person}{{Percy Liang}}, \bibinfo{person}{{Vijay S. Pande}}, {and}
  \bibinfo{person}{{Jure Leskovec}}.} \bibinfo{year}{2020}\natexlab{a}.
\newblock \showarticletitle{Strategies for pre-training graph neural networks}.
  In \bibinfo{booktitle}{\emph{The International Conference on Learning
  Representations ({ICLR})}}.
\newblock


\bibitem[\protect\citeauthoryear{{Weihua Hu}, {Matthias Fey}, {Marinka Zitnik},
  {Yuxiao Dong}, {Hongyu Ren}, {Bowen Liu}, {Michele Catasta}, and {Jure
  Leskovec}}{{Weihua Hu} et~al\mbox{.}}{2020b}]%
        {372}
\bibfield{author}{\bibinfo{person}{{Weihua Hu}}, \bibinfo{person}{{Matthias
  Fey}}, \bibinfo{person}{{Marinka Zitnik}}, \bibinfo{person}{{Yuxiao Dong}},
  \bibinfo{person}{{Hongyu Ren}}, \bibinfo{person}{{Bowen Liu}},
  \bibinfo{person}{{Michele Catasta}}, {and} \bibinfo{person}{{Jure
  Leskovec}}.} \bibinfo{year}{2020}\natexlab{b}.
\newblock \showarticletitle{Open graph benchmark: datasets for machine learning
  on graphs}.
\newblock  (\bibinfo{year}{2020}).
\newblock
\urldef\tempurl%
\url{https://arxiv.org/abs/2005.00687}
\showURL{%
\tempurl}


\bibitem[\protect\citeauthoryear{{Weilin Chiang}, {Xuanqing Liu}, {Si Si},
  {Yang Li}, {Samy Bengio}, and {Cho-Jui Hsieh}}{{Weilin Chiang}
  et~al\mbox{.}}{2019}]%
        {23}
\bibfield{author}{\bibinfo{person}{{Weilin Chiang}}, \bibinfo{person}{{Xuanqing
  Liu}}, \bibinfo{person}{{Si Si}}, \bibinfo{person}{{Yang Li}},
  \bibinfo{person}{{Samy Bengio}}, {and} \bibinfo{person}{{Cho-Jui Hsieh}}.}
  \bibinfo{year}{2019}\natexlab{}.
\newblock \showarticletitle{Cluster-{GCN}: an efficient algorithm for training
  deep and large graph convolutional networks}. In
  \bibinfo{booktitle}{\emph{The {ACM} International Conference on Knowledge
  Discovery and Data Mining ({SIGKDD})}}. \bibinfo{pages}{257--266}.
\newblock


\bibitem[\protect\citeauthoryear{{Wenbing Huang}, {Tong Zhang}, {Yu Rong}, and
  {Junzhou Huang}}{{Wenbing Huang} et~al\mbox{.}}{2018}]%
        {9}
\bibfield{author}{\bibinfo{person}{{Wenbing Huang}}, \bibinfo{person}{{Tong
  Zhang}}, \bibinfo{person}{{Yu Rong}}, {and} \bibinfo{person}{{Junzhou
  Huang}}.} \bibinfo{year}{2018}\natexlab{}.
\newblock \showarticletitle{Adaptive sampling towards fast graph representation
  learning}. In \bibinfo{booktitle}{\emph{The International Conference on
  Neural Information Processing Systems ({NeurPS})}}.
  \bibinfo{pages}{4563--4572}.
\newblock


\bibitem[\protect\citeauthoryear{{Wenchao Yu}, {Cheng Zheng}, and {Wei
  Cheng}}{{Wenchao Yu} et~al\mbox{.}}{2018}]%
        {135}
\bibfield{author}{\bibinfo{person}{{Wenchao Yu}}, \bibinfo{person}{{Cheng
  Zheng}}, {and} \bibinfo{person}{{Wei Cheng}}.}
  \bibinfo{year}{2018}\natexlab{}.
\newblock \showarticletitle{Learning deep network representations with
  adversarially regularized autoencoders}. In \bibinfo{booktitle}{\emph{The
  {ACM} International Conference on Knowledge Discovery and Data Mining
  ({SIGKDD})}}. \bibinfo{pages}{2663--2671}.
\newblock


\bibitem[\protect\citeauthoryear{{Wenwu Zhu}, {Xin Wang}, and {Peng
  Cui}}{{Wenwu Zhu} et~al\mbox{.}}{2020}]%
        {139}
\bibfield{author}{\bibinfo{person}{{Wenwu Zhu}}, \bibinfo{person}{{Xin Wang}},
  {and} \bibinfo{person}{{Peng Cui}}.} \bibinfo{year}{2020}\natexlab{}.
\newblock \showarticletitle{Deep learning for learning graph representations}.
\newblock  (\bibinfo{year}{2020}).
\newblock
\urldef\tempurl%
\url{https://arxiv.org/abs/2001.00293v1}
\showURL{%
\tempurl}


\bibitem[\protect\citeauthoryear{{William L. Hamilton}, {Rex Ying}, and {Jure
  Leskovec}}{{William L. Hamilton} et~al\mbox{.}}{2017}]%
        {104}
\bibfield{author}{\bibinfo{person}{{William L. Hamilton}},
  \bibinfo{person}{{Rex Ying}}, {and} \bibinfo{person}{{Jure Leskovec}}.}
  \bibinfo{year}{2017}\natexlab{}.
\newblock \showarticletitle{Inductive representation learning on large graphs}.
  In \bibinfo{booktitle}{\emph{The International Conference on Neural
  Information Processing Systems (NIPS)}}. \bibinfo{pages}{1024--1034}.
\newblock


\bibitem[\protect\citeauthoryear{{Wouter Kool}, {Herke van Hoof}, and {Max
  Welling}}{{Wouter Kool} et~al\mbox{.}}{2019}]%
        {215}
\bibfield{author}{\bibinfo{person}{{Wouter Kool}}, \bibinfo{person}{{Herke van
  Hoof}}, {and} \bibinfo{person}{{Max Welling}}.}
  \bibinfo{year}{2019}\natexlab{}.
\newblock \showarticletitle{Attention, learn to solve routing problems}. In
  \bibinfo{booktitle}{\emph{The International Conference on Learning
  Representations (ICLR)}}.
\newblock


\bibitem[\protect\citeauthoryear{{Xavier Bresson} and {Thomas Laurent}}{{Xavier
  Bresson} and {Thomas Laurent}}{2017}]%
        {119}
\bibfield{author}{\bibinfo{person}{{Xavier Bresson}} {and}
  \bibinfo{person}{{Thomas Laurent}}.} \bibinfo{year}{2017}\natexlab{}.
\newblock \showarticletitle{Residual gated graph ConvNets}.
\newblock  (\bibinfo{year}{2017}).
\newblock
\urldef\tempurl%
\url{https://arxiv.org/abs/1711.07553}
\showURL{%
\tempurl}


\bibitem[\protect\citeauthoryear{{Xiang Wang}, {Xiangnan He}, {Yixin Cao},
  {Meng Liu}, and {Tat-Seng Chua}}{{Xiang Wang} et~al\mbox{.}}{2019}]%
        {349}
\bibfield{author}{\bibinfo{person}{{Xiang Wang}}, \bibinfo{person}{{Xiangnan
  He}}, \bibinfo{person}{{Yixin Cao}}, \bibinfo{person}{{Meng Liu}}, {and}
  \bibinfo{person}{{Tat-Seng Chua}}.} \bibinfo{year}{2019}\natexlab{}.
\newblock \showarticletitle{{KGAT}: knowledge graph attention network for
  recommendation}. In \bibinfo{booktitle}{\emph{The International Conference on
  Knowledge Discovery and Data Mining ({SIGKDD})}}. \bibinfo{pages}{950--958}.
\newblock


\bibitem[\protect\citeauthoryear{{Xiao Huang}, {Qingquan Song}, {Yuening Li},
  and {Xia Hu}}{{Xiao Huang} et~al\mbox{.}}{2019}]%
        {116}
\bibfield{author}{\bibinfo{person}{{Xiao Huang}}, \bibinfo{person}{{Qingquan
  Song}}, \bibinfo{person}{{Yuening Li}}, {and} \bibinfo{person}{{Xia Hu}}.}
  \bibinfo{year}{2019}\natexlab{}.
\newblock \showarticletitle{Graph recurrent networks with attributed random
  walks}. In \bibinfo{booktitle}{\emph{The {ACM} International Conference on
  Knowledge Discovery and Data Mining ({SIGKDD})}}. \bibinfo{pages}{732--740}.
\newblock


\bibitem[\protect\citeauthoryear{{Xiao Shen} and {Fulai Chung}}{{Xiao Shen} and
  {Fulai Chung}}{2020}]%
        {141}
\bibfield{author}{\bibinfo{person}{{Xiao Shen}} {and} \bibinfo{person}{{Fulai
  Chung}}.} \bibinfo{year}{2020}\natexlab{}.
\newblock \showarticletitle{Deep network embedding for graph representation
  learning in signed networks}.
\newblock \bibinfo{journal}{\emph{{IEEE} Transactions on Cybernetics}}
  \bibinfo{volume}{50}, \bibinfo{number}{4} (\bibinfo{year}{2020}),
  \bibinfo{pages}{1556--1568}.
\newblock


\bibitem[\protect\citeauthoryear{{Xiao Wang}, {Houye Ji}, {Chuan Shi}, {Bai
  Wang}, {Peng Cui}, {Philip S. Yu}, and {Yanfang Ye}}{{Xiao Wang}
  et~al\mbox{.}}{2018}]%
        {75}
\bibfield{author}{\bibinfo{person}{{Xiao Wang}}, \bibinfo{person}{{Houye Ji}},
  \bibinfo{person}{{Chuan Shi}}, \bibinfo{person}{{Bai Wang}},
  \bibinfo{person}{{Peng Cui}}, \bibinfo{person}{{Philip S. Yu}}, {and}
  \bibinfo{person}{{Yanfang Ye}}.} \bibinfo{year}{2018}\natexlab{}.
\newblock \showarticletitle{Heterogeneous Graph Attention Network}. In
  \bibinfo{booktitle}{\emph{The World Wide Web Conference}}.
  \bibinfo{pages}{2022--2032}.
\newblock


\bibitem[\protect\citeauthoryear{{Xiao Wang}, {Ruijia Wang}, {Chuan Shi},
  {Guojie Song}, and {Qingyong Li}}{{Xiao Wang} et~al\mbox{.}}{2020}]%
        {344}
\bibfield{author}{\bibinfo{person}{{Xiao Wang}}, \bibinfo{person}{{Ruijia
  Wang}}, \bibinfo{person}{{Chuan Shi}}, \bibinfo{person}{{Guojie Song}}, {and}
  \bibinfo{person}{{Qingyong Li}}.} \bibinfo{year}{2020}\natexlab{}.
\newblock \showarticletitle{Multi-component graph convolutional collaborative
  filtering}. In \bibinfo{booktitle}{\emph{The {AAAI} Conference on Artificial
  Intelligence ({AAAI})}}.
\newblock


\bibitem[\protect\citeauthoryear{{Xiaodan Liang}, {Liang Lin}, {Xiaohui Shen},
  {Jiashi Feng}, {Shuicheng Yan}, and {Eric P. Xing}}{{Xiaodan Liang}
  et~al\mbox{.}}{2017}]%
        {267}
\bibfield{author}{\bibinfo{person}{{Xiaodan Liang}}, \bibinfo{person}{{Liang
  Lin}}, \bibinfo{person}{{Xiaohui Shen}}, \bibinfo{person}{{Jiashi Feng}},
  \bibinfo{person}{{Shuicheng Yan}}, {and} \bibinfo{person}{{Eric P. Xing}}.}
  \bibinfo{year}{2017}\natexlab{}.
\newblock \showarticletitle{Interpretable structure-evolving {LSTM}}. In
  \bibinfo{booktitle}{\emph{The {IEEE} International Conference on Computer
  Vision and Pattern Recognition ({CVPR})}}. \bibinfo{pages}{1010--1019}.
\newblock


\bibitem[\protect\citeauthoryear{{Xiaodan Liang}, {Xiaohui Shen}, {Jiashi
  Feng}, {Liang Lin}, and {Shuicheng Yan}}{{Xiaodan Liang}
  et~al\mbox{.}}{2016}]%
        {121}
\bibfield{author}{\bibinfo{person}{{Xiaodan Liang}}, \bibinfo{person}{{Xiaohui
  Shen}}, \bibinfo{person}{{Jiashi Feng}}, \bibinfo{person}{{Liang Lin}}, {and}
  \bibinfo{person}{{Shuicheng Yan}}.} \bibinfo{year}{2016}\natexlab{}.
\newblock \showarticletitle{Semantic Object Parsing with Graph {LSTM}}. In
  \bibinfo{booktitle}{\emph{The European Conference on Computer Vision
  ({ECCV})}}. \bibinfo{pages}{125--143}.
\newblock


\bibitem[\protect\citeauthoryear{{Xiaojie Guo}, {Lingfei Wu}, and {Liang
  Zhao}}{{Xiaojie Guo} et~al\mbox{.}}{2018}]%
        {145}
\bibfield{author}{\bibinfo{person}{{Xiaojie Guo}}, \bibinfo{person}{{Lingfei
  Wu}}, {and} \bibinfo{person}{{Liang Zhao}}.} \bibinfo{year}{2018}\natexlab{}.
\newblock \showarticletitle{Deep Graph Translation}.
\newblock  (\bibinfo{year}{2018}).
\newblock
\urldef\tempurl%
\url{https://arxiv.org/abs/1805.09980}
\showURL{%
\tempurl}


\bibitem[\protect\citeauthoryear{{Xin Jiang}, {Kewei Cheng}, {Song Jiang}, and
  {Yizhou Sun}}{{Xin Jiang} et~al\mbox{.}}{2020}]%
        {24}
\bibfield{author}{\bibinfo{person}{{Xin Jiang}}, \bibinfo{person}{{Kewei
  Cheng}}, \bibinfo{person}{{Song Jiang}}, {and} \bibinfo{person}{{Yizhou
  Sun}}.} \bibinfo{year}{2020}\natexlab{}.
\newblock \showarticletitle{{Chordal-GCN}: exploiting sparsity in training
  large-scale graph convolutional networks}.
\newblock  (\bibinfo{year}{2020}).
\newblock


\bibitem[\protect\citeauthoryear{{Xinyi Zhang} and {Lihui Chen}}{{Xinyi Zhang}
  and {Lihui Chen}}{2019}]%
        {52}
\bibfield{author}{\bibinfo{person}{{Xinyi Zhang}} {and} \bibinfo{person}{{Lihui
  Chen}}.} \bibinfo{year}{2019}\natexlab{}.
\newblock \showarticletitle{Capsule Graph Neural Network}. In
  \bibinfo{booktitle}{\emph{The International Conference on Learning
  Representations ({ICLR})}}.
\newblock


\bibitem[\protect\citeauthoryear{{Yang Gao}, {Hong Yang}, {Peng Zhang}, {Chuan
  Zhou}, and {Yue Hu}}{{Yang Gao} et~al\mbox{.}}{2019}]%
        {173}
\bibfield{author}{\bibinfo{person}{{Yang Gao}}, \bibinfo{person}{{Hong Yang}},
  \bibinfo{person}{{Peng Zhang}}, \bibinfo{person}{{Chuan Zhou}}, {and}
  \bibinfo{person}{{Yue Hu}}.} \bibinfo{year}{2019}\natexlab{}.
\newblock \showarticletitle{{GraphNAS}: graph neural architecture search with
  reinforcement learning}.
\newblock  (\bibinfo{year}{2019}).
\newblock
\urldef\tempurl%
\url{https://arxiv.org/abs/1904.09981v1}
\showURL{%
\tempurl}


\bibitem[\protect\citeauthoryear{{Yao Ma}, {Suhang Wang}, {Charu C. Aggarwal},
  {Dawei Yin}, and {Jiliang Tang}}{{Yao Ma} et~al\mbox{.}}{2018}]%
        {36}
\bibfield{author}{\bibinfo{person}{{Yao Ma}}, \bibinfo{person}{{Suhang Wang}},
  \bibinfo{person}{{Charu C. Aggarwal}}, \bibinfo{person}{{Dawei Yin}}, {and}
  \bibinfo{person}{{Jiliang Tang}}.} \bibinfo{year}{2018}\natexlab{}.
\newblock \showarticletitle{Multi-dimensional graph convolutional networks}.
\newblock  (\bibinfo{year}{2018}).
\newblock
\urldef\tempurl%
\url{https://arxiv.org/abs/1808.06099}
\showURL{%
\tempurl}


\bibitem[\protect\citeauthoryear{{Yao Ma}, {Suhang Wang}, {Charu C. Aggarwal},
  and {Jiliang Tang}}{{Yao Ma} et~al\mbox{.}}{2019}]%
        {82}
\bibfield{author}{\bibinfo{person}{{Yao Ma}}, \bibinfo{person}{{Suhang Wang}},
  \bibinfo{person}{{Charu C. Aggarwal}}, {and} \bibinfo{person}{{Jiliang
  Tang}}.} \bibinfo{year}{2019}\natexlab{}.
\newblock \showarticletitle{Graph convolutional networks with {EigenPooling}}.
  In \bibinfo{booktitle}{\emph{The International Conference on Knowledge
  Discovery and Data Mining ({SIGKDD})}}. \bibinfo{pages}{723--731}.
\newblock


\bibitem[\protect\citeauthoryear{{Yao Mao}, {Ziyi Guo}, {Zhaochun Ren}, {Eric
  Zhao}, {Jiliang Tang}, and {Dawei Yin}}{{Yao Mao} et~al\mbox{.}}{2018}]%
        {107}
\bibfield{author}{\bibinfo{person}{{Yao Mao}}, \bibinfo{person}{{Ziyi Guo}},
  \bibinfo{person}{{Zhaochun Ren}}, \bibinfo{person}{{Eric Zhao}},
  \bibinfo{person}{{Jiliang Tang}}, {and} \bibinfo{person}{{Dawei Yin}}.}
  \bibinfo{year}{2018}\natexlab{}.
\newblock \showarticletitle{Streaming graph neural networks}.
\newblock  (\bibinfo{year}{2018}).
\newblock
\urldef\tempurl%
\url{https://arxiv.org/abs/1810.10627}
\showURL{%
\tempurl}


\bibitem[\protect\citeauthoryear{{Yawei Luo}, {Tao Guan}, {Junqing Yu}, {Ping
  Liu}, and {Yi Yang}}{{Yawei Luo} et~al\mbox{.}}{2018}]%
        {33}
\bibfield{author}{\bibinfo{person}{{Yawei Luo}}, \bibinfo{person}{{Tao Guan}},
  \bibinfo{person}{{Junqing Yu}}, \bibinfo{person}{{Ping Liu}}, {and}
  \bibinfo{person}{{Yi Yang}}.} \bibinfo{year}{2018}\natexlab{}.
\newblock \showarticletitle{Every node counts: self-ensembling graph
  convolutional networks for semi-supervised learning}.
\newblock  (\bibinfo{year}{2018}).
\newblock
\urldef\tempurl%
\url{https://arxiv.org/abs/1809.09925v1}
\showURL{%
\tempurl}


\bibitem[\protect\citeauthoryear{{Yedid Hoshen}}{{Yedid Hoshen}}{2017}]%
        {338}
\bibfield{author}{\bibinfo{person}{{Yedid Hoshen}}.}
  \bibinfo{year}{2017}\natexlab{}.
\newblock \showarticletitle{{VAIN}: attentional multi-agent predictive
  modeling}. In \bibinfo{booktitle}{\emph{The International Conference on
  Neural Information Processing Systems ({NIPS})}}.
  \bibinfo{pages}{2698--2708}.
\newblock


\bibitem[\protect\citeauthoryear{{Yelong Shen}, {Jianshu Chen}, {Po-Sen Huang},
  {Yuqing Guo}, and {Jianfeng Gao}}{{Yelong Shen} et~al\mbox{.}}{2018}]%
        {188}
\bibfield{author}{\bibinfo{person}{{Yelong Shen}}, \bibinfo{person}{{Jianshu
  Chen}}, \bibinfo{person}{{Po-Sen Huang}}, \bibinfo{person}{{Yuqing Guo}},
  {and} \bibinfo{person}{{Jianfeng Gao}}.} \bibinfo{year}{2018}\natexlab{}.
\newblock \showarticletitle{M-Walk: learning to walk over graphs using monte
  carlo tree search}. In \bibinfo{booktitle}{\emph{The International Conference
  on Neural Information Processing Systems ({NIPS})}}.
  \bibinfo{pages}{6786--6797}.
\newblock


\bibitem[\protect\citeauthoryear{{Yifan Feng}, {Haoxuan You}, {Zizhao Zhang},
  {Rongrong Ji}, and {Yue Gao}}{{Yifan Feng} et~al\mbox{.}}{2019}]%
        {65}
\bibfield{author}{\bibinfo{person}{{Yifan Feng}}, \bibinfo{person}{{Haoxuan
  You}}, \bibinfo{person}{{Zizhao Zhang}}, \bibinfo{person}{{Rongrong Ji}},
  {and} \bibinfo{person}{{Yue Gao}}.} \bibinfo{year}{2019}\natexlab{}.
\newblock \showarticletitle{Hypergraph Neural Networks}. In
  \bibinfo{booktitle}{\emph{The {AAAI} Conference on Artificial Intelligence
  ({AAAI})}}. \bibinfo{pages}{3558--3565}.
\newblock


\bibitem[\protect\citeauthoryear{{Yihe Dong}, {Will Sawin}, and {Yoshua
  Bengio}}{{Yihe Dong} et~al\mbox{.}}{2020}]%
        {174}
\bibfield{author}{\bibinfo{person}{{Yihe Dong}}, \bibinfo{person}{{Will
  Sawin}}, {and} \bibinfo{person}{{Yoshua Bengio}}.}
  \bibinfo{year}{2020}\natexlab{}.
\newblock \showarticletitle{{HNHN}: hypergraph networks with hyperedge
  neurons}.
\newblock  (\bibinfo{year}{2020}).
\newblock
\urldef\tempurl%
\url{https://arxiv.org/abs/2006.12278}
\showURL{%
\tempurl}


\bibitem[\protect\citeauthoryear{{Yinwei Wei}, {Xiang Wang}, {Liqiang Nie},
  {Xiangnan He}, {Richang Hong}, and {Tat-Seng Chua}}{{Yinwei Wei}
  et~al\mbox{.}}{2019}]%
        {346}
\bibfield{author}{\bibinfo{person}{{Yinwei Wei}}, \bibinfo{person}{{Xiang
  Wang}}, \bibinfo{person}{{Liqiang Nie}}, \bibinfo{person}{{Xiangnan He}},
  \bibinfo{person}{{Richang Hong}}, {and} \bibinfo{person}{{Tat-Seng Chua}}.}
  \bibinfo{year}{2019}\natexlab{}.
\newblock \showarticletitle{{MMGCN}: multi-modal graph convolution network for
  personalized recommendation of micro-video}. In \bibinfo{booktitle}{\emph{The
  {ACM} International Conference on Multimedia ({MM})}}.
  \bibinfo{pages}{1437--1445}.
\newblock


\bibitem[\protect\citeauthoryear{{Yizhou Zhang}, {Yun Xiong}, {Xiangnan Kong},
  {Shanshan Li}, {Jinhong Mi}, and {Yangyong Zhu}}{{Yizhou Zhang}
  et~al\mbox{.}}{2018}]%
        {62}
\bibfield{author}{\bibinfo{person}{{Yizhou Zhang}}, \bibinfo{person}{{Yun
  Xiong}}, \bibinfo{person}{{Xiangnan Kong}}, \bibinfo{person}{{Shanshan Li}},
  \bibinfo{person}{{Jinhong Mi}}, {and} \bibinfo{person}{{Yangyong Zhu}}.}
  \bibinfo{year}{2018}\natexlab{}.
\newblock \showarticletitle{Deep Collective Classification in Heterogeneous
  Information Networks}. In \bibinfo{booktitle}{\emph{The World Wide Web
  Conference ({WWW})}}. \bibinfo{pages}{399--408}.
\newblock


\bibitem[\protect\citeauthoryear{{Yotam Hechtlinger}, {Purvasha Chakravarti},
  and {Jining Qin}}{{Yotam Hechtlinger} et~al\mbox{.}}{2017}]%
        {31}
\bibfield{author}{\bibinfo{person}{{Yotam Hechtlinger}},
  \bibinfo{person}{{Purvasha Chakravarti}}, {and} \bibinfo{person}{{Jining
  Qin}}.} \bibinfo{year}{2017}\natexlab{}.
\newblock \showarticletitle{A generalization of convolutional neural networks
  to graph-structured data}.
\newblock  (\bibinfo{year}{2017}).
\newblock
\urldef\tempurl%
\url{https://arxiv.org/abs/1704.08165}
\showURL{%
\tempurl}


\bibitem[\protect\citeauthoryear{{Youngjoo Seo}, {Micha\"{e}l Defferrard},
  {Pierre Vandergheynst}, and {Xavier Bresson}}{{Youngjoo Seo}
  et~al\mbox{.}}{2018}]%
        {109}
\bibfield{author}{\bibinfo{person}{{Youngjoo Seo}},
  \bibinfo{person}{{Micha\"{e}l Defferrard}}, \bibinfo{person}{{Pierre
  Vandergheynst}}, {and} \bibinfo{person}{{Xavier Bresson}}.}
  \bibinfo{year}{2018}\natexlab{}.
\newblock \showarticletitle{Structured sequence modeling with graph
  convolutional recurrent networks}. In \bibinfo{booktitle}{\emph{The
  International Conference on Neural Information Processing ({ICONIP})}}.
  \bibinfo{pages}{362--373}.
\newblock


\bibitem[\protect\citeauthoryear{{Yu Jin} and {Joseph F. JaJa}}{{Yu Jin} and
  {Joseph F. JaJa}}{2018}]%
        {143}
\bibfield{author}{\bibinfo{person}{{Yu Jin}} {and} \bibinfo{person}{{Joseph F.
  JaJa}}.} \bibinfo{year}{2018}\natexlab{}.
\newblock \showarticletitle{Learning graph-level representations with recurrent
  neural networks}.
\newblock  (\bibinfo{year}{2018}).
\newblock
\urldef\tempurl%
\url{https://arxiv.org/abs/1805.07683}
\showURL{%
\tempurl}


\bibitem[\protect\citeauthoryear{{Yu Rong}, {Wenbing Huang}, {Tingyang Xu}, and
  {Junzhou Huang}}{{Yu Rong} et~al\mbox{.}}{2020}]%
        {379}
\bibfield{author}{\bibinfo{person}{{Yu Rong}}, \bibinfo{person}{{Wenbing
  Huang}}, \bibinfo{person}{{Tingyang Xu}}, {and} \bibinfo{person}{{Junzhou
  Huang}}.} \bibinfo{year}{2020}\natexlab{}.
\newblock \showarticletitle{{DropEdge}: towards deep graph convolutional
  networks on node classification}. In \bibinfo{booktitle}{\emph{The
  International Conference on Learning Representations ({ICLR})}}.
\newblock


\bibitem[\protect\citeauthoryear{{Yu Zhou}, {Jianbin Huang}, {Heli Sun},
  {Yizhou Sun}, {Shaojie Qiao}, and {Stephen Wambura}}{{Yu Zhou}
  et~al\mbox{.}}{2019}]%
        {395}
\bibfield{author}{\bibinfo{person}{{Yu Zhou}}, \bibinfo{person}{{Jianbin
  Huang}}, \bibinfo{person}{{Heli Sun}}, \bibinfo{person}{{Yizhou Sun}},
  \bibinfo{person}{{Shaojie Qiao}}, {and} \bibinfo{person}{{Stephen Wambura}}.}
  \bibinfo{year}{2019}\natexlab{}.
\newblock \showarticletitle{Recurrent meta-structure for robust similarity
  measure in heterogeneous information networks}.
\newblock \bibinfo{journal}{\emph{ACM Transactions on Knowledge Discovery from
  Data}} (\bibinfo{year}{2019}), \bibinfo{pages}{No. 64}.
\newblock


\bibitem[\protect\citeauthoryear{{Yuan Li}, {Xiaodan Liang}, {Zhiting Hu},
  {Yinbo Chen}, and {Eric P. Xing}}{{Yuan Li} et~al\mbox{.}}{2019}]%
        {110}
\bibfield{author}{\bibinfo{person}{{Yuan Li}}, \bibinfo{person}{{Xiaodan
  Liang}}, \bibinfo{person}{{Zhiting Hu}}, \bibinfo{person}{{Yinbo Chen}},
  {and} \bibinfo{person}{{Eric P. Xing}}.} \bibinfo{year}{2019}\natexlab{}.
\newblock \showarticletitle{Graph Transformer}.
\newblock  (\bibinfo{year}{2019}).
\newblock
\urldef\tempurl%
\url{https://openreview.net/forum?id=HJei-2RcK7}
\showURL{%
\tempurl}


\bibitem[\protect\citeauthoryear{{Yue Zhang}, {Qi Liu}, and {Linfeng
  Song}}{{Yue Zhang} et~al\mbox{.}}{2018}]%
        {120}
\bibfield{author}{\bibinfo{person}{{Yue Zhang}}, \bibinfo{person}{{Qi Liu}},
  {and} \bibinfo{person}{{Linfeng Song}}.} \bibinfo{year}{2018}\natexlab{}.
\newblock \showarticletitle{Sentence-State {LSTM} for text representation}. In
  \bibinfo{booktitle}{\emph{The Annual Meeting of the Association for
  Computational Linguistics}}. \bibinfo{pages}{317--327}.
\newblock


\bibitem[\protect\citeauthoryear{{Yujia Li}, {Daniel Tarlow}, {Marc
  Brockschmidt}, and {Richard Zemel}}{{Yujia Li} et~al\mbox{.}}{[n.d.]}]%
        {122}
\bibfield{author}{\bibinfo{person}{{Yujia Li}}, \bibinfo{person}{{Daniel
  Tarlow}}, \bibinfo{person}{{Marc Brockschmidt}}, {and}
  \bibinfo{person}{{Richard Zemel}}.} \bibinfo{year}{[n.d.]}\natexlab{}.
\newblock \showarticletitle{Gated graph sequence neural networks}.
\newblock


\bibitem[\protect\citeauthoryear{{Yujia Li}, {Oriol Vinyals}, {Chris Dyer},
  {Razvan Pascanu}, and {Peter Battaglia}}{{Yujia Li} et~al\mbox{.}}{2018}]%
        {150}
\bibfield{author}{\bibinfo{person}{{Yujia Li}}, \bibinfo{person}{{Oriol
  Vinyals}}, \bibinfo{person}{{Chris Dyer}}, \bibinfo{person}{{Razvan
  Pascanu}}, {and} \bibinfo{person}{{Peter Battaglia}}.}
  \bibinfo{year}{2018}\natexlab{}.
\newblock \showarticletitle{Learning Deep Generative Models of Graphs}. In
  \bibinfo{booktitle}{\emph{The International Conference on Learning
  Representations ({ICLR} Workshop)}}.
\newblock


\bibitem[\protect\citeauthoryear{{Yujun Cai}, {Liuhao Ge}, {Jun Liu}, {Jianfei
  Cai}, {Tat-Jen Cham}, {Junsong Yuan}, and {Nadia Magnenat Thalmann}}{{Yujun
  Cai} et~al\mbox{.}}{2019}]%
        {57}
\bibfield{author}{\bibinfo{person}{{Yujun Cai}}, \bibinfo{person}{{Liuhao Ge}},
  \bibinfo{person}{{Jun Liu}}, \bibinfo{person}{{Jianfei Cai}},
  \bibinfo{person}{{Tat-Jen Cham}}, \bibinfo{person}{{Junsong Yuan}}, {and}
  \bibinfo{person}{{Nadia Magnenat Thalmann}}.}
  \bibinfo{year}{2019}\natexlab{}.
\newblock \showarticletitle{Exploiting spatial-temporal relationships for 3d
  pose estimation via graph convolutional networks}. In
  \bibinfo{booktitle}{\emph{The {IEEE} International Conference on Computer
  Vision ({ICCV})}}. \bibinfo{pages}{2272--2281}.
\newblock


\bibitem[\protect\citeauthoryear{{Yuyu Zhang}, {Xinshi Chen}, {Yuan Yang},
  {Arun Ramamurthy}, {Bo Li}, {Yuan Qi}, and {Le Song}}{{Yuyu Zhang}
  et~al\mbox{.}}{2020}]%
        {168}
\bibfield{author}{\bibinfo{person}{{Yuyu Zhang}}, \bibinfo{person}{{Xinshi
  Chen}}, \bibinfo{person}{{Yuan Yang}}, \bibinfo{person}{{Arun Ramamurthy}},
  \bibinfo{person}{{Bo Li}}, \bibinfo{person}{{Yuan Qi}}, {and}
  \bibinfo{person}{{Le Song}}.} \bibinfo{year}{2020}\natexlab{}.
\newblock \showarticletitle{Efficient probabilistic logic reasoning with graph
  neural networks}. In \bibinfo{booktitle}{\emph{The International Conference
  on Learning Representations ({ICLR})}}.
\newblock


\bibitem[\protect\citeauthoryear{{Yuzhou Chen}, {Yulia R. Gel}, and {Konstanin
  Avrachenkov}}{{Yuzhou Chen} et~al\mbox{.}}{2020}]%
        {44}
\bibfield{author}{\bibinfo{person}{{Yuzhou Chen}}, \bibinfo{person}{{Yulia R.
  Gel}}, {and} \bibinfo{person}{{Konstanin Avrachenkov}}.}
  \bibinfo{year}{2020}\natexlab{}.
\newblock \showarticletitle{Fractional graph convolutional networks ({FGCN})
  for semi-supervised learning}.
\newblock  (\bibinfo{year}{2020}).
\newblock
\urldef\tempurl%
\url{https://openreview.net/forum?i}
\showURL{%
\tempurl}


\bibitem[\protect\citeauthoryear{Zhang, {Shali Jiang}, {Zhicheng Cui}, {Roman
  Garnett}, and {Yixin Chen}}{Zhang et~al\mbox{.}}{2019}]%
        {79}
\bibfield{author}{\bibinfo{person}{Muhan Zhang}, \bibinfo{person}{{Shali
  Jiang}}, \bibinfo{person}{{Zhicheng Cui}}, \bibinfo{person}{{Roman Garnett}},
  {and} \bibinfo{person}{{Yixin Chen}}.} \bibinfo{year}{2019}\natexlab{}.
\newblock \showarticletitle{{D-VAE}: A Variational Autoencoder for Directed
  Acyclic Graphs}.
\newblock  (\bibinfo{year}{2019}).
\newblock
\urldef\tempurl%
\url{https://arxiv.org/abs/1904.11088}
\showURL{%
\tempurl}


\bibitem[\protect\citeauthoryear{{Zhen Zhang}, {Hongxia Yang}, {Jiajun Bu},
  {Sheng Zhou}, {Pinggang Yu}, {Jianwei Zhang}, {Martin Ester}, and {Can
  Wang}}{{Zhen Zhang} et~al\mbox{.}}{2018}]%
        {138}
\bibfield{author}{\bibinfo{person}{{Zhen Zhang}}, \bibinfo{person}{{Hongxia
  Yang}}, \bibinfo{person}{{Jiajun Bu}}, \bibinfo{person}{{Sheng Zhou}},
  \bibinfo{person}{{Pinggang Yu}}, \bibinfo{person}{{Jianwei Zhang}},
  \bibinfo{person}{{Martin Ester}}, {and} \bibinfo{person}{{Can Wang}}.}
  \bibinfo{year}{2018}\natexlab{}.
\newblock \showarticletitle{ANRL: attributed network representation learning
  via deep neural networks}. In \bibinfo{booktitle}{\emph{The International
  Joint Conference on Artificial Intelligence ({IJCAI})}}.
  \bibinfo{pages}{3155--3161}.
\newblock


\bibitem[\protect\citeauthoryear{{Zhihong Zhang}, {Dongdong Chen}, {Jianjia
  Wang}, {Lu Bai}, and {Edwin R. Hancock}}{{Zhihong Zhang}
  et~al\mbox{.}}{2019a}]%
        {51}
\bibfield{author}{\bibinfo{person}{{Zhihong Zhang}}, \bibinfo{person}{{Dongdong
  Chen}}, \bibinfo{person}{{Jianjia Wang}}, \bibinfo{person}{{Lu Bai}}, {and}
  \bibinfo{person}{{Edwin R. Hancock}}.} \bibinfo{year}{2019}\natexlab{a}.
\newblock \showarticletitle{Quantum-based subgraph convolutional neural
  networks}.
\newblock \bibinfo{journal}{\emph{Pattern Recognition}} \bibinfo{volume}{88},
  \bibinfo{number}{2019} (\bibinfo{year}{2019}), \bibinfo{pages}{38--49}.
\newblock


\bibitem[\protect\citeauthoryear{{Zhihong Zhang}, {Dongdong Chen}, {Zeli Wang},
  {Heng Li}, {Lu Bai}, and {Edwin R. Hancock}}{{Zhihong Zhang}
  et~al\mbox{.}}{2019b}]%
        {131}
\bibfield{author}{\bibinfo{person}{{Zhihong Zhang}}, \bibinfo{person}{{Dongdong
  Chen}}, \bibinfo{person}{{Zeli Wang}}, \bibinfo{person}{{Heng Li}},
  \bibinfo{person}{{Lu Bai}}, {and} \bibinfo{person}{{Edwin R. Hancock}}.}
  \bibinfo{year}{2019}\natexlab{b}.
\newblock \showarticletitle{Depth-based subgraph convolutional auto-encoder for
  network representation learning}.
\newblock \bibinfo{journal}{\emph{Pattern Recognition}} \bibinfo{volume}{90},
  \bibinfo{number}{2019} (\bibinfo{year}{2019}), \bibinfo{pages}{363--376}.
\newblock


\bibitem[\protect\citeauthoryear{{Zhijiang Guo}, {Yan Zhang}, and {Wei
  Lu}}{{Zhijiang Guo} et~al\mbox{.}}{2019}]%
        {312}
\bibfield{author}{\bibinfo{person}{{Zhijiang Guo}}, \bibinfo{person}{{Yan
  Zhang}}, {and} \bibinfo{person}{{Wei Lu}}.} \bibinfo{year}{2019}\natexlab{}.
\newblock \showarticletitle{Attention guided graph convolutional networks for
  relation extraction}. In \bibinfo{booktitle}{\emph{The Annual Meeting of the
  Association for Computational Linguistics ({ACL})}}.
  \bibinfo{pages}{241--251}.
\newblock


\bibitem[\protect\citeauthoryear{{Zhijie Deng}, {Yinpeng Dong}, and {Jun
  Zhu}}{{Zhijie Deng} et~al\mbox{.}}{2019}]%
        {176}
\bibfield{author}{\bibinfo{person}{{Zhijie Deng}}, \bibinfo{person}{{Yinpeng
  Dong}}, {and} \bibinfo{person}{{Jun Zhu}}.} \bibinfo{year}{2019}\natexlab{}.
\newblock \showarticletitle{Batch virtual adversarial training for graph
  convolutional networks}. In \bibinfo{booktitle}{\emph{The International
  Conference on Machine Learning ({ICML} Workshop)}}.
\newblock


\bibitem[\protect\citeauthoryear{{Zhitao Ying}, {Dylan Bourgeois}, {Jiaxuan
  You}, {Marinka Zitnik}, and {Jure Leskovec}}{{Zhitao Ying}
  et~al\mbox{.}}{2019}]%
        {166}
\bibfield{author}{\bibinfo{person}{{Zhitao Ying}}, \bibinfo{person}{{Dylan
  Bourgeois}}, \bibinfo{person}{{Jiaxuan You}}, \bibinfo{person}{{Marinka
  Zitnik}}, {and} \bibinfo{person}{{Jure Leskovec}}.}
  \bibinfo{year}{2019}\natexlab{}.
\newblock \showarticletitle{{GNNExplainer}: generating explanations for graph
  neural networks}. In \bibinfo{booktitle}{\emph{The International Conference
  on Neural Information Processing Systems ({NeurPS})}}.
  \bibinfo{pages}{9244--9255}.
\newblock


\bibitem[\protect\citeauthoryear{{Ziniu Hu}, {Yuxiao Dong}, {Kuansan Wang}, and
  {Yizhou Sun}}{{Ziniu Hu} et~al\mbox{.}}{2020}]%
        {114}
\bibfield{author}{\bibinfo{person}{{Ziniu Hu}}, \bibinfo{person}{{Yuxiao
  Dong}}, \bibinfo{person}{{Kuansan Wang}}, {and} \bibinfo{person}{{Yizhou
  Sun}}.} \bibinfo{year}{2020}\natexlab{}.
\newblock \showarticletitle{Heterogeneous Graph Transformer}.
\newblock  (\bibinfo{year}{2020}).
\newblock
\urldef\tempurl%
\url{https://arxiv.org/abs/2003.01332}
\showURL{%
\tempurl}


\bibitem[\protect\citeauthoryear{{Ziqi Liu}, {Chaochao Chen}, {Longfei Li},
  {Jun Zhou}, {Xiaolong Li}, {Le Song}, and {Yuan Qi}}{{Ziqi Liu}
  et~al\mbox{.}}{2019}]%
        {34}
\bibfield{author}{\bibinfo{person}{{Ziqi Liu}}, \bibinfo{person}{{Chaochao
  Chen}}, \bibinfo{person}{{Longfei Li}}, \bibinfo{person}{{Jun Zhou}},
  \bibinfo{person}{{Xiaolong Li}}, \bibinfo{person}{{Le Song}}, {and}
  \bibinfo{person}{{Yuan Qi}}.} \bibinfo{year}{2019}\natexlab{}.
\newblock \showarticletitle{{GeniePath}: graph neural networks with adaptive
  receptive paths}. In \bibinfo{booktitle}{\emph{The {AAAI} Conference on
  Artificial Intelligence ({AAAI})}}. \bibinfo{pages}{4424--4431}.
\newblock


\bibitem[\protect\citeauthoryear{{Ziwei Zhang}, {Peng Cui}, and {Wenwu
  Zhu}}{{Ziwei Zhang} et~al\mbox{.}}{2018}]%
        {2}
\bibfield{author}{\bibinfo{person}{{Ziwei Zhang}}, \bibinfo{person}{{Peng
  Cui}}, {and} \bibinfo{person}{{Wenwu Zhu}}.} \bibinfo{year}{2018}\natexlab{}.
\newblock \showarticletitle{Deep learning on graphs: a survey}.
\newblock  (\bibinfo{year}{2018}).
\newblock
\urldef\tempurl%
\url{https://arxiv.org/abs/1812.04202}
\showURL{%
\tempurl}


\bibitem[\protect\citeauthoryear{{Zonghan Wu}, {Shirui Pan}, {Fengwen Chen},
  {Guodong Long}, {Chengqi Zhang}, and {Philip S. Yu}}{{Zonghan Wu}
  et~al\mbox{.}}{2019}]%
        {4}
\bibfield{author}{\bibinfo{person}{{Zonghan Wu}}, \bibinfo{person}{{Shirui
  Pan}}, \bibinfo{person}{{Fengwen Chen}}, \bibinfo{person}{{Guodong Long}},
  \bibinfo{person}{{Chengqi Zhang}}, {and} \bibinfo{person}{{Philip S. Yu}}.}
  \bibinfo{year}{2019}\natexlab{}.
\newblock \showarticletitle{A comprehensive survey on graph neural networks}.
\newblock  (\bibinfo{year}{2019}).
\newblock
\urldef\tempurl%
\url{https://arxiv.org/abs/1901.00596}
\showURL{%
\tempurl}


\bibitem[\protect\citeauthoryear{Z\"{u}gner, {Amir Akbarnejad}, and {Stephan
  G\"{u}nnemann}}{Z\"{u}gner et~al\mbox{.}}{2018}]%
        {181}
\bibfield{author}{\bibinfo{person}{Daniel Z\"{u}gner}, \bibinfo{person}{{Amir
  Akbarnejad}}, {and} \bibinfo{person}{{Stephan G\"{u}nnemann}}.}
  \bibinfo{year}{2018}\natexlab{}.
\newblock \showarticletitle{Adversarial attacks on neural networks for graph
  data}. In \bibinfo{booktitle}{\emph{The International Conference on Knowledge
  Discovery and Data Mining ({SIGKDD})}}. \bibinfo{pages}{2847--2856}.
\newblock


\end{thebibliography}

%\appendix
%xxx
%xxx
%xxx

\end{document}